\documentclass[12pt]{article}

\usepackage[utf8]{inputenc}
\usepackage[T1]{fontenc}
\usepackage{times}
\usepackage{amsmath,amsthm,amssymb}
\usepackage{graphicx}
\usepackage{natbib}
\usepackage{url}
\usepackage{hyperref}
\usepackage{geometry}
\usepackage{lineno}
\usepackage{setspace}
\usepackage{booktabs}
\usepackage{subcaption}
\usepackage{bm}

\usepackage{multirow}
\usepackage{lscape}

\usepackage[utf8]{inputenc}
\usepackage{newunicodechar}

\newunicodechar{￥}{\textyen}
\DeclareTextCommandDefault{\textyen}{%
  \vphantom{Y}%
  {\ooalign{Y\cr\hidewidth\yenbars\hidewidth\cr}}%
}

\newcommand{\yenbars}{%
  \vbox{
     \hrule height.1ex width.4em
     \kern.15ex
     \hrule height.1ex width.4em
     \kern.3ex
  }%
}

\usepackage{pgfplots}
\pgfplotsset{width=7cm,compat=1.18}
\usepackage{pgfplotstable}

\hypersetup{
    colorlinks=true,
    linkcolor=blue,
    citecolor=blue,
    urlcolor=blue
}

\definecolor{airforceblue}{rgb}{0.36,0.54,0.66}

\title{SX-GeoTree: \underline{S}elf-e\underline{X}plaining \underline{Geo}spatial Regression \underline{Tree} Incorporating the Spatial Similarity of Feature Attributions}

\author{
    Chaogui Kang\textsuperscript{a*}, Lijian Luo\textsuperscript{a}, Qingfeng Guan\textsuperscript{b}, Yu Liu\textsuperscript{c}
}

\date{}

\newcommand{\affiliations}{
    \textsuperscript{a}National Engineering Research Center of Geographic Information System,\\ China University of Geosciences, Wuhan, China;\\
    \textsuperscript{b}School of Geography and Information Engineering,\\ China University of Geosciences, Wuhan, China;\\
    \textsuperscript{c}Institute of Remote Sensing and Geographic Information System,\\ School of Earth and Space Sciences, Peking University, Beijing, China\\
    \textsuperscript{*}Corresponding author: kangchaogui@cug.edu.cn
}

\begin{document}

\maketitle

\begin{center}
	\affiliations\\
\end{center}

\vspace{0.5cm}

\vspace{1cm}

\newpage
\begin{abstract}
\noindent
Decision trees remain central for tabular prediction but struggle with (i) capturing spatial dependence and (ii) producing locally stable (robust) explanations. We present SX-GeoTree, a self-explaining geospatial regression tree that integrates three coupled objectives during recursive splitting: impurity reduction (MSE), spatial residual control (global Moran's I), and explanation robustness via modularity maximization on a consensus similarity network formed from (a) geographically weighted regression (GWR) coefficient distances (stimulus-response similarity) and (b) SHAP attribution distances (explanatory similarity). We recast local Lipschitz continuity of feature attributions as a network community preservation problem, enabling scalable enforcement of spatially coherent explanations without per-sample neighborhood searches. Experiments on two exemplar tasks (county-level GDP in Fujian, n=83; point-wise housing prices in Seattle, n=21,613) show SX-GeoTree maintains competitive predictive accuracy (within 0.01 $R^{2}$ of decision trees) while improving residual spatial evenness and doubling attribution consensus (modularity: Fujian 0.19 vs 0.09; Seattle 0.10 vs 0.05). Ablation confirms Moran’s I and modularity terms are complementary; removing either degrades both spatial residual structure and explanation stability. The framework demonstrates how spatial similarity - extended beyond geometric proximity through GWR-derived local relationships - can be embedded in interpretable models, advancing trustworthy geospatial machine learning and offering a transferable template for domain-aware explainability.

\vspace{0.5cm}
\end{abstract}

\newpage
\section{Introduction}

Decision trees are arguably the most fundamental and popular model in data analytics and machine learning. They form the main building block of many more advanced algorithms like random forests and gradient boosting machines, which can perform on par or better with large and deep neural networks for tabular data \citep{Grinsztajn2022}. Compared with black-boxed neural networks, decision trees excel at handling mixed feature types, offering crystal interpretability for users to understand the reasoning behind predictions, and requiring few parameters to optimize on. However, decision trees have also been increasingly criticized for the limited ability to capture spatial patterns \citep{Geerts2024a} and the unwarranted robustness of model explanations \citep{Letoffe2025} in recent literature. Therefore, it is essential to adapt decision trees for geospatial tabular data and towards robust interpretability. 

To address above issues, previous research attempts have successfully adapted decision tree regressors for the geospatial prediction task \citep{Geerts2024a}. Given the geospatial tabular data containing both the attributes and location of instances, the adapted decision trees introduced multivariate geospatial splits (e.g., oblique splits and Gaussian splits) to learn from $x$- and $y$-coordinates simultaneously, enabling the predictive model to recognize spatial patterns in geographic location data more accurately and with simpler trees \citep{Geerts2023}. For simplicity, such adapted decision tree regressors are often named the geospatial regression tree (or GeoTree, for short), which provide not only more accurate predictions but also more intuitive decision boundaries, making them more interpretable than aspatial decision tree algorithms. In other words, GeoTrees improve the robustness of model explanations in terms of \emph{explicitness}, i.e. learning immediate and understandable spatial splits beyond the axis parallel splits \citep{Burkart2021}, at the cost of computational efficiency. Unfortunately, the robustness of model explanations in terms of \emph{stability}, i.e. the consensus between model explanations for similar/neighboring instances \citep{Alvarez-Melis2018a}, remains underexplored. 

Revisiting the advances in GeoTrees and model interpretability, we argue that the stability of model predictions and explanations is of utmost importance for building trustable GeoTrees. First and foremost, decision trees are prone to overfitting, where the model captures noise in the training data, leading to serious bias-variance problem \citep{Dietterich1995}. Since the bias-variance trade-off is a golden standard for the evaluation of machine learning models' performance, GeoTrees should output similar predictions for similar input instances in order to avoid over-fitting. Secondly, under the condition that GeoTrees achieve a good balance between the bias-variance trade-off, modifying either the attributes or the location of the input instance being explained slightly will not change the prediction of the model dramastically. According to the definition of feature importances/attributions, if both the model input and output are similar for two instances under inspection, model explanations for each target feature should also be similar for the corresponding instances \citep{Alvarez-Melis2018b}. Therefore, in addition to similar model predictions, GeoTrees should also output similar model explanations for similar input instances in order to assure robustness. More specifically, we argue that, \emph{in order to meet the above two requirements for enhancing GeoTrees' robustness,  the similarity between the model's input instances should be evaluated in both the attributive space and the geographical space under the context of spatial similarity assessment}. To achieve this objective, \emph{this article introduces a novel self-explaining geospatial regression tree model that considers the spatial similarity of feature attributions} (which is a popular approach for quantifying model explanations).

The remainder of this article is organized into four parts. Section 2 summarizes the methods for building the GeoTree, evaluating its predictive performance, and explaining input feature importances. Section 3 describes the proposed self-explaining geospatial regression tree incorporating the spatial similarity of feature attributions (SX-GeoTree) in details. Section 4 presents two empirical experiments and results by comparing the proposed SX-GeoTree model with baseline models. The final Section remarks on our conclusion and discussion.

\section{Literature Review}

In this section, we focus on binary decision trees for regression with numerical features, and the game theoretic Shapley values that explain the predictions of tree-based machine learning models. First, we discuss how the decision tree regressor can be adapted for geospatial prediction tasks (Section 2.1). Then, we discuss the exact TreeSHAP algorithm focused specifically on tree models that estimates Shapley values very efficiently, and how the Shapley value framework can be applied to the quantification of the importance of location and the synergies between locational and attributive features in a model (Section 2.2). Last, since this research tries to ensure the robustness of GeoTrees by satisfying multiple requirements (i.e., performance and stability), we discuss how the self-learning framework can be applied to the optimization of competing objectives in machine learning models (Section 2.3).

\subsection{GeoTree}

\begin{figure}[h!]
     \centering
     \begin{subfigure}{0.323\textwidth}
       \centering
       \includegraphics[width=\textwidth, trim={8.4cm 3cm 9.85cm 5.5cm},clip]{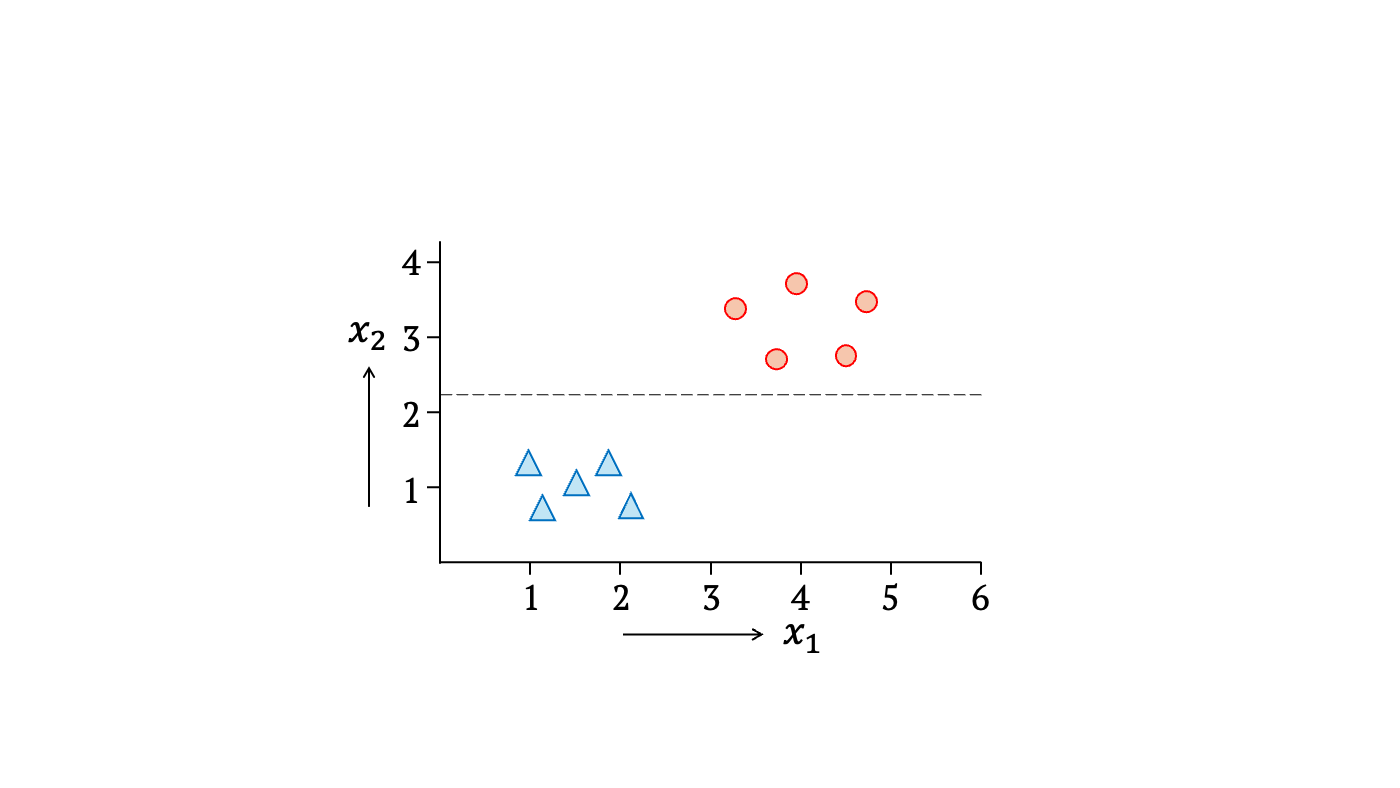}
       \caption{Axis-parallel split}
       \label{fig1a}
     \end{subfigure}
     \hfill
     \begin{subfigure}{0.323\textwidth}
       \centering
       \includegraphics[width=\textwidth, trim={8.4cm 3cm 9.85cm 5.5cm},clip]{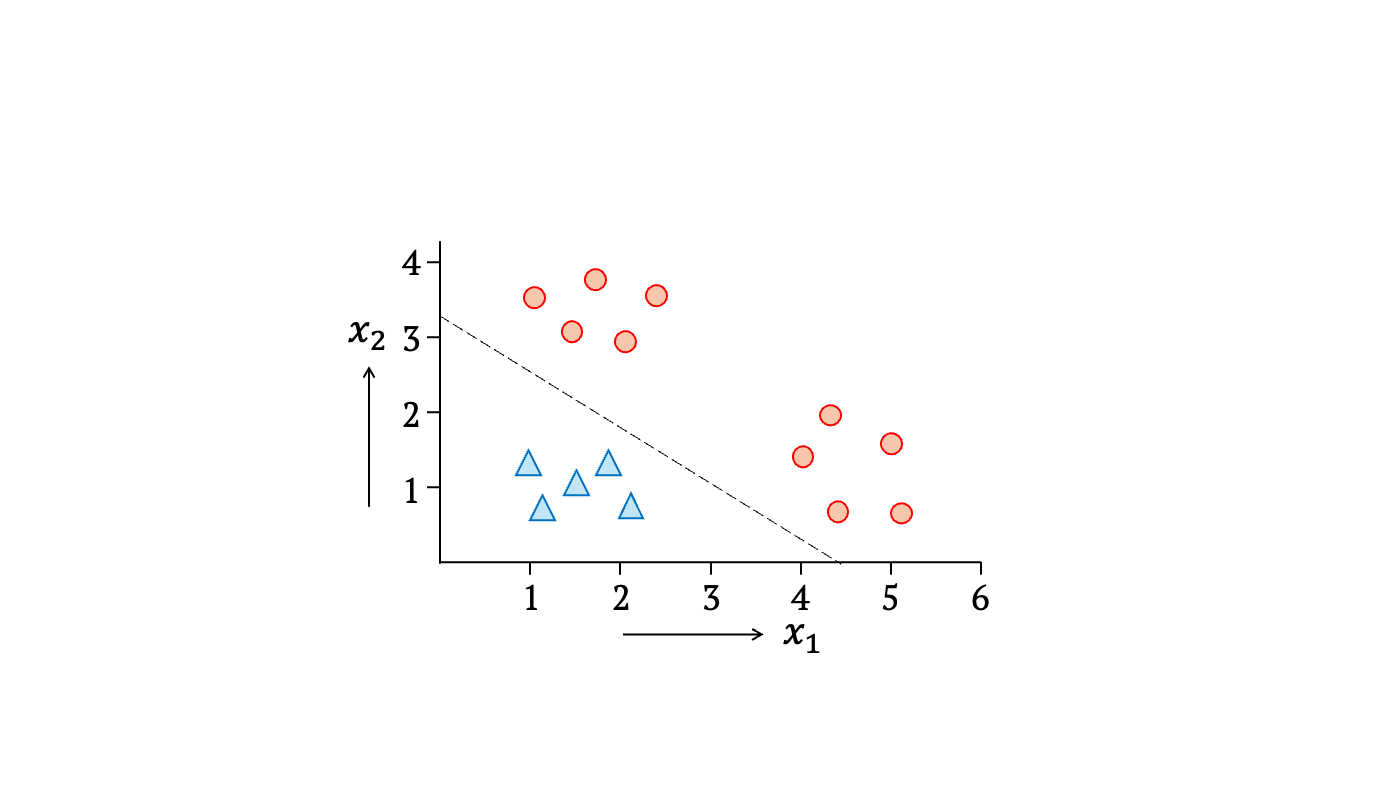}
       \caption{Oblique split}
       \label{fig1b}
     \end{subfigure}
     \hfill
     \begin{subfigure}{0.323\textwidth}
       \centering
       \includegraphics[width=\textwidth, trim={8.4cm 3cm 9.85cm 5.5cm},clip]{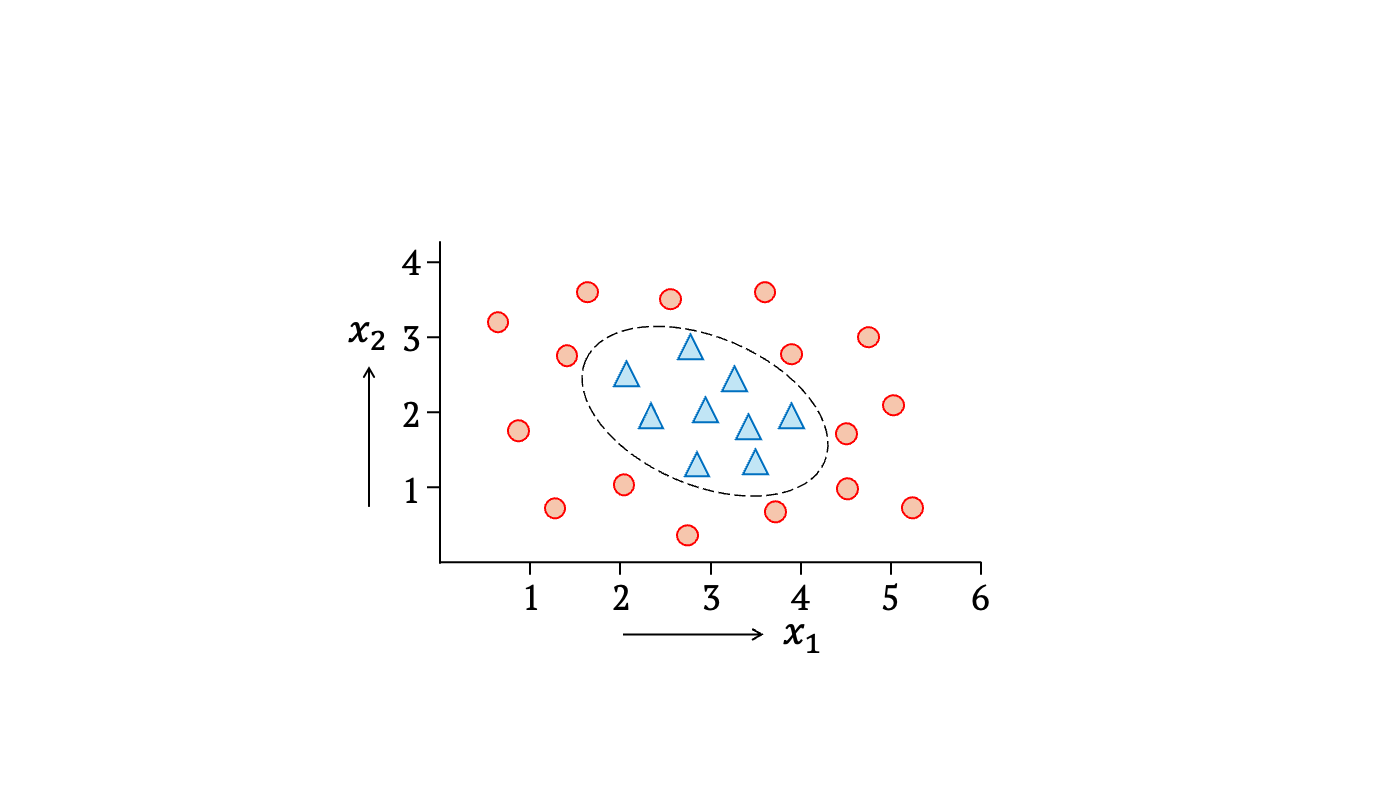}
       \caption{Gaussian split}
       \label{fig1c}
     \end{subfigure}     
     \caption{The multivariate decision tree splits in GeoTrees}
     \label{fig1}
   \end{figure}

A decision tree for regression is a model that predicts numerical values using a tree-like structure \citep{Breiman2017}. It splits data based on key features, starting from a root question and branching out. Each node asks about a feature, dividing data further until reaching leaf nodes with final predictions. Given a geospatial tabular data (see Table~\ref{tab1} for example), a regression tree takes as input an observation ($x_{1}^{\mathsf{s}}$, $x_{2}^{\mathsf{s}}$, ..., $x_{i}^{\mathsf{a}}$, $x_{i+1}^{\mathsf{a}}$, ..., $x_{n}^{\mathsf{a}}$, $y$) where $x_i$ are the features which are real-valued and the superscripts $\mathsf{s}$ and $\mathsf{a}$ indicate features are locational and attributive respectively, $n$ is the number of features and $y$ is the real-valued target variable. Then, a trained regression tree consists of splitting conditions of the form $x_i \leq c$ with $c$ a constant, producing axis-parallel splits (see Figure~\ref{fig1a}) which are not capable for recognizing complex linear and/or non-linear interactions between different spatial features \citep{Yildiz2001}. 

\begin{table}[h!]
   \centering
   \caption{Geospatial tabular data containing both locational and attributive features}
   \begin{tabular}{@{} c ccc ccccc @{}} 
      \toprule
      \textbf{Instance} & \multicolumn{3}{c}{\textbf{Location}} & \multicolumn{4}{c}{\textbf{Attribute}} & \textbf{Target} \\
      \cmidrule(r){2-4} \cmidrule(r){5-8}  
      $id$    & $longitude$ & $latitude$ & ... & $area$ ($m^2$) & $floor$ (\#)  & $room$ (\#) & ... & $price$ (\$) \\
      \midrule
      1      & 20.78 & 43.65 & ... & 100.5 & 15 & 3 & ... & 1025\\
      2         & 15.94     &  50.01 & ... & 78.8 & 10 & 2 & ... & 890\\
      3       & 17.88  & 32.50 & ... & 95.2  & 25 & 3 & ... & 975 \\
      ... & ...   &  ... & ... &  ... & ...   &  ... & ... & ... \\
      \bottomrule
   \end{tabular}
   \label{tab1}
\end{table}

To address this limitation, GeoTrees extend the conventional decision tree for regression by the inclusion of two specific bivariate decision tree splits (see Figure \ref{fig1b} and \ref{fig1c} for example): (1) \emph{the oblique split} that divides the input space in two regions based on the linear combination of locational coordinates, $w_{1}*x_{1}^{\mathsf{s}} + w_{2}*x_{2}^{\mathsf{s}} \leq c$, where $w_{1}$ and $w_{2}$ are learnable weights \citep{Murthy1994}; and (2) \emph{the Gaussian split} that defines an ellipse in a two-dimensional space based on the non-linear combination of locational coordinates, $\sqrt{ (x_{1}^{\mathsf{s}} - x_1^{\mathcal{F}_{1}})^2 + (x_{2}^{\mathsf{s}} - x_2^{\mathcal{F}_{1}})^2} + \sqrt{ (x_{1}^{\mathsf{s}} - x_1^{\mathcal{F}_{2}})^2 + (x_{2}^{\mathsf{s}} - x_2^{\mathcal{F}_{2}})^2} \leq c$, where $\mathcal{F}_{1}$ and $\mathcal{F}_{2}$ are two learnable focal points, and then separates input data located on or inside of the ellipse from input data located outside of the ellipse \citep{Geerts2023}. The combination of these bivariate split types with the conventional axis-parallel split enables GeoTrees to recognize more immediate and understandable spatial patterns in geographic location data and with simpler trees. Due to their enhanced ability in spatial splitting, GeoTrees and their variations can even outperform traditional spatial statistical models (including GWR, Kriging) and machine learning and deep learning methods (including MLP, GNN) across a range of geospatial prediction tasks through an extensive benchmark \citep{Geerts2024a}. 

To build an optimal GeoTree, an additional requirement is to utilize the pruning regularization method to make the tree smaller to prevent overfitting. The key challenge in regularization is striking a balance between underfitting and overfitting, i.e. the bias-variance trade-off \citep{Belkin2019}. Under-regularized trees might be too deep and capture noise (overfit), while over-regularized trees might be too shallow and miss important patterns (underfit). For decision trees, pruning regularization is often achieved by: (1) Early stopping that halts the growth of a decision tree during the training process based on certain predefined criteria including maximum depth of the tree, minimum number of samples required to split a node, and minimum information gain; or (2) Post-pruning that first builds a complete tree and then removes or collapses branches that don’t significantly contribute to the model’s performance based on the cost-complexity measure, $R_{\alpha}(T)=R(T)+\alpha |\tilde{T}|$, where $\alpha$ ($\geq 0$) is the learnable complexity parameter, $|\tilde{T}|$ is the number of leaf nodes in a given tree $T$, and $R(T)$ is the total sample weighted squared errors of the leaf nodes. Ideally, the regularization parameters should be determined using cross-validation or a separate validation dataset, ensuring that the settings generalize well to new data \citep{Ghojogh2019}.

With all the above considerations, a well-regularized GeoTree learns a prediction function $f(\boldsymbol{x}) \rightarrow y$ that is robust, i.e., not overly sensitive to changes in the input $\boldsymbol{x}=$\{$x_{1}^{\mathsf{s}}$, $x_{2}^{\mathsf{s}}$, ..., $x_{i}^{\mathsf{a}}$, $x_{i+1}^{\mathsf{a}}$, ..., $x_{n}^{\mathsf{a}}$\}. To put it more accurately, the learned predictive function holds the Lipschitz continuity, $L(\boldsymbol{x}) = \mathop{\arg\max}\limits_{\boldsymbol{x}_{i} \neq \boldsymbol{x}_{j}} \frac{\left\Vert f(\boldsymbol{x}_{i}) - f(\boldsymbol{x}_{j}) \right\Vert_{2}}{\left\Vert \boldsymbol{x}_{i} - \boldsymbol{x}_{j} \right\Vert_{2}} \leq K$, where the smallest Lipschitz constant $K$ ($\geq 0$) indicates the maximum absolute change in the function value per unit norm change in the input \citep{Gouk2021}. Typically, achieving point-wise robustness in the entire input domain (i.e., the global Lipschitz criterion associated with a extremely small Lipschitz constant $K$) would imply an excessively large bias and essentially a useless prediction function \citep{Geman1992}. Instead, the learned predictive function $f(\boldsymbol{x})$ is preferable to be locally Lipschitz continuous for every point $\boldsymbol{x}_{i}$ of interest, $L(\boldsymbol{x}_{i}) = \mathop{\arg\max}\limits_{\boldsymbol{x}_{j} \in \mathcal{N}_{\epsilon}(\boldsymbol{x}_{i})} \frac{\left\Vert f(\boldsymbol{x}_{i}) - f(\boldsymbol{x}_{j}) \right\Vert_{2}}{\left\Vert \boldsymbol{x}_{i} - \boldsymbol{x}_{j} \right\Vert_{2}} \leq K$, over a large and representative subset of the input space $\mathcal{N}_{\epsilon}(\boldsymbol{x}_{i})=\{\boldsymbol{x}_{j} \in \boldsymbol{x} \mid \left\Vert \boldsymbol{x}_{i} - \boldsymbol{x}_{j} \right\Vert_{2} \leq \epsilon\}$ with respect to the underlying data distribution (like within a bandwidth $\epsilon$ of the training samples). Otherwise, we can expect that the learned function which varies too drastically in response to minuscule changes in the input, especially when the input is near the decision boundary or from another representative region, would struggle to generalize well on unseen test data. 

Despite the pruning regularization and the local Lipschitz continuity, GeoTrees are unable to fully capture spatial effects in geographically referenced data, overlooking spatial disparities in their predictions \citep{Kopczewska2022}. Recent advancements prove that significant levels of spatial autocorrelation remain present in the errors of machine learning models \citep{Song2023}. Residual spatial autocorrelation suggests the potential to gain predictive power by pushing the model to better capture spatial relationships in the data, and three approaches have often been applied for addressing spatial dependencies in prediction methods: spatial feature generation, spatial dependency in model structure, and spatial regularization in the objective function. Among them, methods that account for spatial autocorrelation in the objective function, allow for a more direct approach to learn spatial dependencies in the data and simultaneously increase prediction accuracy \citep{Geerts2024b, Zhuang2025}. Leveraging a tailored loss function combining the global Moran’s I of model residuals and/or the local Moran’s I (e.g., the local indicator of spatial association, LISA) of model predictions with the conventional regression loss, the tailored models optimize not only the error with respect to the target but also the loss with respect to the spatial autocorrelation of the target in the regularization form: $\mathcal{L}(f(\boldsymbol{x})) = \mathcal{L}_{1}(y, \hat{y}) + \lambda_{1} \mathcal{L}_{2}(\rho_{_{LISA}}(y), \rho_{_{LISA}}(\hat{y})) + \lambda_{2} | \rho_{_{MoranI}}(y-\hat{y}) |$, where $\mathcal{L}_{1}$ and $\mathcal{L}_2$ are any regression loss such as mean squared error (MSE), $y$ is the target variable, $\hat{y}$ is the predicted variable, $\lambda_{1}$ and $\lambda_{2}$ are the regularization parameters balancing the contribution of each term, and $\rho$ is a function of spatial autocorrelation (such as Moran's I and LISA). To build an optimal GeoTree, the final loss function therefore consists of a combination of the loss of the main task and the loss of the auxiliary task, i.e. the predicted global/local Moran’s I.

\subsection{TreeSHAP}

Model interpretability is essential for creating GeoTree models that are not only accurate but also trustworthy and actionable. It not only boosts trust but also helps us uncover potential biases, improve model performance, and even explain results to non-technical stakeholders \citep{Lipton2018}.  One of the simplest ways to understand a model is feature importance that tells which variables (or “features”) in the input dataset have the most influence on a model’s predictions from a global view, i.e. an overarching idea of which features are important across all predictions. For decision trees, it is often measured using the Gini importance by ranking features based on how much they reduce uncertainty in splits. However, according to the Gini importance in decision trees, high-cardinality features (like unique IDs or very detailed categories) can end up looking more important than they actually are, thus presenting a skewed understanding of what really drives a model’s decisions \citep{Nembrini2018}.

\begin{figure}[h!] 
   \centering
   \includegraphics[width=\textwidth, trim={0 6.5cm 0 4cm},clip]{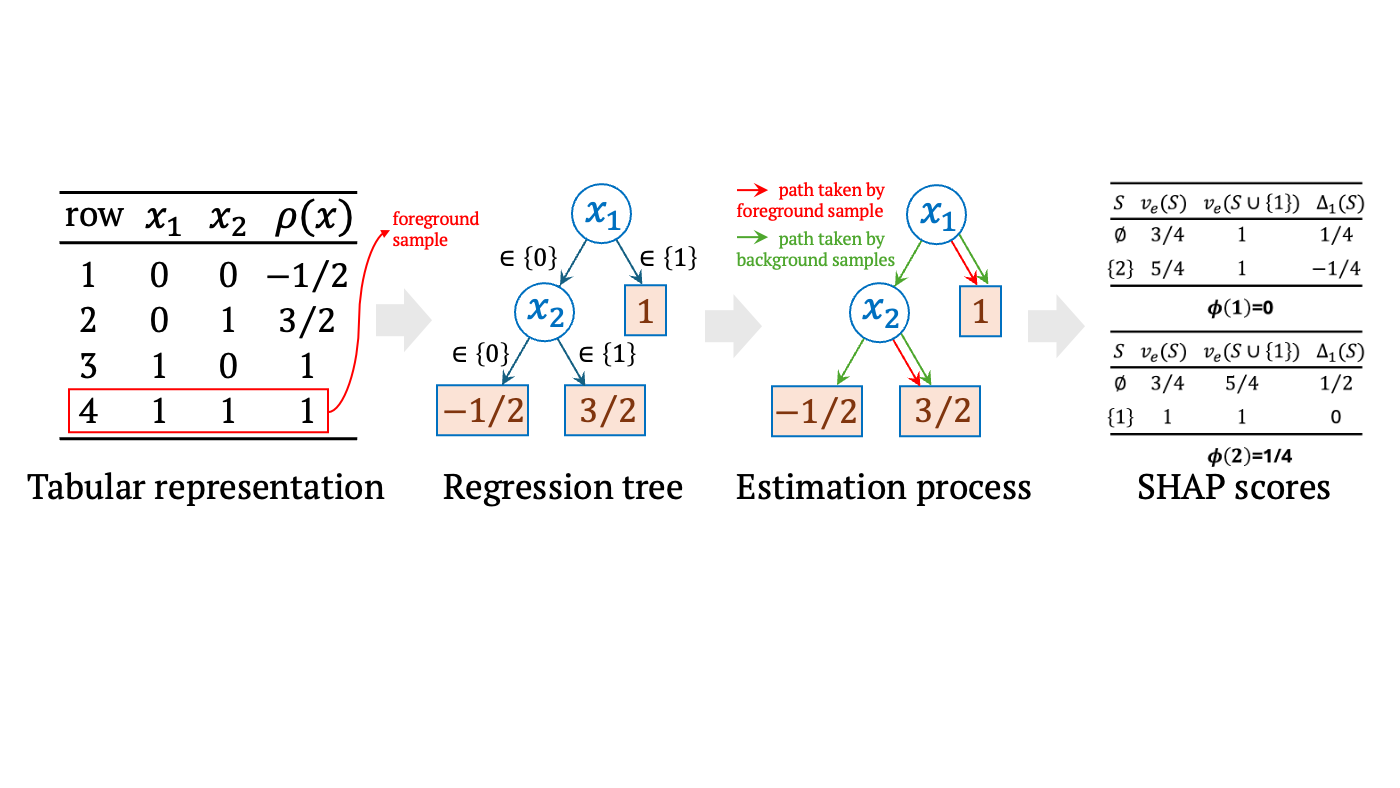} 
   \caption{Illustration of the feature contribution and expected value estimation process using Tree SHAP. Note that $v_{e}$ is the expected values given the known features.}
   \label{fig2}
\end{figure}

An alternative approach is feature attribution that determines fair payouts in coalitional games based on the average marginal contribution where multiple features contribute to a final outcome. Due to new efficient algorithms such as SHapley Additive exPlanations (SHAP), game theoretic Shapley values that determines the individual contribution of each player to the overall outcome considering all possible combinations of features \citep{Shapley1953}, $\phi_{i}(f, \boldsymbol{x})=\sum\limits_{S \subseteq N \backslash \{i\}}\frac{|S|!(n-|S|-1)!}{n!} [f_{\boldsymbol{x}}(S\cup\{i\})-f_{\boldsymbol{x}}(S)]$ (where $N$ is the entire set of features, $S$ represents any subset of features that doesn’t include the $i$-th feature and $|S|$ is the size of that subset, and $f$ (or $f_{x}(\cdot)$) represents the learned prediction function), have recently become a popular way to explain the predictions of tree-based machine learning models. In specific, when applying Shapley's theory to explain decision trees (see Figure~\ref{fig2} for example), SHAP values for a specific sample $\boldsymbol{x}^{\mathsf{f}}=\{x_{1}^{\mathsf{f}}, ... , x_{i}^{\mathsf{f}}, ... , x_{n}^{\mathsf{f}}\} =x_{\{1, ..., i, ..., n\}}^{\mathsf{f}} $ can be efficiently estimated as an average of simpler problems \citep{Lundberg2020}: $\phi_{i}(f, \boldsymbol{x}^{\mathsf{f}}) =  \frac{1}{|D|} \sum\limits_{\boldsymbol{x}^{\mathsf{b}} \in D} \phi_{i}(f, \boldsymbol{x}^{\mathsf{f}}, \boldsymbol{x}^{\mathsf{b}}) = \frac{1}{n|D|} \sum\limits_{\boldsymbol{x}^{\mathsf{b}} \in D} \sum\limits_{S \subseteq N \backslash \{i\}} \allowbreak\binom{n-1}{|S|}^{-1} [ f(x_{S \cup \{i\}}^{\mathsf{f}}, x_{N \backslash \{S \cup \{i\}\}}^{\mathsf{b}} ) - f(x_{S}^{\mathsf{f}}, x_{N \backslash S}^{\mathsf{b}}) ]$, by intervening on features in the \emph{foreground} sample $\boldsymbol{x}^{\mathsf{f}}$ (i.e., the sample being explained) with features from the \emph{background} sample $\boldsymbol{x}^{\mathsf{b}}= \{x_{1}^{\mathsf{b}}, ... , x_{i}^{\mathsf{b}}, ... , x_{n}^{\mathsf{b}}\} = x_{\{1, ..., i, ..., n\}}^{\mathsf{b}}$ (i.e., the sample the foreground sample is compared to), where $|D|$ is the number of samples in the background (underlying) distribution. In essence, SHAP values quantify not just the overall importance of a feature (e.g., by averaging local feature attributions over all samples in the input dataset $\boldsymbol{X}$ like traditional methods, $\phi_{i}(f)=\frac{1}{|\boldsymbol{X}|} \sum\limits_{\boldsymbol{x}^{\mathsf{f}} \in \boldsymbol{X}} |\phi_{i}(f, \boldsymbol{x}^{\mathsf{f}})|$), but also how that feature impacts each individual prediction (i.e., $\phi_{i}(f, \boldsymbol{x}^{\mathsf{f}})$). Unlike the Gini importance, tree SHAP values are robust to bias from high-cardinality \citep{Young1985} or correlated features \citep{Fujimoto2006}, and therefore provide a more reliable and nuanced quantification of feature importances.

The original Shapley value considers only individual features. However, in the context of geospatial regression analyses, treating locational features, such as pairs of coordinates or more complicated location embeddings, as separate features in post-hoc model explanations is not sufficient to uncover the underlying spatial effects between them. To adapt Shapley values for the combination of (locational) features, indicators for measuring the average contribution of a feature combination (also known as a coalition) with and without individual contributions of its constituent elements have been introduced: (1) The former is often realized based on the Shapley generalized values \citep{Marichal2007} that directly apply the original Shapley value for feature sets. For example, the GeoShapley value \citep{Li2024} was proposed to accommodate locational features as a single joint player $x_{\mathsf{GEO}} = \{x_{1}^{\mathsf{s}}, x_{2}^{\mathsf{s}}, ..., x_{g}^{\mathsf{s}}\}$ (i.e., $\mathsf{GEO}=\{1, 2, ..., g\}$ where $g$ is the number of locational features) in the model explanation stage: $\phi_{\mathsf{GEO}}(f, \boldsymbol{x}) = \sum\limits_{S \subseteq N \backslash \{\mathsf{GEO}\}} \frac{|S|!(n-|S|-g)!}{(n-g+1)!} [f_{\boldsymbol{x}}(S \cup \mathsf{GEO}) - f_{\boldsymbol{x}}(S)]$. Other estimators based on the Shapley generalized values include the Shapley group value \citep{Flores2019}, the Shapely sets \citep{Sivill2023}, to list a few. These new indices have enabled the direct measurement of the total marginal contribution to the model prediction from all locational features while holding other features constant; (2) The latter attributes the interaction effects by defining the Shapley interaction value between features isolated from each individuals' contributions \citep{Grabisch1999}. The most popular estimator in this strand is the SHAP interaction value \citep{Lundberg2018, Muschalik2024} that measures the local feature interaction effect $\phi_{i,j}(f, \boldsymbol{x})$ of the $i$-th feature w.r.t. the $j$-th feature:  $\phi_{i,j}(f, \boldsymbol{x}) = \sum\limits_{S \subseteq N \backslash \{\mathsf{i,j}\}}\frac{|S|!(n-|S|-2)!}{2(n-1)!} \nabla_{i,j}(f, \boldsymbol{x}, S)$, where $\nabla_{i,j}(f, \boldsymbol{x}, S) = [f_{\boldsymbol{x}}(S \cup \{i,j\}) - f_{\boldsymbol{x}}(S \cup \{j\})] - [f_{\boldsymbol{x}}(S \cup \{i\}) - f_{\boldsymbol{x}}(S)]$. Instead of the average marginal contribution of different features together, the SHAP interaction value can be interpreted as (a half of) the difference between SHAP values of the $i$-th feature with and without the $j$-th feature. Other axiomatic estimators of the interaction value include the added Shapley value \citep{Alshebli2019}, the Shapley-Taylor interaction index \citep{Sundararajan2020}, the Shapley residuals \citep{Kumar2021}, the joint Shapley value \citep{Harris2022} and the faith-interaction index \citep{Tsai2023}. By utilizing these methods, the main effect of an individual feature and its interaction effects with other features can be explored thoroughly.

Despite the many advances in tree explainers, recent work has uncovered that exact SHAP scores might assign unsatisfactory importance to features, which mislead with respect to relative feature influence \citep{Huang2024, Letoffe2025}. Such observations reveal apparent limitations with the theoretical foundations of SHAP scores. In addition to the problematic SHAP values assigned for different features, recent work has also uncovered the absence of a crucial property of SHAP values assigned for different samples; That is, meaningful interpretability methods should satisfy to generate robust explanations to local perturbations of the input \citep{Alvarez-Melis2018b}. Similar to the local Lipschitz criterion for model predictions, the robustness of an explanation model $f_{shap}$ can be evaluated in terms of its local Lipschitz constant $\hat{L}(\boldsymbol{x}_{i})$, for every point $\boldsymbol{x}_{i}$ of interest, by solving the optimization problem: $\hat{L}(\boldsymbol{x}_{i}) =  \mathop{\arg\max}\limits_{\boldsymbol{x}_{j} \in \mathcal{N}_{\epsilon}(\boldsymbol{x}_{i})} \frac{\left\Vert f_{shap}(\boldsymbol{x}_{i}) - f_{shap}(\boldsymbol{x}_{j}) \right\Vert_{2}}{\left\Vert \boldsymbol{x}_{i} - \boldsymbol{x}_{j} \right\Vert_{2}} \leq K$, where $\mathcal{N}(\boldsymbol{x}_{i})$ is a ball of radius $\epsilon$ centered at the anchor point $\boldsymbol{x}_{i}$ as aforementioned. Therefore, during the training process of machine learning models, an additional penalty $\mathcal{L}(f_{shap}(\boldsymbol{x}))$ with regard to the robustness of continuous model explanations should be considered together with the conventional optimization of model predictions $\mathcal{L}(f(\boldsymbol{x}))$ on the objective \citep{Alvarez-Melis2018a}, yielding the refined loss: $\mathcal{L} = \mathcal{L}(f(\boldsymbol{x})) + \xi \mathcal{L}(f_{shap}(\boldsymbol{x}))$, where $\xi$ is a learnable regularization parameter.

\subsection{Learning by Self-Explaining}

As discussed above, to train an optimal GeoTree is to optimize multiple objectives simultaneously (i.e., multi-objective/task learning), concerning the models' accuracy, cost-complexity, local Lipschitz of predictions, spatial disparity of residuals, and local Lipschitz of explanations. According to the taxonomy of explainable AI methods, the task inherently asks for a learning-by-self-explaining paradigm that imposes explainability on a model during training itself \citep{Elton2020}. This is because, for a multi-objective learning task that accounts for both accuracy and explainability, the conventional post-hoc model explanation methods as they try to interpret the decisions of a machine learning model after the model has been trained \citep{Retzlaff2024}, does not provide feedback on quality of explanations to the base (prediction) task. It implies that ``good'' explanations derived from the learner model are not actually used in the models' own decision processes. The learning-by-self-explaining approach overcomes this limitation via imposing feedbacks with regard to explainability on a model during training itself. 

In the learning-by-self-explaining framework, the underlying idea is that a learner model, in addition to optimizing for the original predictive task, is further optimized based on explanatory feedback from an internal critic model \citep{Stammer2024}. Its goal is to ensure that the learner makes predictions based on ``good'' explanations \citep{Alvarez-Melis2018a}. Therefore, the learner jointly optimizes its explanations based on the critic’s feedback, but also based on its predictive performance for the original task. The exact procedure of how to integrate the critic’s feedback depends on the submodel types as well as on the form of the critic’s feedback. The most intuitive way to achieve the goal is using the Lagrangian approach \citep{Lin1979} that first defines a specific loss for each objective and then combine all the losses linearly as one weighted objective function based on regularization techniques. However, in such a multi-objective problem, different tasks may inherently conflict, necessitating a trade-off \citep{Sener2018}. The naive workaround that minimizes a weighted linear combination of per-task losses is only valid when the tasks do not compete. For example, much research has shown that just adding a regularization term can overlook complex interdependencies between objectives and lead to suboptimal trade-offs \citep{Wegel2025}. Therefore, to balance performance and interpretability, a one-size-fits-all loss that folds the secondary objective as a weighted regularization term should be applied with caution in multi-objective learning tasks.

Alternatively, more recent work has proposed a bilevel optimization strategy that treats the multi-objective problem as two linked sub-problems aligning with a Stackelberg game (i.e., a leader model and a follower model) instead of a single blended objective \citep{Yazdani2024}. Having the two models that interact with a defined order of optimization (i.e., the leader optimizes the primary loss (e.g., accuracy) with respect to its own parameters first, then the follower treats the leader’s parameters as fixed (for that iteration) and optimizes the secondary loss (e.g., interpretability or another constraint) with respect to its own parameters), bilevel optimization gives each objective its own ``space'' (layers, parameters, even optimizer), yielding cleaner design and often better performance on the primary task all while meeting secondary goals to a Pareto-optimal degree. It is noteworthy that the bilevel setup is particularly suitable for problems that can be modeled by neural networks in that a bilevel neural network can be easily solved by alternating optimization \citep{Bezdek2003}: at each training iteration, the model updates one set of parameters while holding the other fixed, and then vice versa. Fortunately, a decision tree can be effectively reframed as a one-layer neural network with softmax as its activation function \citep{Yang2018}: $\pi = f_{w,b,\tau}(x)=\text{softmax}((wx+b)/\tau)$, where $w$ is a constant rather than a trainable variable and its value is set as $w=[1,2,...,n+1]$, $b$ is constructed as $b=[0, -\beta_{1}, -\beta_{1}-\beta_{2}, ..., -\beta_{1}-\beta_{2}-...-\beta_{n}]$ and the learnable parameters ($\beta_{i}, \beta_{i+1}$) define the interval that $x$ is split, and $\tau>0$ is a temperature factor. Therefore, the bilevel neural network approach can provide a promising tool for building optimal GeoTrees in the form of neural networks (also known as soft decision trees). Compared with hard decision tree, such as neural network based approach is more complicated, parameter-heavy, and difficult to train and scale.

\section{Methodology}

In this section, we introduce the proposed self-explaining geospatial regression tree (i.e. SX-GeoTree) model. First, Section 3.1 presents the overall architecture of the proposed model, which includes modules tailored for bridging the model's performance and robustness. Secondly, Section 3.2 elaborates on the method to quantify the spatial similarities of the model's inputs, predictions and explanations by leveraging geographically weighted regression and network modularization techniques. Lastly, Section 3.3 describes the metrics for the comparative analyses between the proposed model and the baseline models.

\subsection{Architecture of the Proposed Model}

Our motivation is to predict a response $Y$ from inputs (as known as predictors) $X_{1}$, $X_{2}$, ..., $X_{n}$ using GeoTree. The general idea is that we will segment the high dimensional predictor space - that is, the set of possible values for $X_{1}$, $X_{2}$, ..., $X_{n}$ -  into $J$ distinct and non-overlapping regions, $R_{1}$, $R_{2}$, ..., $R_{J}$, by growing a binary tree. In keeping with the tree analogy, the regions $R_{j}$ are known as terminal nodes or leaves of the tree. The points along the tree where the predictor space is split are referred to as internal nodes. At each internal node in the tree, we apply a test (i.e. the splitting rule including axis-parallel split, oblique split and Gaussian split) to the inputs, say $X_{i}$, and then, depending on the outcome of the test, go to either the left-hand branch (corresponds to $X_{i} < t_{k}$) or the right-hand branch (corresponds to $X_{i} \geq t_{k}$) resulting from that split. As a GeoTree is drawn upside down following a series of splits, we eventually come to a leaf node (i.e. a region $R_{j}$), where we make a prediction which is simply the mean of the response values for the training observations in that region.

\begin{figure}[h!] 
   \centering
   \includegraphics[width=\textwidth]{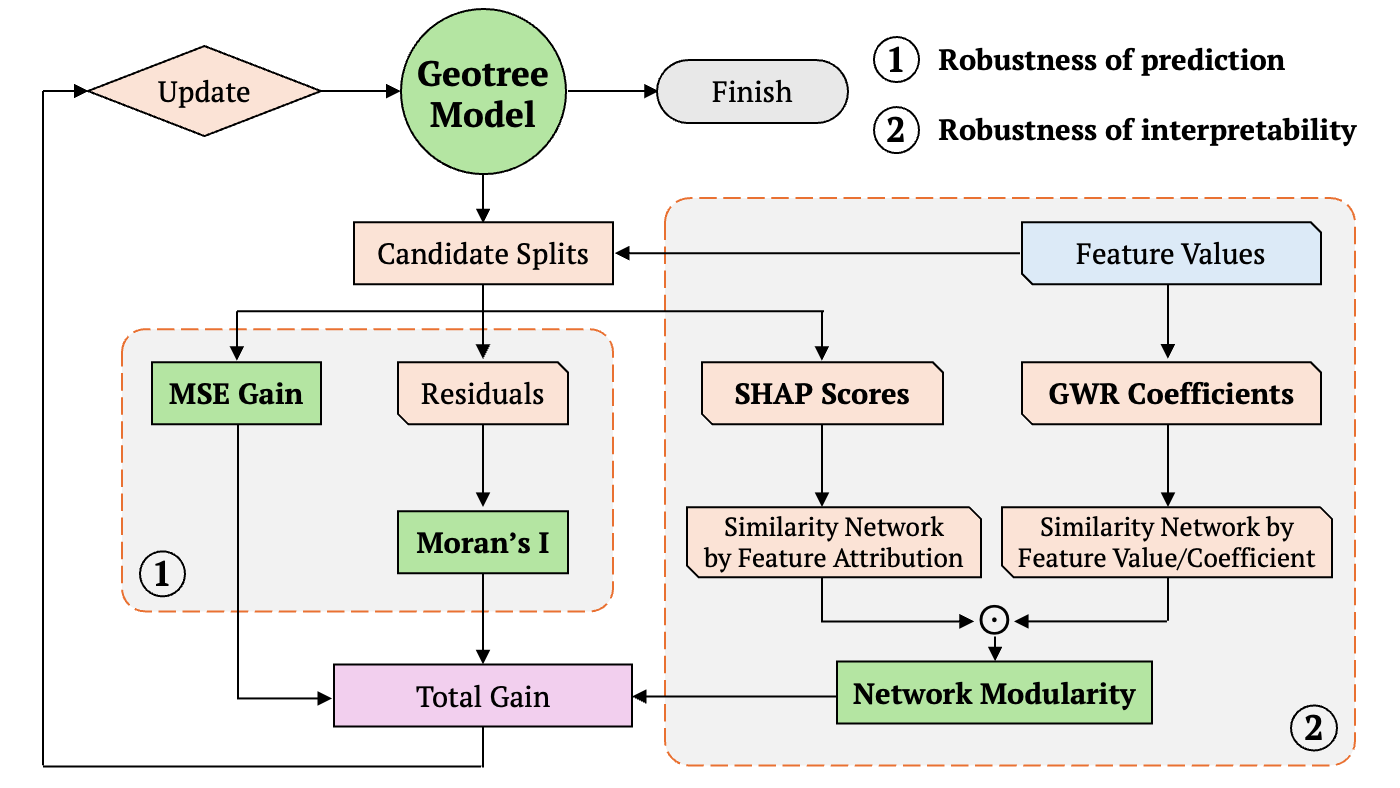} 
   \caption{The proposed SX-GeoTree model}
   \label{fig3:framework}
\end{figure}

To facilitate self-explaining, the GeoTree is built according to the top-down, greedy approach called recursive binary splitting. It begins at the top of the tree (all observations below to a single region) and then successively splits the predictor space. Each split is indicated via two new branches further down on the tree. We refer to the subtree consisting of the nodes and segments that is tested before the current splitting nodes as a \emph{sub-model}. During the tree building process, the best split that follows the current sub-model is greedily made at that particular split rather than looking ahead and picking a split that will lead to a better tree in a future split. To choose the best split next to the current sub-model (as illustrated in Figure \ref{fig3:framework}), we explicitly develop ad-hoc modules that concerns the updated sub-model's reduction of impurity, spatial disparity of residuals and robustness of explanations. First, in the module with regard to the reduction of impurity, the mean squared error (MSE) that measures the average squared difference between the estimated values (i.e. predictions) and the actual value (i.e. observations) is taken as a crucial criterion to each candidate split nodes:
\begin{align} 
   \mathcal{L}_{MSE} = \frac{1}{n} \sum_{i} (y_{i} - \hat{y}_{_{R_{m^{'}}}})^2  - \frac{1}{n} \sum_{i} (y_{i} - \hat{y}_{_{R_{m}}})^2 
\end{align}
where $n$ is the number of observations, $R_{m^{'}}$ and $R_{m}$ are the regions corresponding to the $m$-th terminal node in the sub-model before and after the current splitting node is considered, $\hat{y}_{_{R_{m^{'}}}}$ and $\hat{y}_{_{R_{m}}}$ are the predicted responses associated with $R_{m^{'}}$ and $R_{m}$, respectively. This module aims to maximize the difference between the MSE of the parent node and the weighted MSE of the child nodes. Secondly, in the module with regard to the spatial disparity of residuals, the global Moran's I index that measures the levels of spatial autocorrelations remaining in the predictive errors of the sub-model is further considered:
\begin{align} 
   \mathcal{L}_{MoranI} = \frac{N}{W} \frac{\sum_{i}\sum_{j} w_{ij}(r_{i} - \bar{r})(r_{j} - \bar{r})}{\sum_{i} (r_{i} - \bar{r})^2}
\end{align}
where $N$ is the number of spatial units (i.e. instances, $N=n$) indexed by $i$ and $j$, $W$ is the sum of all $w_{ij}$; $r_{i} =\hat{y}_{_{R_{m}}} - y_{i}$ is the residual of model prediction at the $i$-th location, $\bar{r}$ is the mean of $r$, and $w_{ij}$ is the adjacency matrix of spatial weights. This module mitigates spatial disparity in the model's prediction. Lastly, in the module with regard to the robustness of explanations, we extend the local Lipschitz continuity by considering similar instances from a geographical perspective and solve the local continuity requirement as a novel network modularity maximization problem. Compared with previous modules, this module is the core methodological contribution of the proposed SX-GeoTree model. Therefore, we elaborate on the proposed network modularity maximization module in a separate section below. 

\subsection{Enhancing the Spatial Similarity of Feature Attributions}


As discussed in Section 2.2, the robustness of an explanation model $f_{shap}$ can be evaluated in terms of its local Lipschitz function $\hat{L}(\boldsymbol{x}_{i})$, for every point $\boldsymbol{x}_{i}$ of interest, by solving the optimization problem: $\hat{L}(\boldsymbol{x}_{i}) =  \mathop{\arg\max}\limits_{\boldsymbol{x}_{j} \in \mathcal{N}_{\epsilon}(\boldsymbol{x}_{i})} \frac{\left\Vert f_{shap}(\boldsymbol{x}_{i}) - f_{shap}(\boldsymbol{x}_{j}) \right\Vert_{2}}{\left\Vert \boldsymbol{x}_{i} - \boldsymbol{x}_{j} \right\Vert_{2}}$, where $\mathcal{N}(\boldsymbol{x}_{i})$ is a ball of radius $\epsilon$ centered at the anchor point $\boldsymbol{x}_{i}$. Therefore, in the proposed SX-GeoTree model, the last module considers an additional penalty $\mathcal{L}_{SHAP}$ with regard to the robustness of continuous explanations on the model's objective. More importantly, under the context of spatial similarity assessment, the numerator and denominator of the local Lipschitz function can be replaced by more flexible metrics of dissimilarity \citep{Mulekar2017}. Here, we argue that, instead of explicitly containing similar instances within the small neighborhood $\mathcal{N}(\boldsymbol{x}_{i})$ of the high dimensional predictor space, two instances (with locational coordinates) can be seen as similar if they are close to each other in the geographical space or if the dependent and explanatory variables at the two locations follow a same predictive relationship (a.k.a., stimulus-response relationships), as Tobler's First Law of Geography \citep{Tobler1970} implies. Usually, closer locations are likely to adhere to similar stimulus-response relationships. Therefore, we utilize the geographically weighted regression (GWR) analysis that effectively applies this law in spatial analyses \citep{Fotheringham2009} to uncover the spatial similarity between different instances.

\begin{figure}[h!] 
   \centering
   \includegraphics[width=\textwidth]{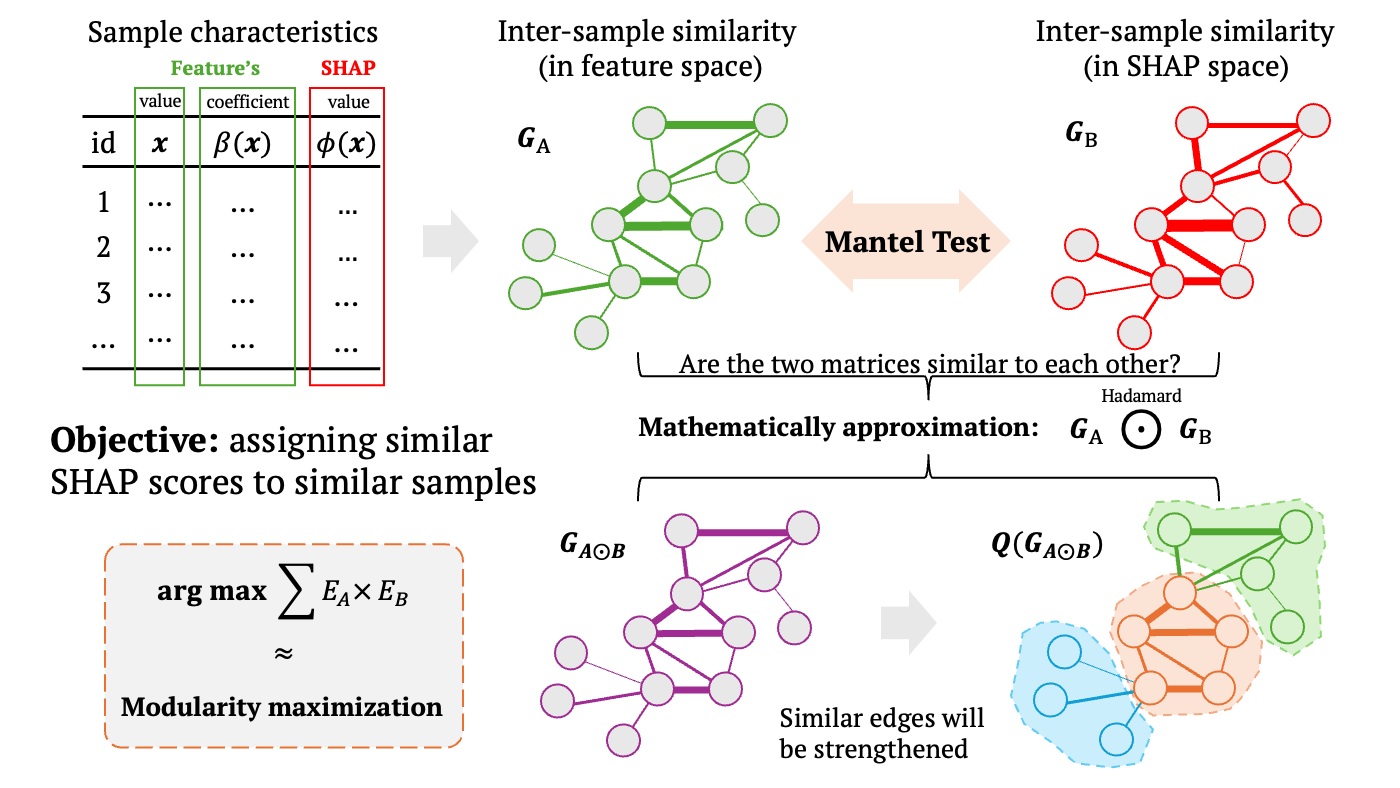} 
   \caption{The construction and partition of the spatial similarity network}
   \label{fig4:modularization}
\end{figure}

According to the GWR model, we first inspect stimulus-response relationships and if/how that relationship varies in space: $y_{i} = \beta_{i,0} + \beta_{i,1} x_{i,1} + \beta_{i,2} x_{i,2} + ... + \beta_{n,1} x_{n,1} + \varepsilon_{i} $, where $i$ refers to a location at which data on $y$ and $x$ are measured and at which local estimates of the parameters $\boldsymbol{\beta}_{i} = \{\beta_{i,0}, \beta_{i,1}, ..., \beta_{i,n}\}$ are obtained. Then, we define spatial dissimilarity of two instances $i$ and $j$ using the Euclidean distance between their GWR coefficients and update the local Lipschitz function as $\hat{L}(\boldsymbol{x}_{i}) =  \mathop{\arg\max}\limits_{\boldsymbol{\beta}_{j} \in \mathcal{S}_{\epsilon}(\boldsymbol{x}_{i})} \frac{\left\Vert f_{shap}(\boldsymbol{x}_{i}) / \boldsymbol{x}_{i}  - f_{shap}(\boldsymbol{x}_{j}) / \boldsymbol{x}_{j} \right\Vert_{2}}{\left\Vert \boldsymbol{\beta}_{i} - \boldsymbol{\beta}_{j} \right\Vert_{2}}$, where $\mathcal{S}(\boldsymbol{x}_{i})$ is a ball of radius $\epsilon$ centered at the anchor point $\boldsymbol{x}_{i}$ in the coefficient space. Specifically, $f_{shap}(\boldsymbol{x}_{i}) / \boldsymbol{x}_{i}$ quantifies the normalized SHAP values for per unit feature value in order to agree with the distance between the coefficient $\boldsymbol{\beta}_{i}$, which tells the increase of model prediction along with the increase of per unit feature value. Unfortunately, the updated Lipschitz function depends on the radius $\epsilon$ that defines the spatial neighborhood of interest. To avoid heavy evaluation of the model's sensibility over the radius $\epsilon$, we instead reframe the Lipschitz function as a spatial similarity network (see Figure \ref{fig4:modularization} for example). According to the local Lipschitz criterion, if two instance $\boldsymbol{x}_{i}$ and $\boldsymbol{x}_{j}$ is separated by a distance $d(\boldsymbol{\beta}_{i}, \boldsymbol{\beta}_{j})$ in the GWR coefficient space, then the ideal distance between the model explanations for the two instances $d(f_{shap}(\boldsymbol{x}_{i})/\boldsymbol{x}_{i}, f_{shap}(\boldsymbol{x}_{j})/\boldsymbol{x}_{j})$ should be $kd(\boldsymbol{\beta}_{i}, \boldsymbol{\beta}_{j})$ where $k$ is a small non-negative constant. 

As the distances for every pair of instances are measured, we obtain two similarity networks: (1) Similarity network of instances in the GWR coefficient space, $G_{_{GWR}}$; (2) Similarity network of instances in the feature attribute space, $G_{_{SHAP}}$. More importantly, if the local Lipschitz criterion holds, the two similarity networks should be structurally identical (since every edge in the first network is $k$ times longer than its counterpart in the second network). Reversely, this property indicates that, the more similar the two networks are, the more likely the model's explanations hold the local Lipschitz criterion. Mathematically, borrowing from the idea of Mantel test \citep{Mantel1967} for quantifying the agreement between two matrices, we compute the Hadamard (element-wise) product of the two matrices $G_{_{GWR}} \odot G_{_{SHAP}}$ and obtain a new similarity matrix $G$, in which the similarity score between two instances will be reinforced to either more similar or more dissimilar if their similarities in $G_{_{GWR}}$ and $G_{_{SHAP}}$ is close to each other, or will be shifted towards neutral otherwise. Such a consensus between the model's inputs and explanations can be uncovered by the modularity of network $G$ that compares the density of links inside communities with the density of links between communities, relative to what would be expected in a random network with the same degree sequence. To put it straightforwardly, if the model's explanations are robust, the modularity of the new similarity network $G$ will be maximized. Therefore, in the module with regard to the robustness of explanations, the modularity \citep{Newman2006} of the similarity network $G$ that implies whether two instances closely located in either the feature value space or the GWR coefficient space are as well close to each other in the feature attribution space is optimized as an additional objective:
\begin{align} 
   \mathcal{L}_{Modularity} = \frac{1}{2m} \sum_{i,j}(A_{ij}-\gamma \frac{k_{i}k_{j}}{2m}) \delta(c_{i}, c_{j})
\end{align}
where $m$ is the sum of all edge weights in the graph, $k_{i}$ is the weighted degree of the $i$-th node, $A_{ij}$ is the adjacency matrix of $G$, $\gamma$ is the resolution parameter, and $\delta{c_{i}, c_{j}}$ is 1 if $i$ and $j$ are in the same community else 0. This module forces the GeoTree model to output robust feature attributes via fast greedy modularity maximization \citep{Blondel2008}. 

Finally, in this work we adhere to the weighted regularization approach rather than the bilevel optimization approach (please refer to Section 2.3 for a detailed comparison between them) to achieve multi-objectives learning in that: (1) The proposed model is a \emph{proof of concept} and we believe that adding regularization terms to the conventional decision tree model is more intuitive and understandable; (2) The bilevel optimization is only applicable to the soft decision tree model which is more complex and requires heavy parameter tuning for training the neural network. Due to this fact, we intend to leave the bilevel optimization approach on deep neural decision trees for future works. Following the regularization approach, we define a tailored information gain function for each candidate splitting node as:
\begin{align} 
   \mathcal{L}_{Gain} = (1 - | \mathcal{L}_{MoranI} | ) \times \mathcal{L}_{Modularity} \times \mathcal{L}_{MSE}
\end{align}
where $\mathcal{L}_{MoranI} $ and $\mathcal{L}_{Modularity}$ are both in the range $[-1, 1]$. We exclude the learnable parameter for each regularization term in order to ease computational load.

\subsection{Performance Metrics for Model Comparisons}

We utilize several well-established evaluation metrics for assessing the performances of the proposed SX-GeoTree models and the baseline models (including the generic regression tree (DT) and the generic GeoTree). As show in Table 2, we first compare those models' performance based on regression evaluation metrics that directly relate to the tailored information gain controlling the decision tree building process: 

(1) The \emph{coefficient of determination}, also known as \emph{R-squared}, which is defined as the fraction of shared variance between two variables, measuring the variability of one variable that is associated with the variability of the other. It is the square of the correlation coefficient and ranges from 0 to 1, where a value of 1 indicates a perfect fit; 

(2) The \emph{root mean square error} (RMSE), which is the square root of MSE (being an integral part of the tailored information gain) and in same units as response variable. In other words, it measures the average magnitude of prediction errors by calculating the square root of the average of squared differences between predicted and actual values; 

(3) The global \emph{Moran's I} index of the model's residuals, which is the tendency for nearby locations to exhibit similar prediction errors. It assesses whether the observed spatial pattern of the residuals between the predications and the observations is clustered, dispersed, or random. A Moran's I value close to +1 indicates a clustered pattern, -1 indicates a dispersed pattern, and 0 indicates a random pattern; 

(4) The \emph{modularity} of the inter-sample similarity network, which measure the strength of division of a network into communities (modules). It quantifies how well a particular partition of nodes into communities captures the community structure of the network. 

Note that these metrics not only help in measuring how well the SX-GeoTree model is able to predict continuous outcomes, but also depict whether the trade-off between accuracy and interpretability using the weighted regularization method ensures Pareto-optimal solutions. In other words, these metrics enable us to check whether improvements in the secondary objective come at an outsized expense of the primary objective, or vice versa. 

\begin{table}[htbp]
   \centering
      \caption{Evaluation metrics for model comparison.}
      \begin{tabular}{@{} lll @{}} 
         \toprule
         \textbf{Metric}    & \textbf{Description} & \textbf{Notes} \\
         \midrule
         R-squared      & Proportion of variance explained by the model. & Higher is better. \\
         RMSE       & Square root of MSE, in same units as response variable. & Lower is better. \\
         Moran's I      & Measures spatial autocorrelation of residuals. & Closer to 0 is better. \\
         Modularity & Measures the strength of community structure in a network. & Higher is better. \\
         \bottomrule
      \end{tabular}
      \label{tab:evaluation_metrics}
\end{table}

In addition, we apply five statistical indicators of the dispersion of model explanations in each community of the underlying similar network (as listed in Table 3) to measure the robustness of explanations derived from different models: 

(1) The \emph{range} (RG), which is a simple measure of variability that represents the difference between the largest (maximum) and smallest (minimum) values in feature contributions. It provides a quick indication of the spread of the data, showing how far apart the most extreme values are; 

(2) The \emph{interquartile range} (IQR), which is a measure of statistical dispersion, specifically representing the range of the middle 50\% of a dataset. It is calculated as the difference between the third quartile (Q3) and the first quartile (Q1) of a dataset; 

(3) The \emph{coefficient of variation} (CV), which is a measure of relative variability, expressing the standard deviation as a percentage of the mean. It's a unit-less measure, allowing for comparison of variability between datasets with different units or scales. A higher CV indicates greater dispersion of data points around the mean, suggesting higher relative variability; 

(4) The \emph{Shannon's entropy} (EPY), which is a measure of the uncertainty or randomness associated with the average contribution of features. It quantifies the skewness of the attributions assigned to all the features. In our case, we prefer to assign high SHAP scores to a limited number of features; 

(5) The \emph{Gini coefficient} (GC), which is a statistical measure of inequality in a distribution. In economics, it's commonly used to measure income inequality within a population, but it can be applied to other distributions as well. The coefficient ranges from 0 to 1 (or 0\% to 100\%), where 0 represents perfect equality (everyone has the same share) and 1 (or 100\%) represents perfect inequality (one person has all the share). 

By utilizing these indicators of dispersion, we evaluate whether the proposed SX-GeoTree model is able to attribute features in a more robust manner. That is, within each spatial community derived from the underlying spatial similar network, the dispersion of feature attributions should be narrow.  

\begin{table}[htbp]
   \centering
   \caption{Dispersion metrics for evaluating intra-community stability of SHAP feature attributions. Lower values indicate more robust (stable) feature attributions.}
   \begin{tabular}{@{} l p{6.8cm} l @{}} 
      \toprule
      \textbf{Metric} & \textbf{Description} & \textbf{Notes} \\
      \midrule
      Range (RG) & Max minus min SHAP value within a community. & Lower is better. \\
      Interquartile Range (IQR) & Middle 50\% spread of SHAP values (Q3–Q1). & Lower is better. \\
      Coefficient of Variation (CV) & Standard deviation divided by mean absolute SHAP value. & Lower is better. \\
      Shannon's Entropy (EPY) & Distributional uncertainty of normalized absolute SHAP values. & Lower is better. \\
      Gini Coefficient (GC) & Inequality of absolute SHAP mass across features. & Lower is better. \\
      \bottomrule
   \end{tabular}
   \label{tab:dispersion_metrics}
\end{table}

Besides, we further extend the indicators listed in Table \ref{tab:dispersion_metrics} as both globally- and locally-averaged, which enable us to assess the robustness of model explanations more comprehensively. We also conduct an ablation study to evaluate the effectiveness of the Moran's I module and the network modularity maximization module.

\section{Experiments and Results}

\subsection{Datasets}

We utilize two representative datasets to evaluate the performance of the proposed SX-GeoTree model. The first dataset consists of the Year-2022 economic statistics of 83 counties in Fujian Province sitting at the Southeastern China. In total, 29 economic and environmental characteristics are recorded in the raw data and we include 8 representative variables informed by the multicollinearity analysis (see Supplementary Material A.1); that is, the VIFs of these selected indicators are below 5. The second dataset records home sales prices and characteristics of 21,613 properties for Seattle and King County, WA in United States from May 2014 to May 2015. This dataset has been widely modeled in existing literature. Similarly, a total of 21 variables are recorded in the raw data and we include 8 representative features with a VIF value below 5 (see Supplementary Material A.2), following previous works. Based on the selected variables (as listed in Table \ref{tab:data}), we build SX-GeoTree models to predict the county-level GDP in Fujian and the house price in Seattle using regression and k-fold cross-validation. During the modeling, all the features included are standardized based on the Z-score method. 

\begin{table}[htbp]
   \centering
   \caption{Datasets and variables used in the experiments.}
   \begin{tabular}{@{} l p{7cm} l @{}} 
      \toprule
      \textbf{Dataset}    & \textbf{Independent Variables} & \textbf{Dependent Variable}\\
      \midrule
       \multirow{7}{*}{Fujian-county}      & \multirow{6}{7cm}{mercator\_x (``X''); mercator\_y (``Y''); primary sector (\%, ``PRI''); tertiary sector (\%, ``TER''); industrial trade profitability (\textyen, ``IND''); household registration population (\#, ``POP''); cultivated land (\%, ``CLD''); impervious surfaces (\%, ``IMP'')} & \multirow{7}{*}{GDP (\textyen)} \\
       & &  \\
       & &  \\
              & &  \\
                     & &  \\
                     & &  \\ 
     & &  \\                     
     \cmidrule(r){1-1}
      \multirow{6}{*}{Seattle-house}       & \multirow{6}{7cm}{mercator\_x (``X''); mercator\_y (``Y''); number of bathrooms (\#, ``BTH''); size of living area (sqft, ``LIV''); size of the lot (sqft, ``LOT''); construction quality (grade, ``GRA''); condition of the house (rank, ``CON''); year built (age, ``AGE'')}  & \multirow{6}{*}{Sale Price ($\log$(\$))} \\
       & &  \\
       & &  \\   
       & &  \\
       & &  \\             
       & &  \\     
      \bottomrule
   \end{tabular}
   \label{tab:data}
\end{table}

\subsection{Model Performance}

We compared the performances of the decision tree regressor (DT), the GeoTree model (GT), and our proposed SX-GeoTree model (using either $\ell_{2}(feature) = \left\Vert \boldsymbol{x}_{i} - \boldsymbol{x}_{j} \right\Vert_{2}$ or $\ell_{2}(gwr)=\left\Vert \boldsymbol{\beta}_{i} - \boldsymbol{\beta}_{j} \right\Vert_{2}$ similarity) for the two predictive tasks. Recall that multi-task learning often suffers from competing objectives. Therefore, we first analyze the prediction accuracy of each model in order to evaluate whether the predictive power is compromised when incorporating the robustness of model interpretability. Then, we analyze the dispersion of SHAP values for similar samples across space to verify whether the proposed model assigns more similar feature attributions to similar samples. Lastly, results of the ablation analysis are reported to confirm the validity of the proposed modules for enhancing the robustness of model interpretability.

\subsubsection{Accuracy}

We determine the optimal parameters, i.e. minimum size leaf (``msl'') and maximum depth (``md''), for the DT model, the GT model, and our proposed SX-GeoTree model using the 5-fold cross-validation approach. The parameters and results are listed in Table \ref{tab5:fujian} for the Fujian-county data and in Table \ref{tab6:seattle} for the Seattle-house data, respectively. 

\begin{table}[h!]
   \centering
      \caption{Model performance on the Fujian-county dataset.}
   \begin{tabular}{@{} llllll@{}} 
      \toprule
      & \textbf{DT} & \textbf{GT} & \multicolumn{2}{c}{\textbf{SX-GeoTree} (ours)} \\
      \cmidrule(r){4-5} 
      Metrics ({\scriptsize{similarity}})    &  &  & \scriptsize{$\ell_{2}(feature)$} & \scriptsize{$\ell_{2}(gwr)$}  \\
      \cmidrule(r){2-5}       
       Parameters ({\scriptsize{optimal}}) & \scriptsize{msl=5,md=5} & \scriptsize{msl=5,md=4} & \scriptsize{msl=5,md=5} & \scriptsize{msl=5,md=5} \\
      \midrule
      \textbf{R}$^{2}$ ({\scriptsize{train}})      & 0.68 		& \textbf{0.73} 	& 0.70 	& 0.72\\
       \textbf{R}$^{2}$ ({\scriptsize{test}}) 	& 0.78	& \textbf{0.86}	& 0.77	& 0.77\\
      \textbf{RMSE} ({\scriptsize{train}})      & 330.71	&\textbf{298.07}	&315.95	&307.44 \\
      \textbf{RMSE} ({\scriptsize{test}})       & 255.54	&\textbf{223.12}	&270.84	&272.24 \\
      \textbf{Moran's I} ({\scriptsize{residuals}}) & -0.0310	&-0.0339	&\textbf{0.0234}	&\textbf{0.0277} \\
      \textbf{Modularity} ({\scriptsize{$\ell_{2}(feature)$}})	& 0.0794		&0.0898	&\textbf{0.1917}	&	\\
     \textbf{Modularity} ({\scriptsize{$\ell_{2}(gwr)$}})				 	& 0.0644		&0.0993	&	& \textbf{0.1679}	\\ 
      \bottomrule
   \end{tabular}
   \label{tab5:fujian}
\end{table}

\begin{table}[h!]
   \centering
   \caption{Model performance on the Seattle-house dataset.}
   \begin{tabular}{@{} lllll @{}} 
      \toprule
      & \textbf{DT} & \textbf{GT} & \multicolumn{2}{c}{\textbf{SX-GeoTree} (ours)} \\
      \cmidrule(r){4-5} 
        Metrics ({\scriptsize{similarity}})    &  &  & \scriptsize{$\ell_{2}(feature)$} & \scriptsize{$\ell_{2}(gwr)$}\\
      \cmidrule(r){2-5}       
       Parameters ({\scriptsize{optimal}}) & \scriptsize{msl=6,md=8} & \scriptsize{msl=6,md=8} & \scriptsize{msl=6,md=8} & \scriptsize{msl=6,md=8} \\
      \midrule
      \textbf{R}$^{2}$ ({\scriptsize{train}})      & \textbf{0.89} 		& \textbf{0.89} 	& 0.88 	& 0.88\\
       \textbf{R}$^{2}$ ({\scriptsize{test}}) 	& \textbf{0.78}	& \textbf{0.78}	& 0.75	& 0.77 \\
      \textbf{RMSE} ({\scriptsize{train}})      & \textbf{24.06}	& 24.49	&25.95	&25.39 \\
      \textbf{RMSE} ({\scriptsize{test}})       & 32.28	&\textbf{32.00}	&34.79	&32.51 \\
      \textbf{Moran's I} ({\scriptsize{residuals}}) & 0.1037	&0.0878	&\textbf{0.0591}	&\textbf{0.0544} \\
      \textbf{Modularity} ({\scriptsize{$\ell_{2}(feature)$}})	& 0.0500		&0.0507	&\textbf{0.1051}	&	\\
     \textbf{Modularity} ({\scriptsize{$\ell_{2}(gwr)$}})				 	& 0.0416		&0.0416	&	& \textbf{0.0765}\\ 
      \bottomrule
   \end{tabular}
   \label{tab6:seattle}
\end{table}

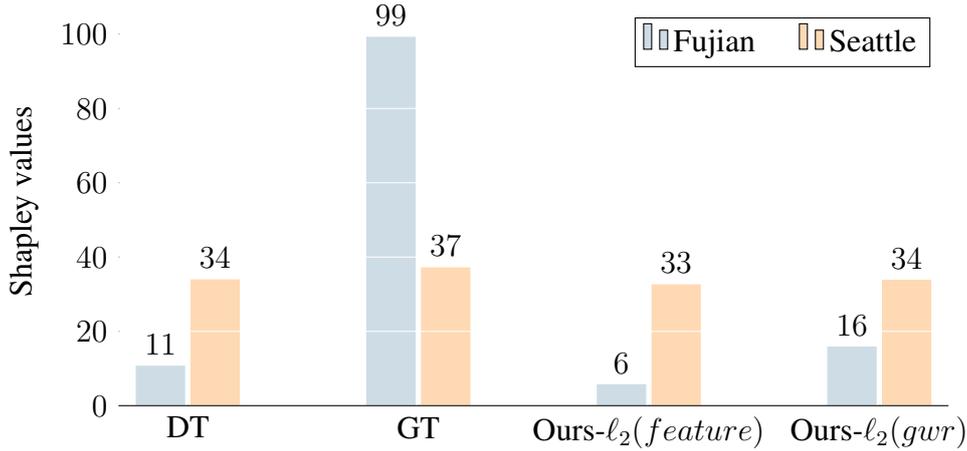
\begin{figure}[h!]
\centering
\begin{tikzpicture}
  \centering
  \begin{axis}[
        ybar, axis on top,
        height=7cm, width=12.5cm,
        bar width=0.65cm,
        ymajorgrids, tick align=inside,
        major grid style={draw=white},
        enlarge y limits={value=.1,upper},
        ymin=0, ymax=100,
        axis x line*=bottom,
        axis y line*=left,
        y axis line style={opacity=0},
        tickwidth=0pt,
        enlarge x limits=true,
        legend style={
            at={(0.8,0.95)},
            anchor=north,
            legend columns=-1,
            /tikz/every even column/.append style={column sep=0.5cm}
        },
        ylabel={Shapley values},
        symbolic x coords={
           DT, GT, Ours-$\ell_{2}(feature)$, Ours-$\ell_{2}(gwr)$},
       xtick=data,
       nodes near coords={
        \pgfmathprintnumber[precision=0]{\pgfplotspointmeta}
       }
    ]
    \addplot [draw=none, fill=airforceblue!30] coordinates {
      (DT,10.76)
      (GT, 99.29) 
      (Ours-$\ell_{2}(feature)$,5.77)
      (Ours-$\ell_{2}(gwr)$,15.92)  };
   \addplot [draw=none,fill=orange!30] coordinates {
      (DT,34.00)
      (GT, 37.22) 
      (Ours-$\ell_{2}(feature)$,32.66)
      (Ours-$\ell_{2}(gwr)$,33.89)  };
    \legend{Fujian,Seattle}
  \end{axis}
  \end{tikzpicture}
   \caption{The estimated SHAP scores for locational features}
   \label{fig5:geoshapley}
 \end{figure}

Interestingly, the improvements of accuracy gained from the GT model varies between the GDP prediction in Fujian and the house price prediction in Seattle. Due to the flexibility of learning oblique and Gaussian splits in space, GT outperforms DT and our model significantly on the first predictive task (with an increase of accuracy close to $0.1$ on the test set). Whereas, on the second predictive task, the three models perform on par with each other, which suggests that for this case study the oblique and Gaussian splits are less frequently adopted by the GT model. This fact is further confirmed by comparing the estimated SHAP scores of the locational features (i.e., the combined attribution of ``X'' and ``Y'' that are utilized for building axis-parallel, oblique and Gaussian splits) across the three models as shown in Figure \ref{fig5:geoshapley}. Moreover, for both cases, our proposed model can perform equally well with the DT model (achieving similar $R^{2}$ values). Unfortunately, under the situation that the oblique and Gaussian splits are overwhelming (as in Fujian), the incorporation of the Moran's I module and the modularity maximization module do compromise the model's predictive power as aforementioned in the multi-objective/task learning framework (see Section 2.3). Therefore, we look forward to adopting the bilevel optimization strategy and the deep neural decision tree method in order to cancel out the trade-off in our future work. 

The trade-off between the accuracy of model's prediction and the robustness of model's explanations are also manifested by the spatial unevenness of residuals and the consensus of feature attributions to similar samples. For both the GDP prediction in Fujian and the house price prediction in Seattle, our proposed models yield a lower value of Moran's I and a higher value of modularity when compared with the DT model and the GT model. As discussed above, the spatial patterns in the Fujian GDP data are more profound than that in the Seattle house price data. Therefore, the modularity of the similarity network in Fujian (around $0.2$) is higher than the modularity of the similarity network in Seattle (around $0.1$). More importantly, our proposed model achieves more robust explanations than the DT model, while does not undermine the prediction accuracy. Particularly, for the Seattle case study, despite that the predictive accuracies of DT, GT and our proposed model are on par, our model performs significantly better in terms the disparity of residuals (about $50\%$ lower) and the consensus of explanations (about $2\times$ higher). This ability showcases the effectiveness of the proposed modules.    

\subsubsection{Robustness}

To further verify whether our proposed model succeeds in assigning similar SHAP scores to similar samples, we visualize the community partitions (labeled as different "Type") derived from the modularity maximization module. In addition, for each spatial community, we inspect the spatial cohesion and dispersion of RG, IQR, CV, EPY and GC metrics, which are ``the lower, the better'' as aforementioned.  For the sake of concise, the detailed statistics are reported in the Supplementary Materials A.3 and A.4. 

\begin{figure}[h!]
\captionsetup[subfigure]{justification=centering}
	\centering
	\begin{subfigure}{0.32\textwidth}
	\centering
	\includegraphics[width=\textwidth]{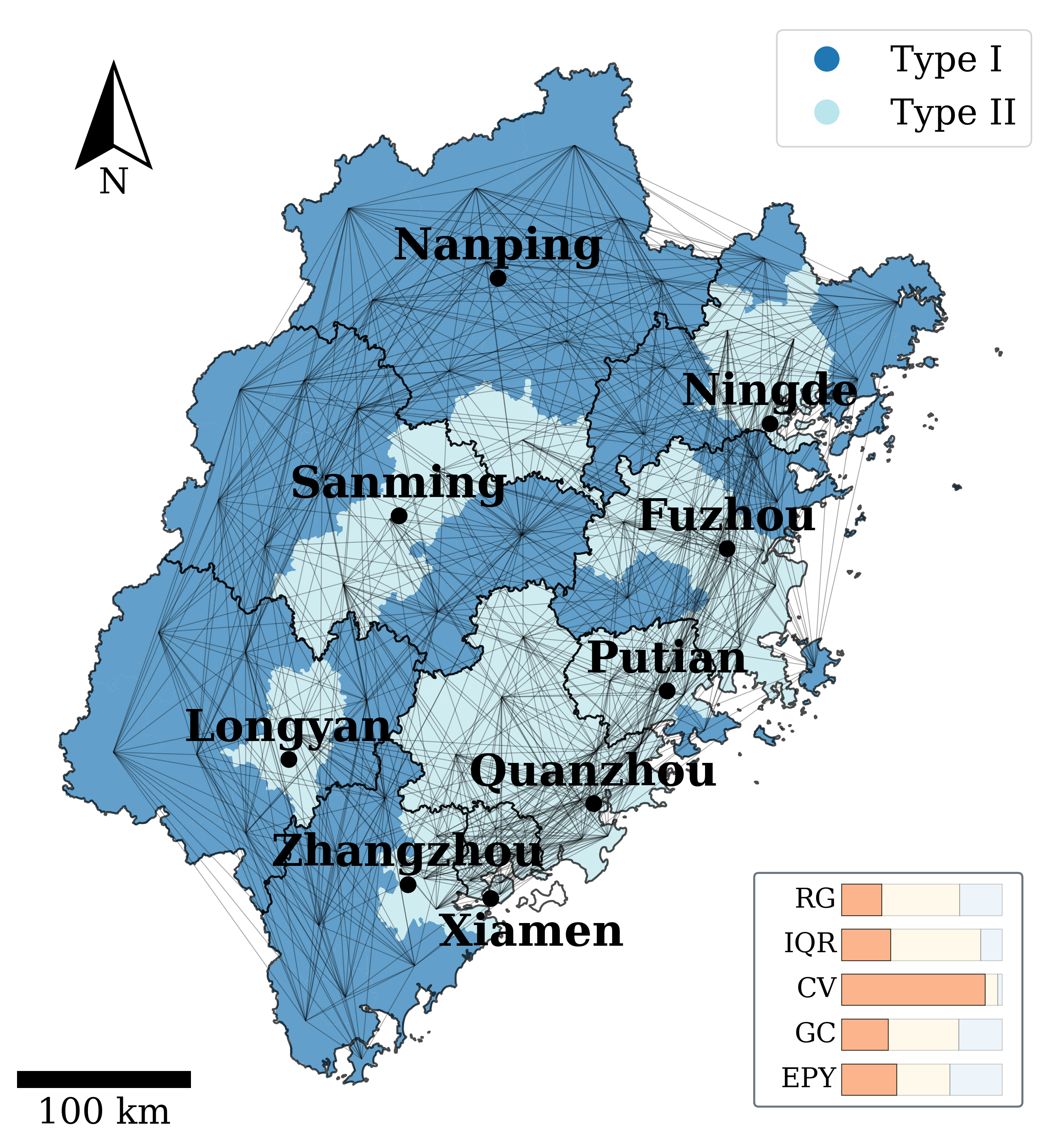}
	 \scriptsize Type I\\
	\includegraphics[width=\textwidth]{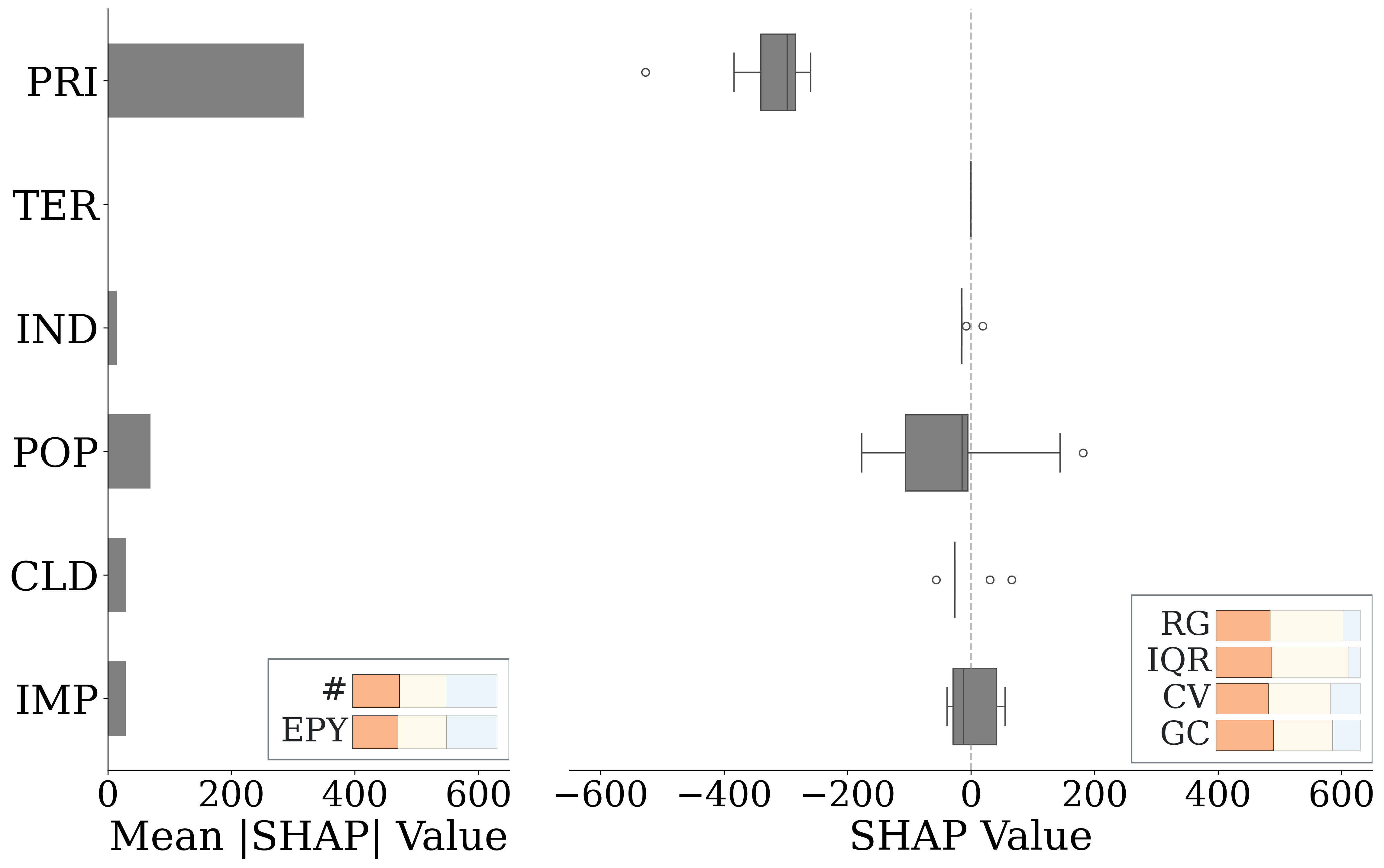}\\
	Type II\\	
	\includegraphics[width=\textwidth]{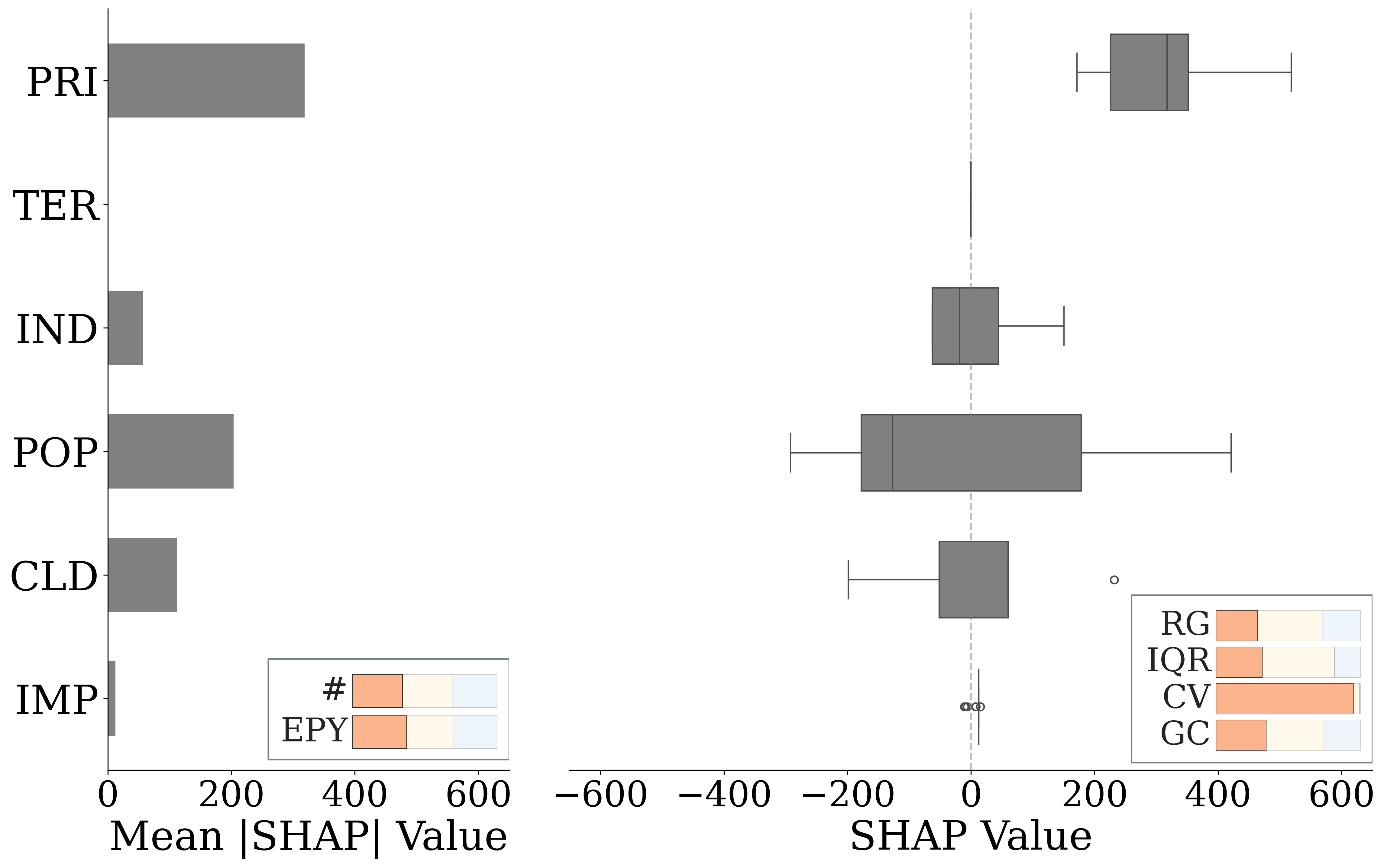}
	 \subcaption{DT\\(msl=5, md=5, $\ell_{2}(feature)$)}
	 \end{subfigure}
	 \begin{subfigure}{0.32\textwidth}
	\centering
	\includegraphics[width=\textwidth]{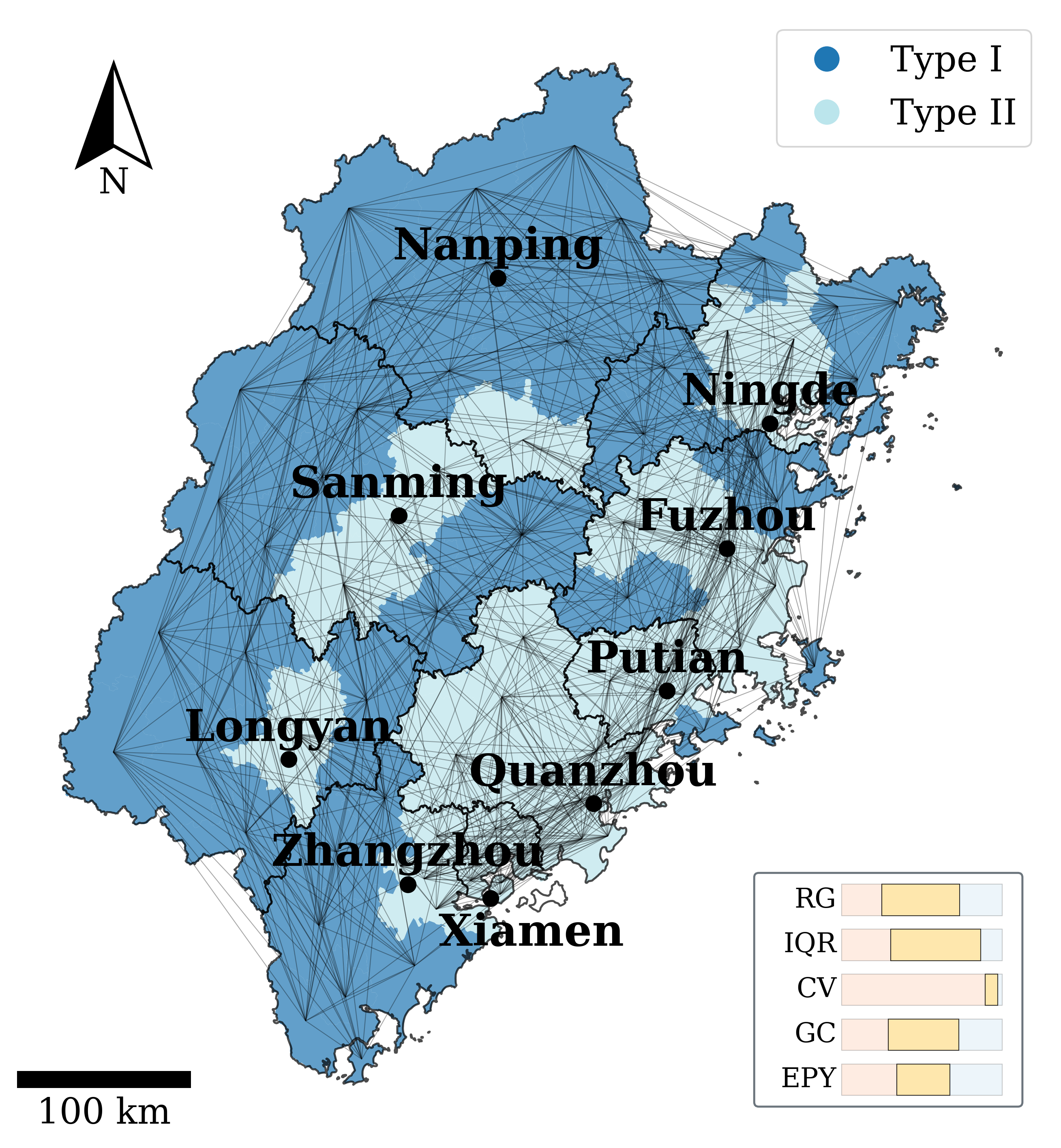}
	 \scriptsize Type I\\
	\includegraphics[width=\textwidth]{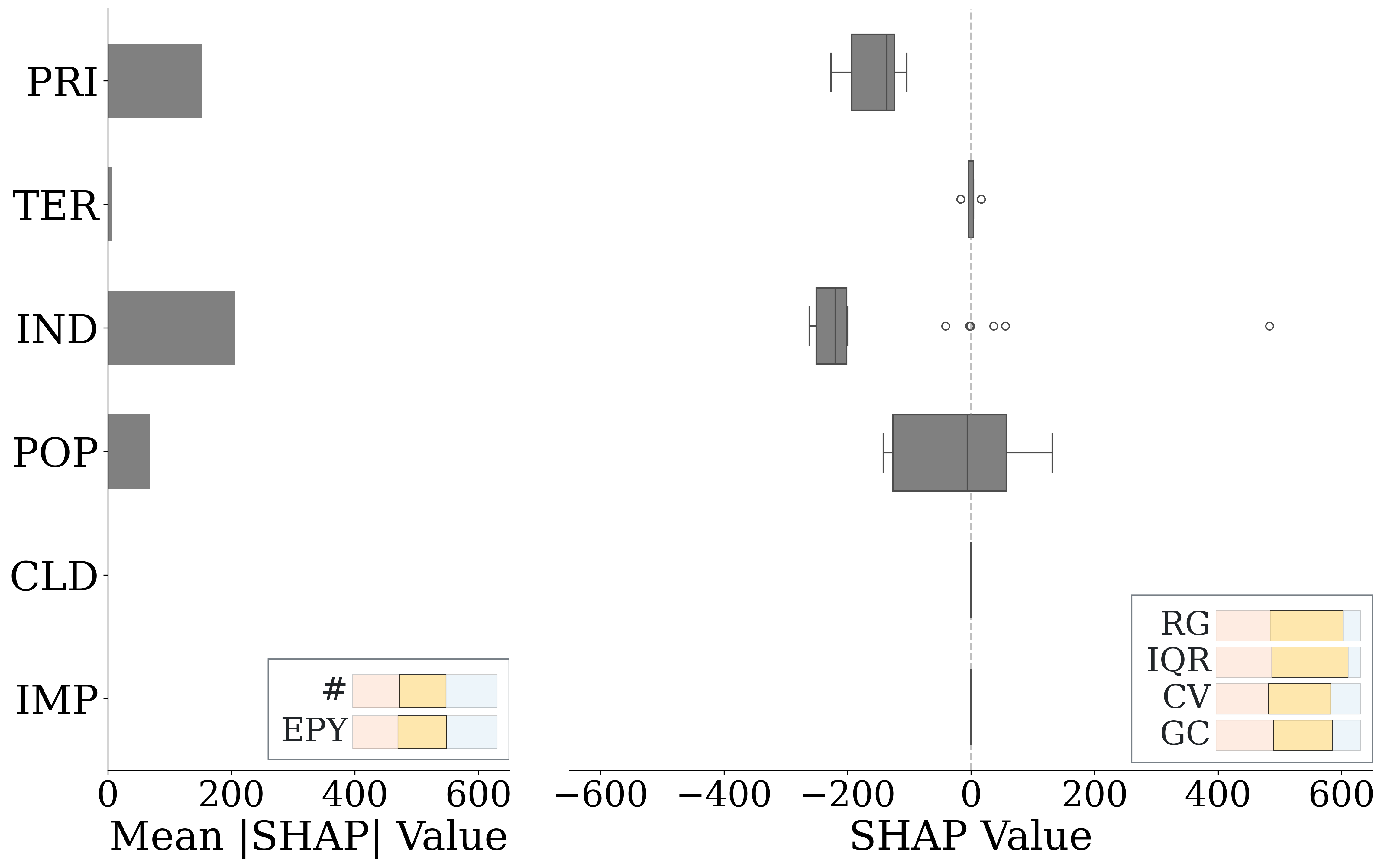}\\
	Type II\\
	\includegraphics[width=\textwidth]{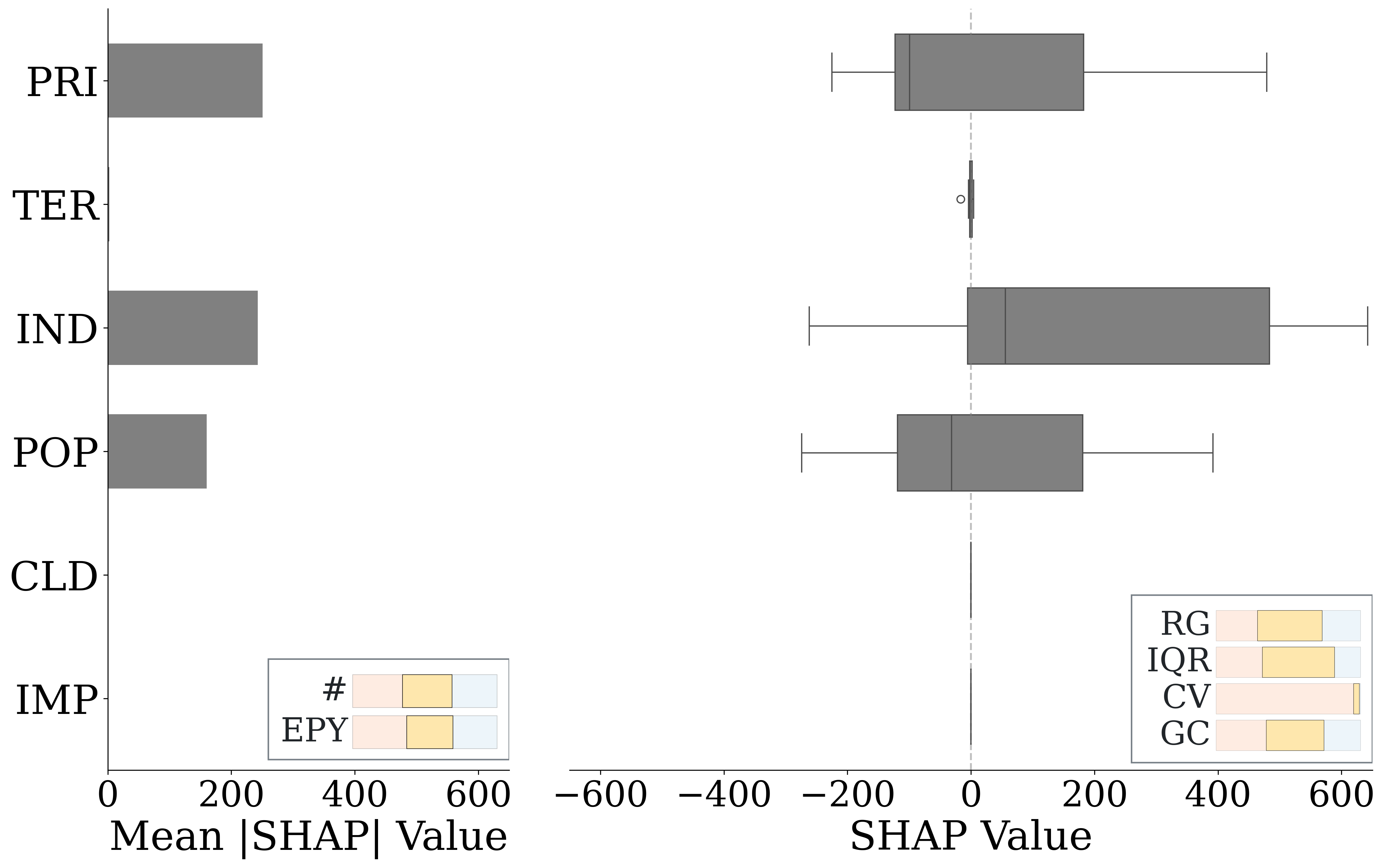}
	 \subcaption{GT\\(msl=5, md=4, $\ell_{2}(feature)$)}
	  \end{subfigure}
	\begin{subfigure}{0.32\textwidth}
	\centering
	\includegraphics[width=\textwidth]{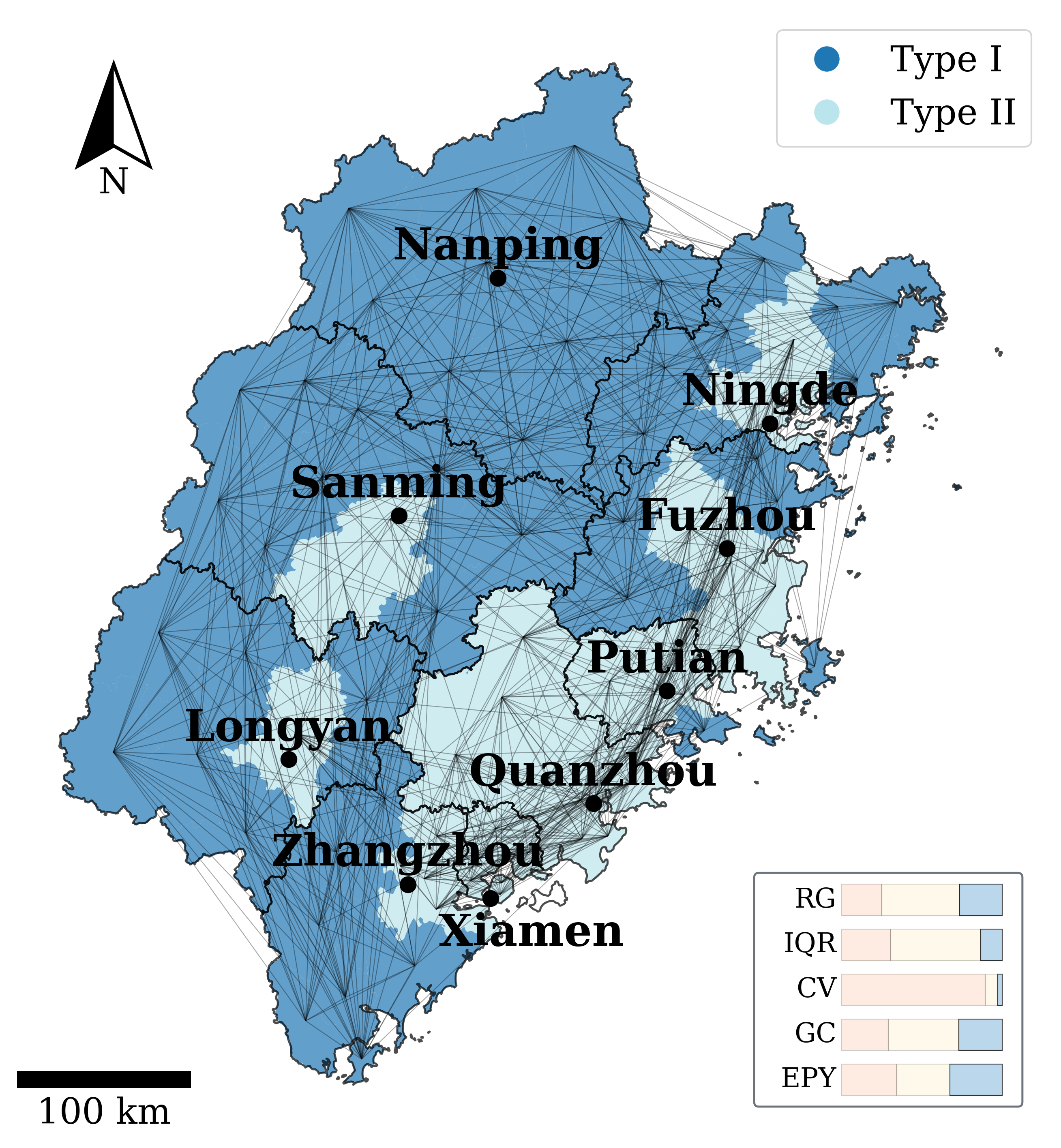}
	 \scriptsize Type I\\
	\includegraphics[width=\textwidth]{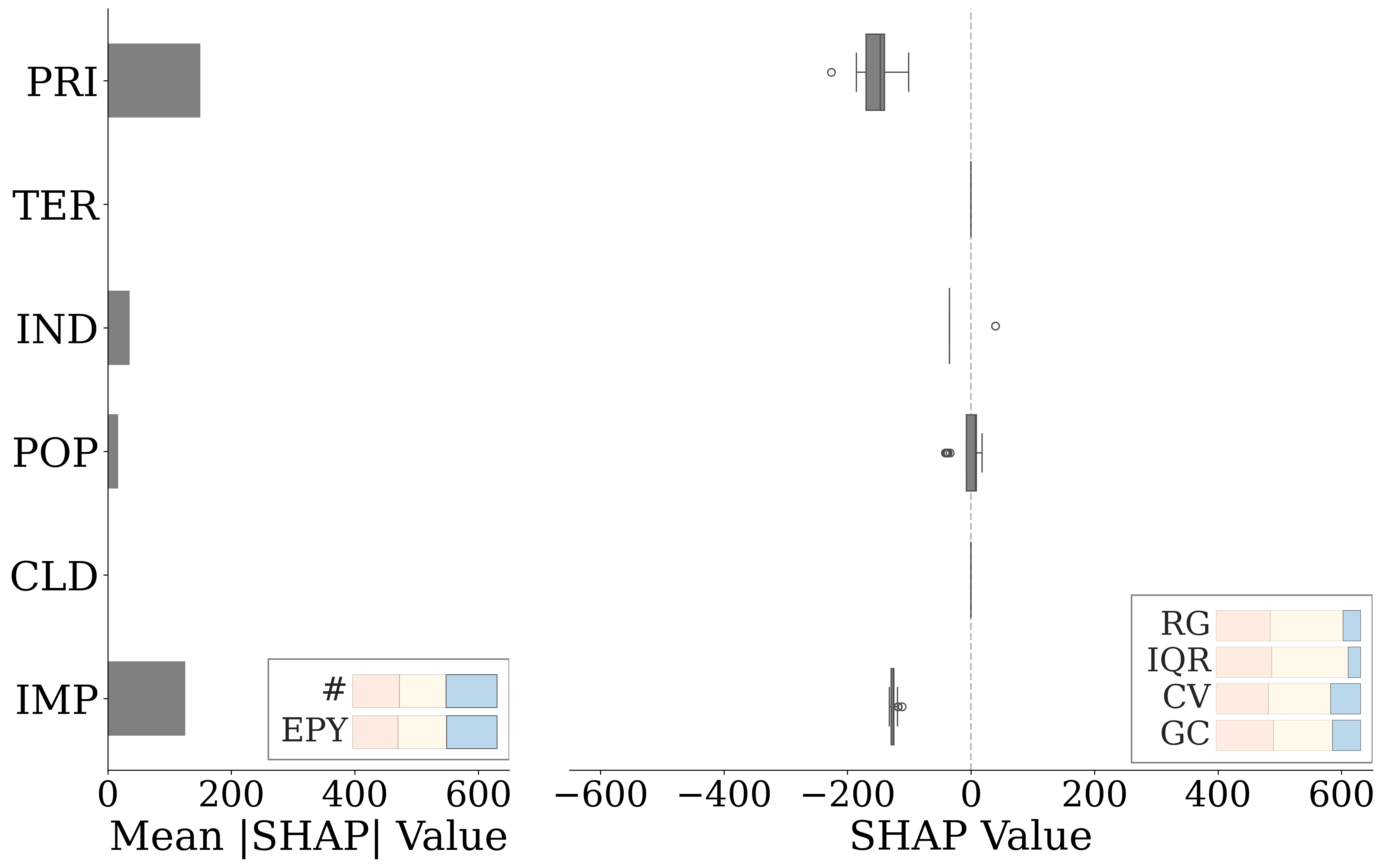}\\
	Type II\\
	\includegraphics[width=\textwidth]{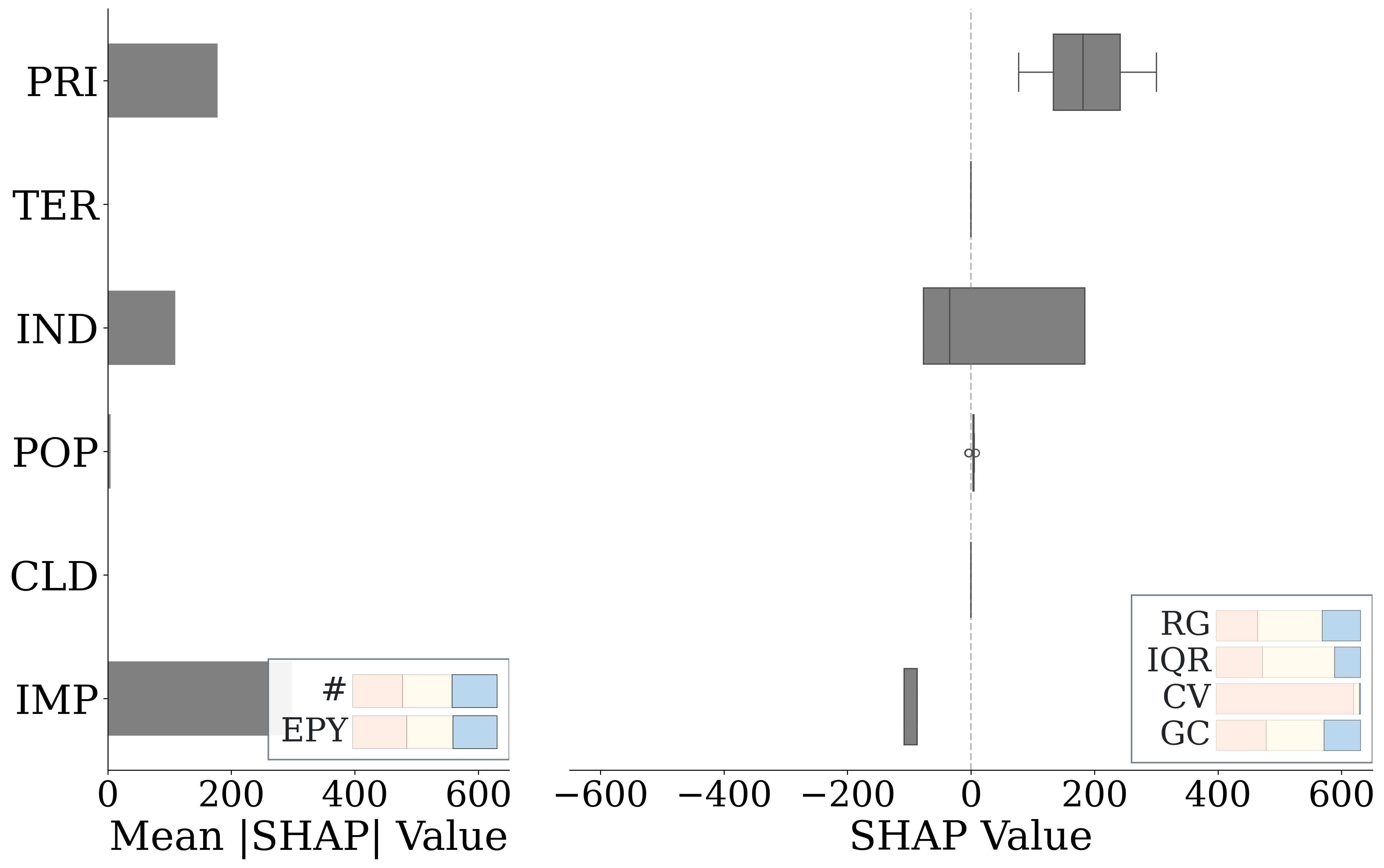}
	 \subcaption{Ours\\(msl=5, md=5, $\ell_{2}(feature)$)}
	 \end{subfigure}
	\caption{The derived spatial communities of counties with similar feature values and attributions in Fujian. The SHAP scores for each feature in each community are compared across the three models based on the predefined dispersion indicators.}
	\label{fig6:fujian-feature}
\end{figure}

As shown in Figure \ref{fig6:fujian-feature}, all the three models predominately separate the municipal centers (counties with dot marker) from the periphery counties in Fujian. Due to the imbalanced development between administrative centers and periphery areas, the socio-economic indicators of ``Type II'' counties (in light blue) are often much higher in magnitude than the indicators of ``Type I'' counties (in dark blue). Considering that the models account for spatial similarity between samples based on their input feature values (i.e., $\ell_{2}(feature)$), they naturally separate the counties with distinctive magnitudes of socio-economic statuses. In specific, in the ``Type II'' areas, the feature importances of IND, POP and IMP are generally higher than in the ``Type I'' areas. Intuitively, in terms of the spatial dispersions of feature attributions to counties within the same spatial community, our proposed model (bars in light blue) performs better than the DT model (bars in light red) and the GT model (bars in light orange). This finding confirms the validity of approximating Mantel test based on the modularity maximization of the element-wise product between the similarity network in feature space and the similarity network in SHAP space. Additionally, the ``Type II'' areas are not spatially cohesive due to the pre-established configuration of administrative centers as the central place theory implies.  

\begin{figure}[h!]
\captionsetup[subfigure]{justification=centering}
	\centering
	\begin{subfigure}{0.32\textwidth}
	\centering
	\includegraphics[width=\textwidth]{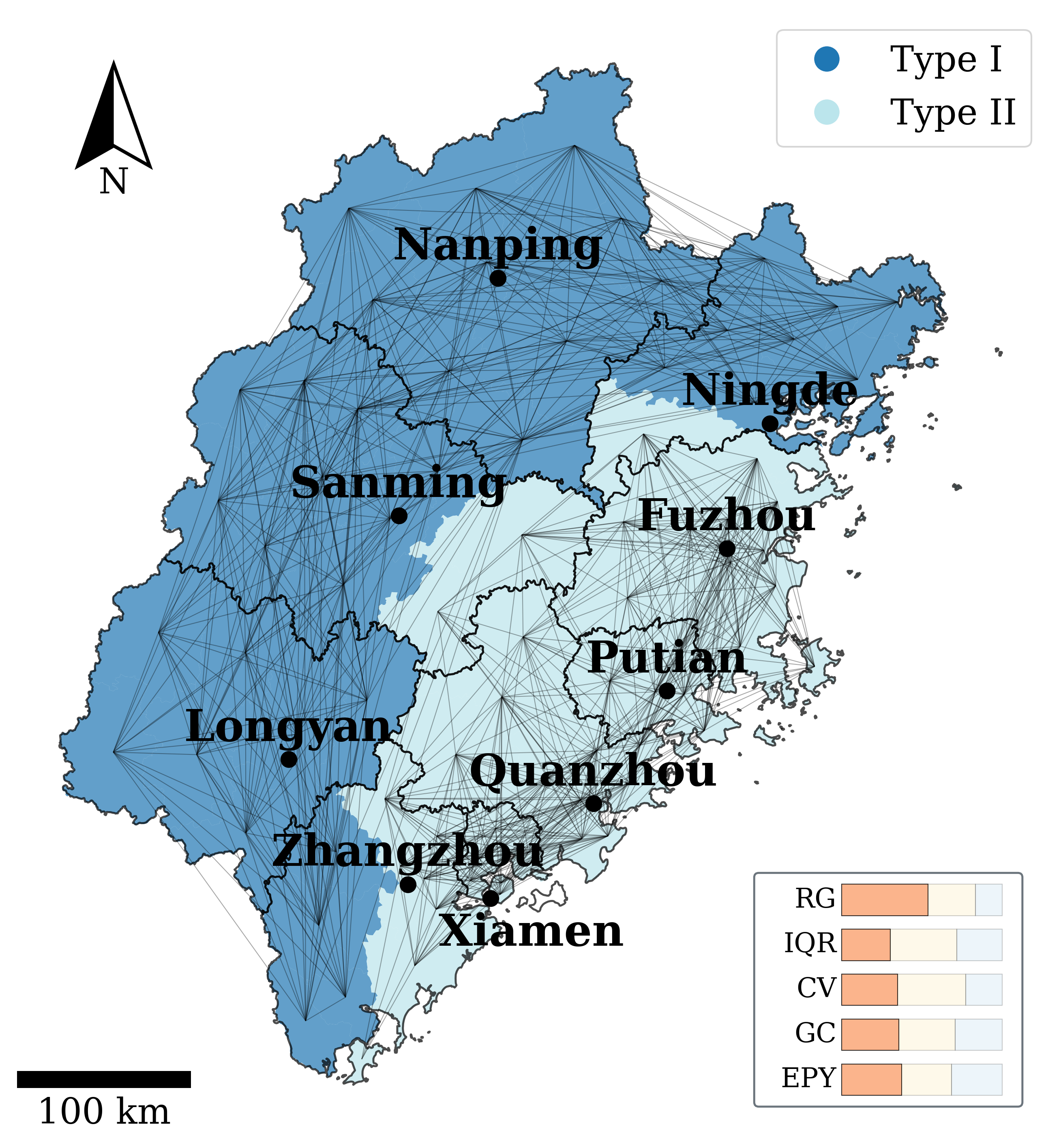}
	 \scriptsize Type I\\
	\includegraphics[width=\textwidth]{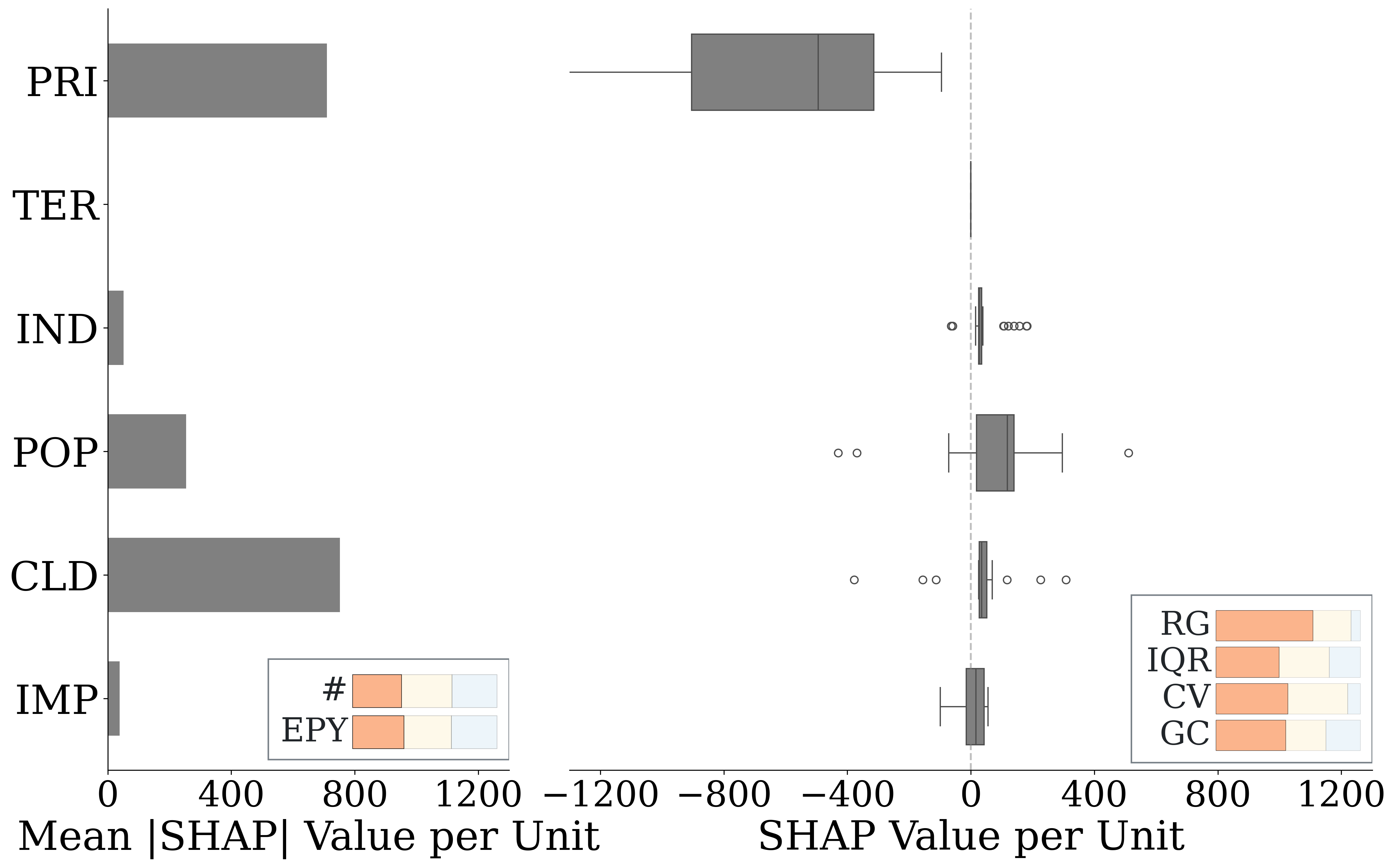}\\
	Type II\\
	\includegraphics[width=\textwidth]{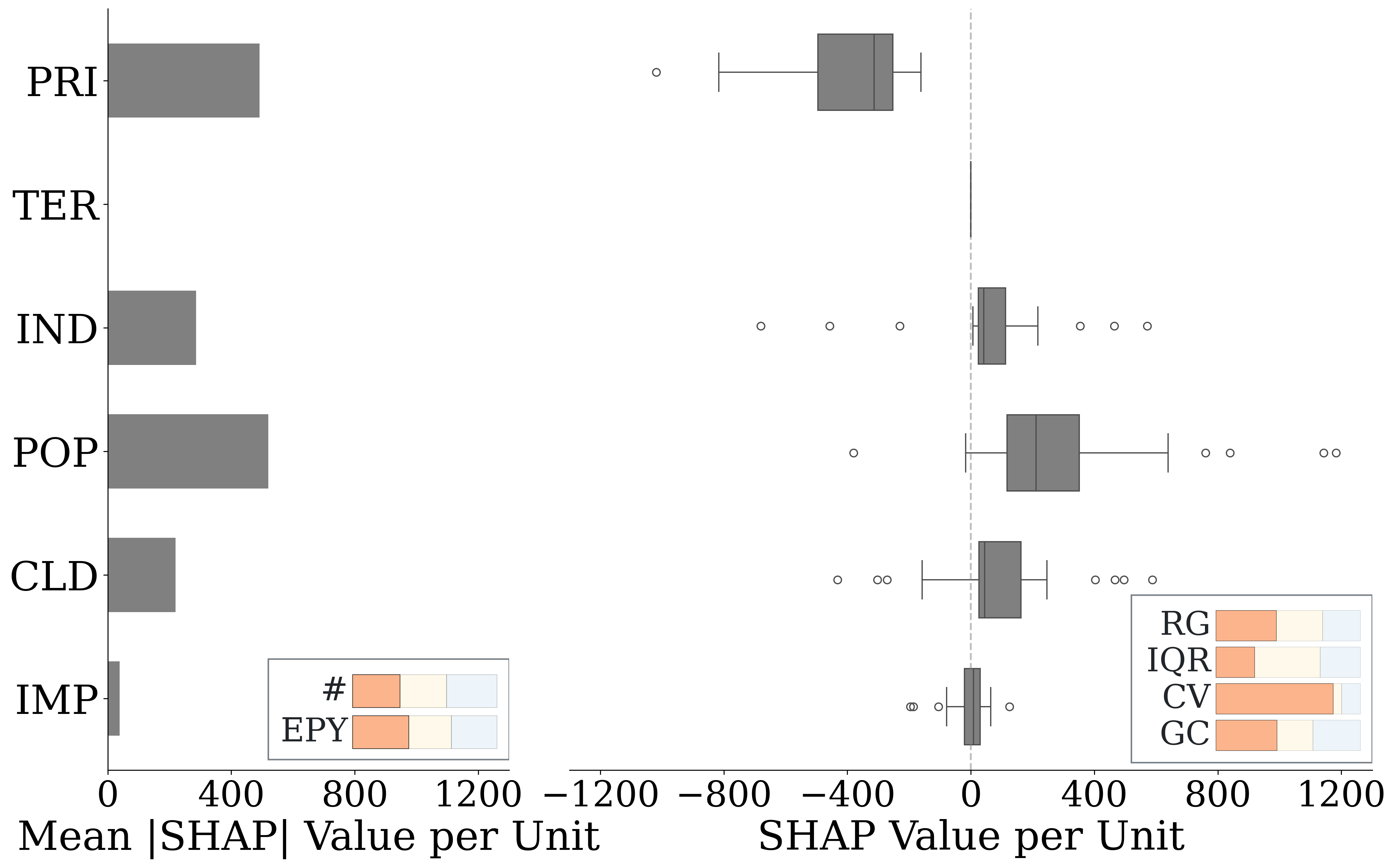}
	 \subcaption{DT\\(msl=5, md=5, $\ell_{2}(gwr)$)}
	 \end{subfigure}
	 \begin{subfigure}{0.32\textwidth}
	\centering
	\includegraphics[width=\textwidth]{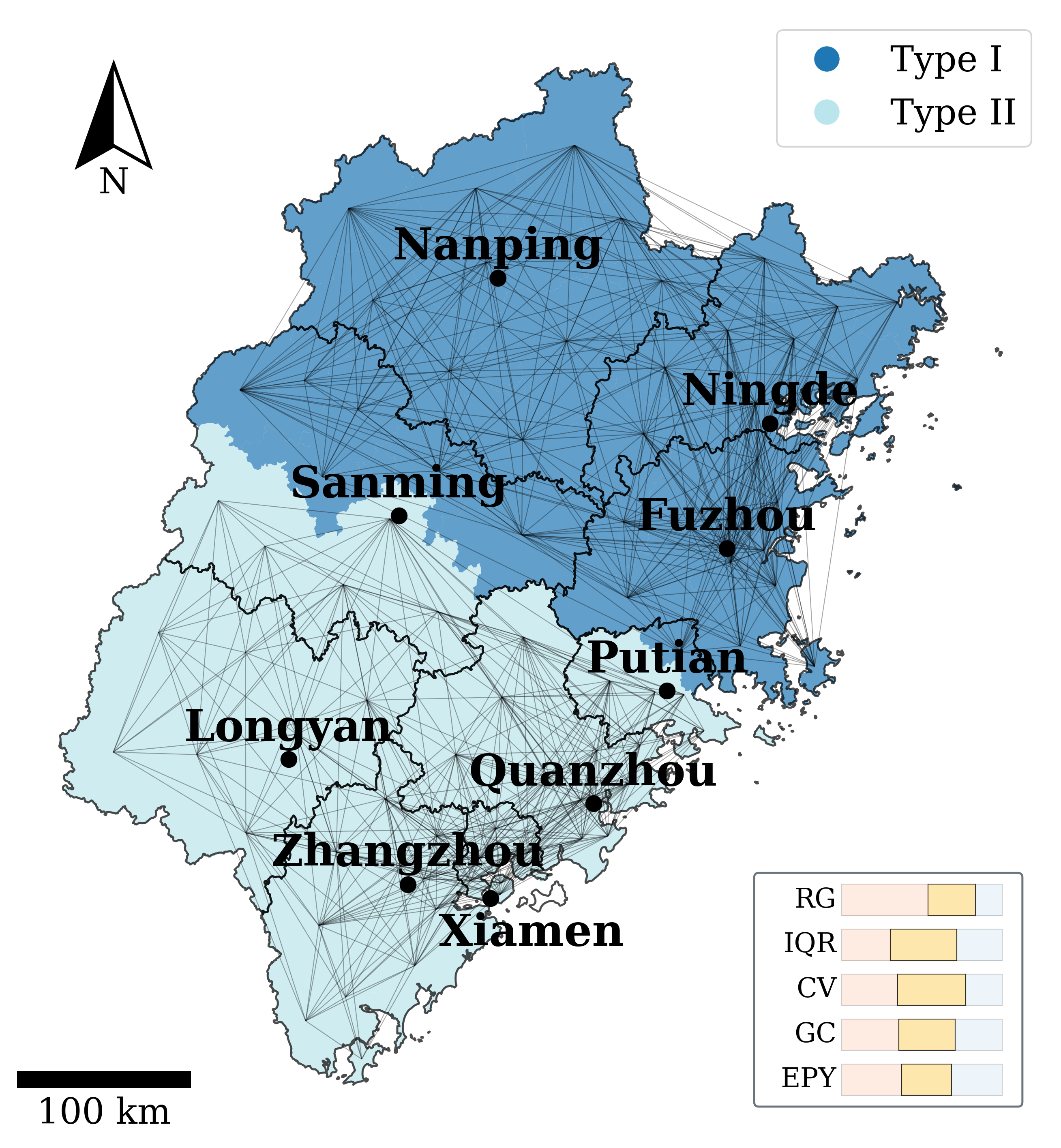}
	\scriptsize Type I\\
	\includegraphics[width=\textwidth]{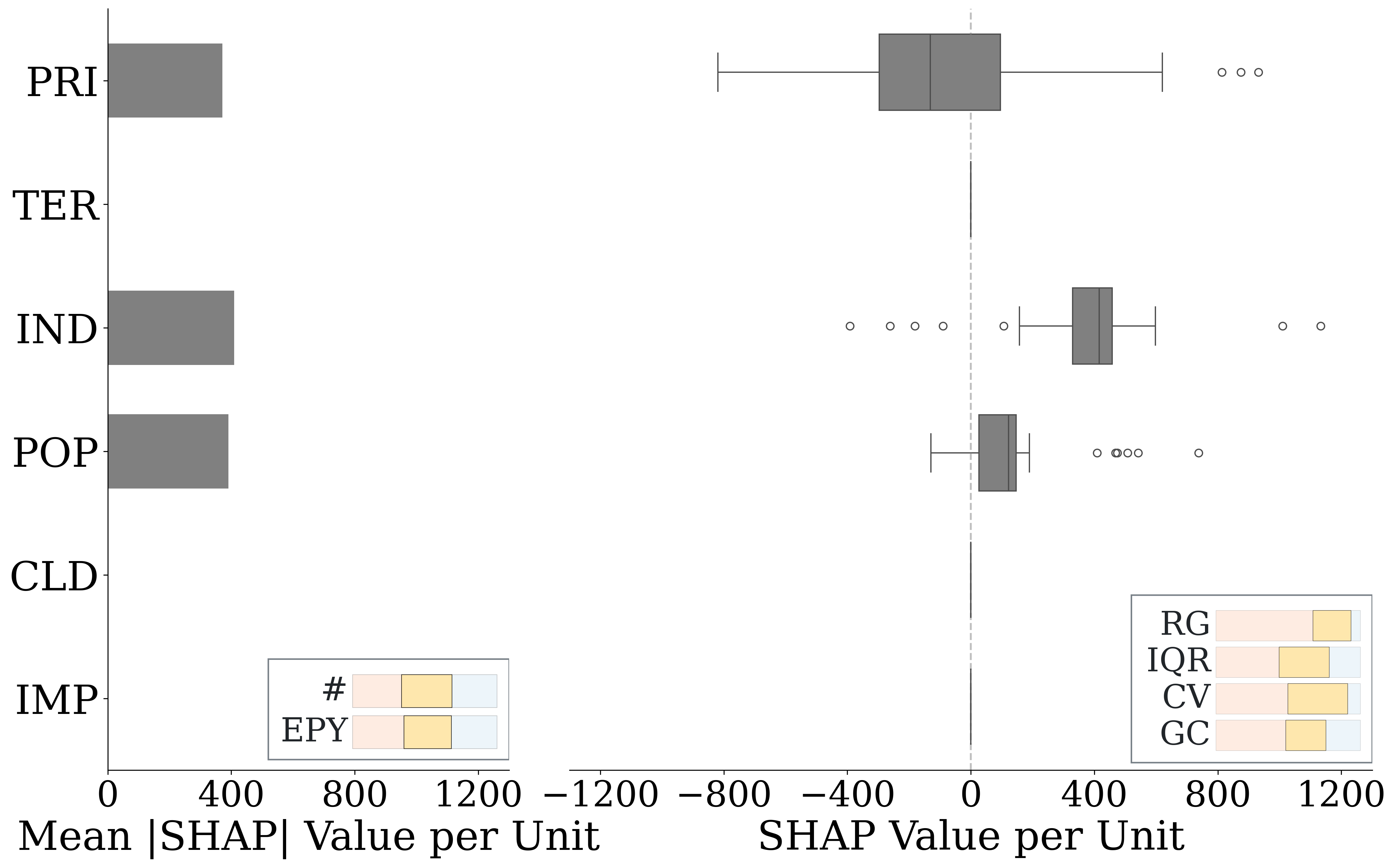}\\
	Type II\\
	\includegraphics[width=\textwidth]{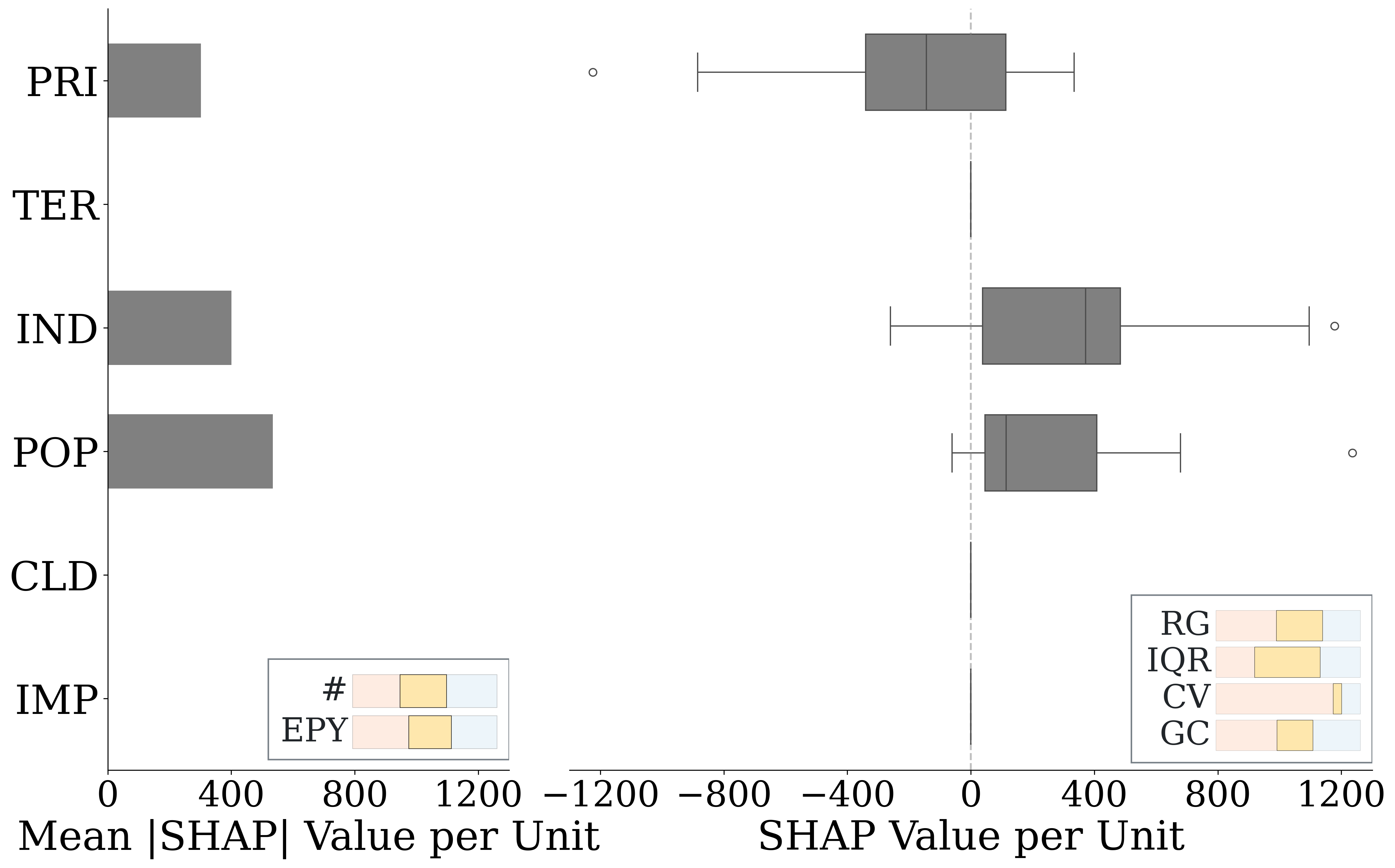}
	 \subcaption{GT\\(msl=5, md=4, $\ell_{2}(gwr)$)}
	  \end{subfigure}
	\begin{subfigure}{0.32\textwidth}
	\centering
	\includegraphics[width=\textwidth]{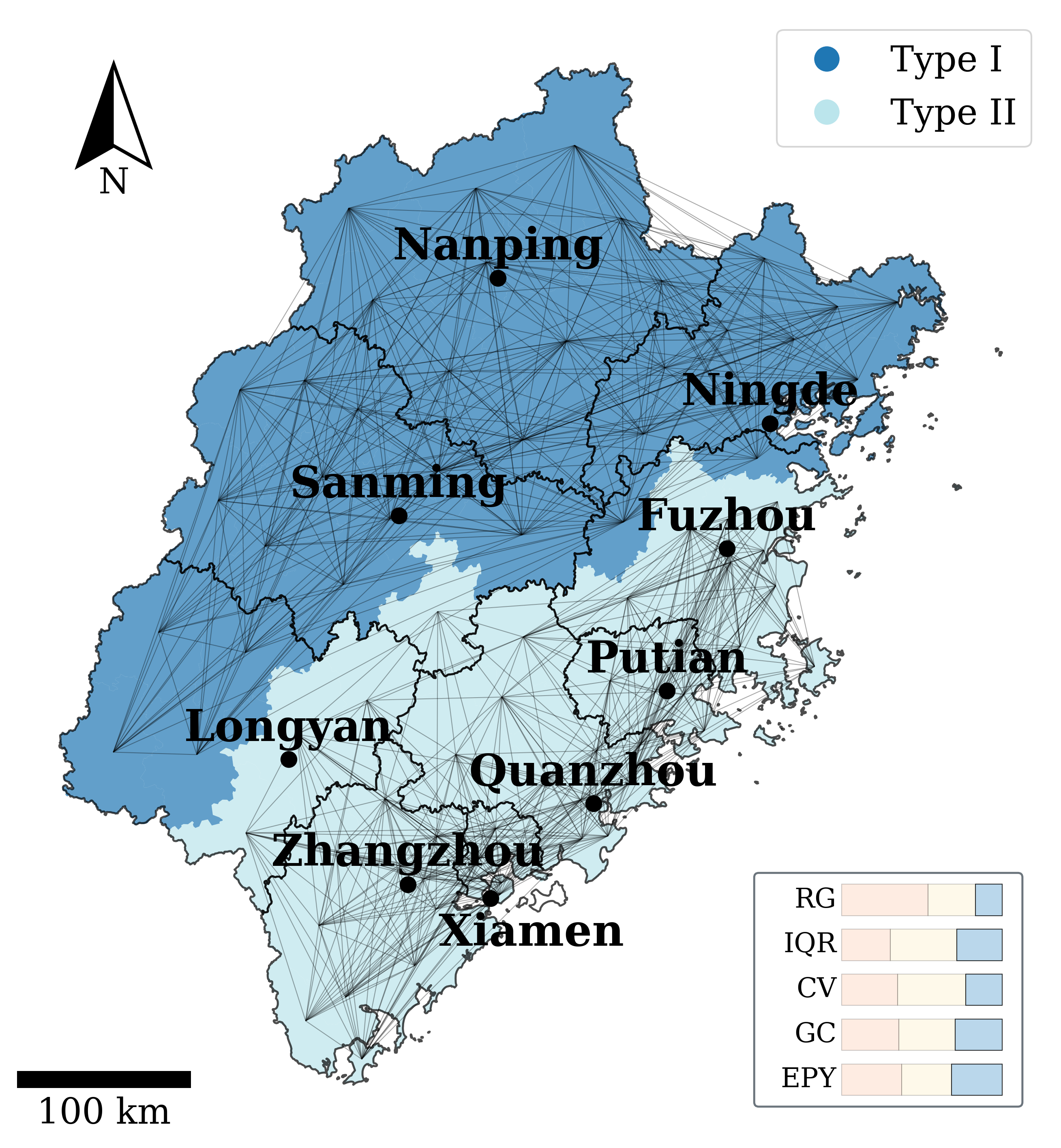}
	 \scriptsize Type I\\
	\includegraphics[width=\textwidth]{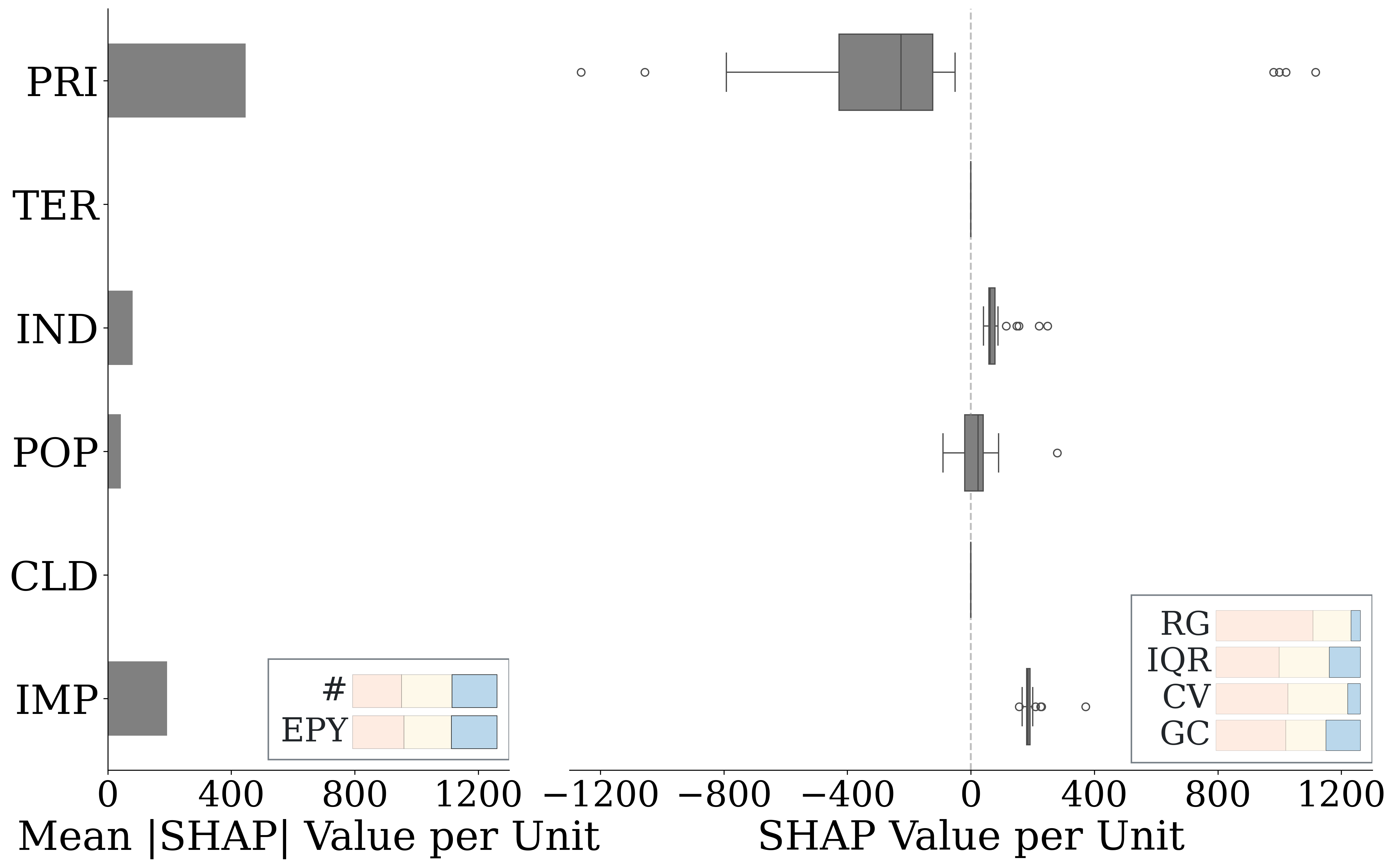}\\
	Type II\\
	\includegraphics[width=\textwidth]{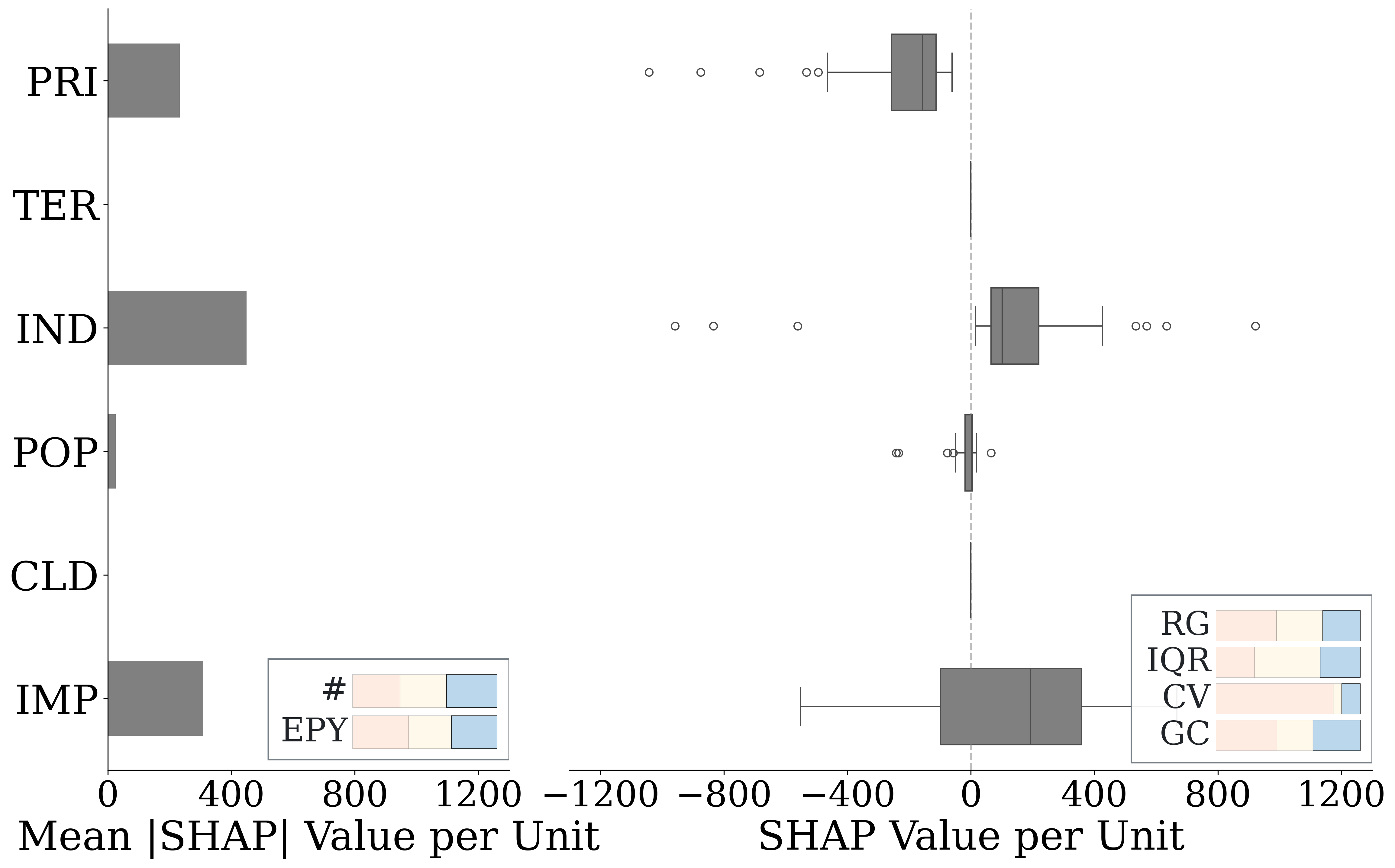}
	 \subcaption{Ours\\(msl=5, md=5, $\ell_{2}(gwr)$)}
	 \end{subfigure}
	\caption{The derived spatial communities of counties with similar feature coefficients and attributions in Fujian. The SHAP scores for each feature in each community are compared across the three models based on the predefined dispersion indicators.}
	\label{fig7:fujian-gwr}
\end{figure}

\begin{figure}[h!]
\captionsetup[subfigure]{justification=centering}
	\centering
	\begin{subfigure}{0.32\textwidth}
	\centering
	\includegraphics[width=\textwidth]{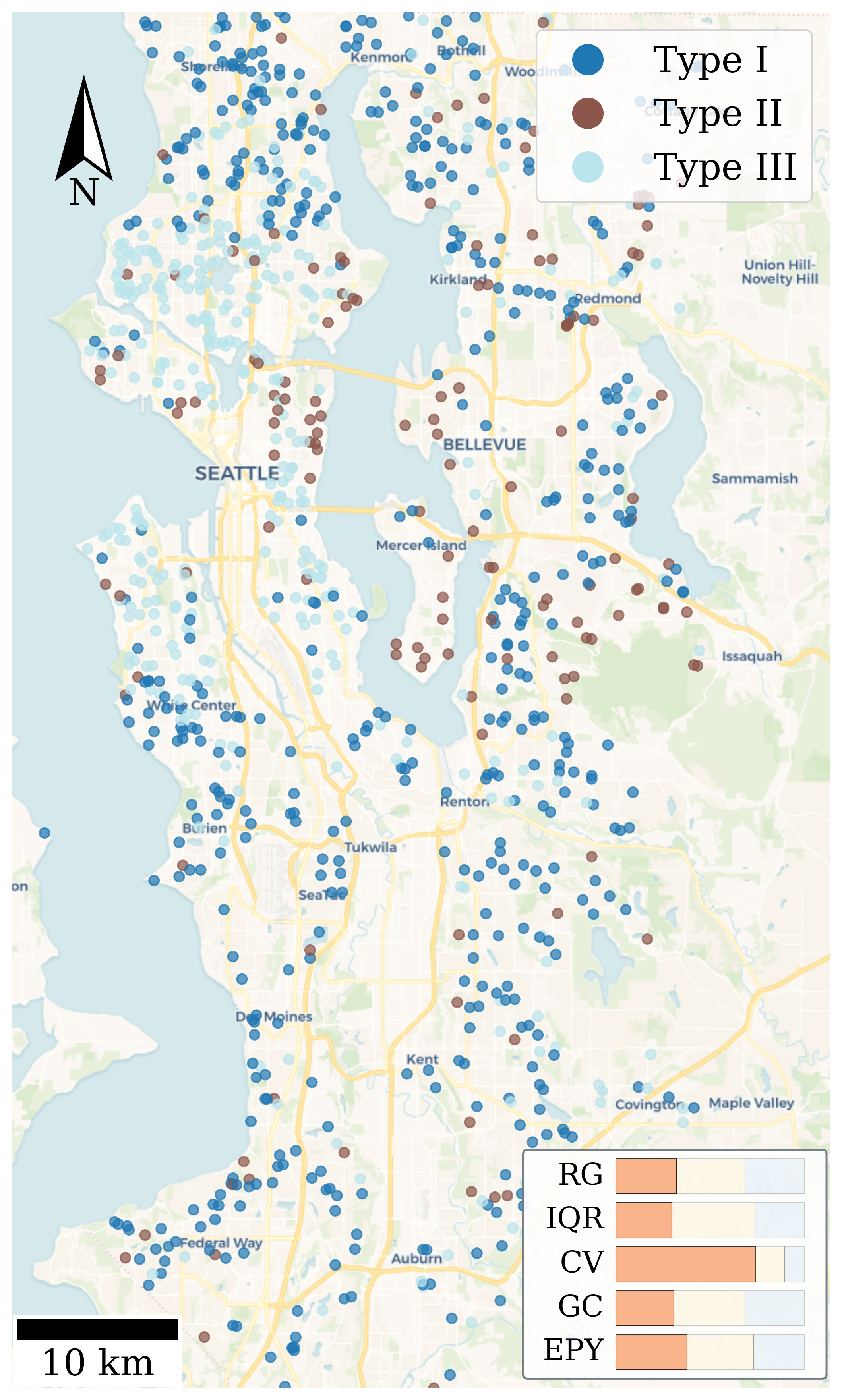}
	  \scriptsize Type I\\
	\includegraphics[width=\textwidth]{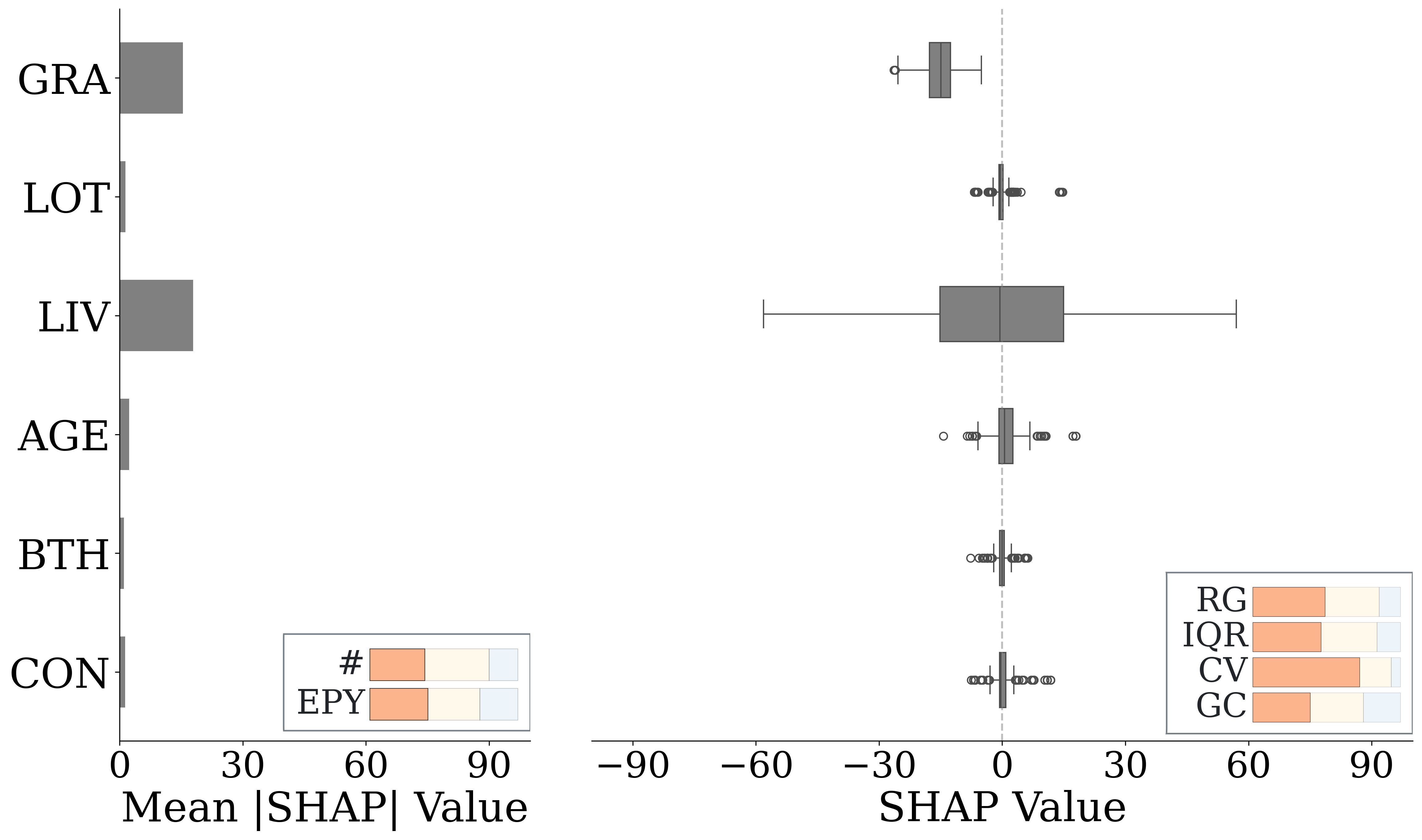}\\
	Type II\\
	\includegraphics[width=\textwidth]{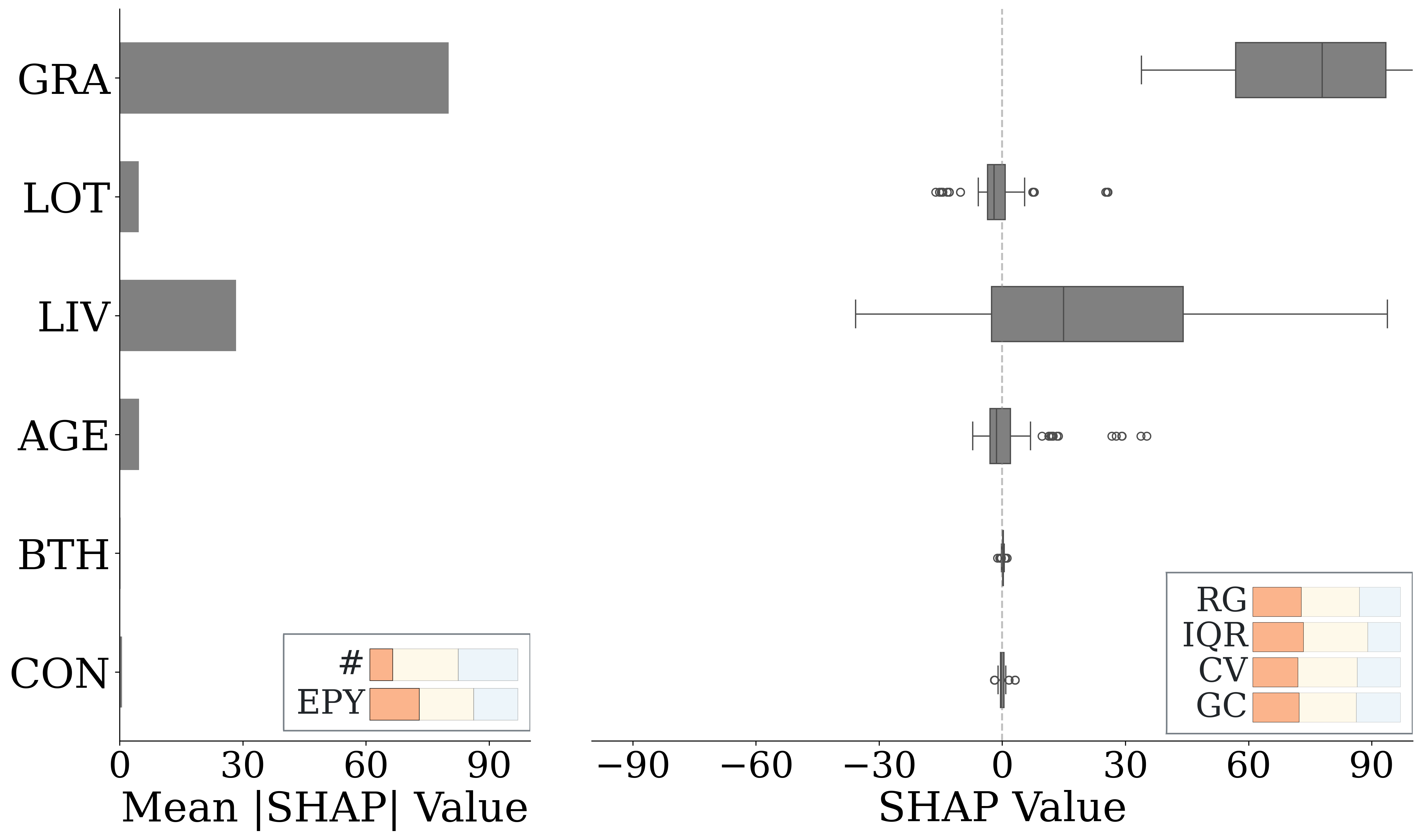}\\
	Type III\\
	\includegraphics[width=\textwidth]{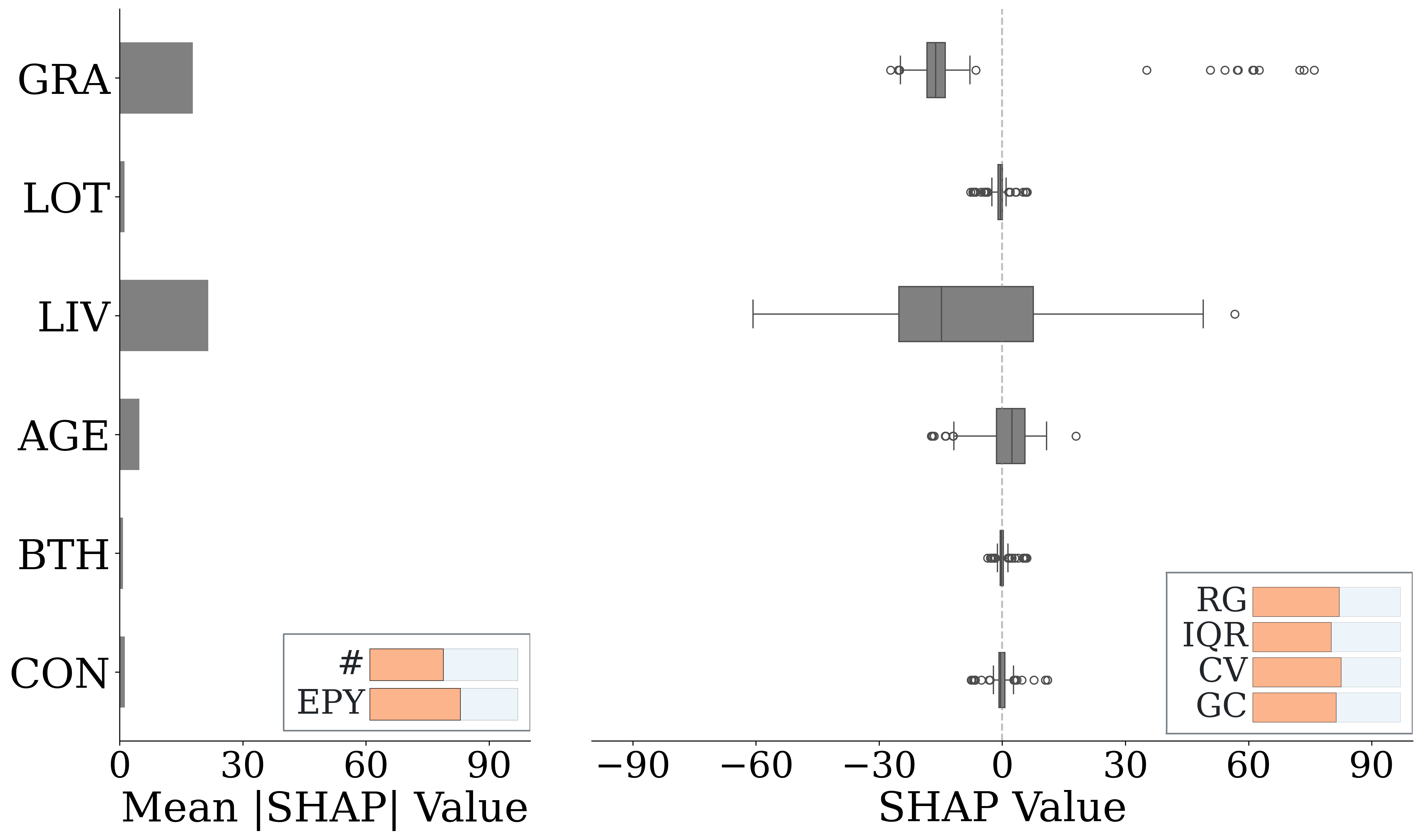}
	 \subcaption{DT\\(msl=6, md=8, $\ell_{2}(feature)$)}
	 \end{subfigure}
	 \begin{subfigure}{0.32\textwidth}
	\centering
	\includegraphics[width=\textwidth]{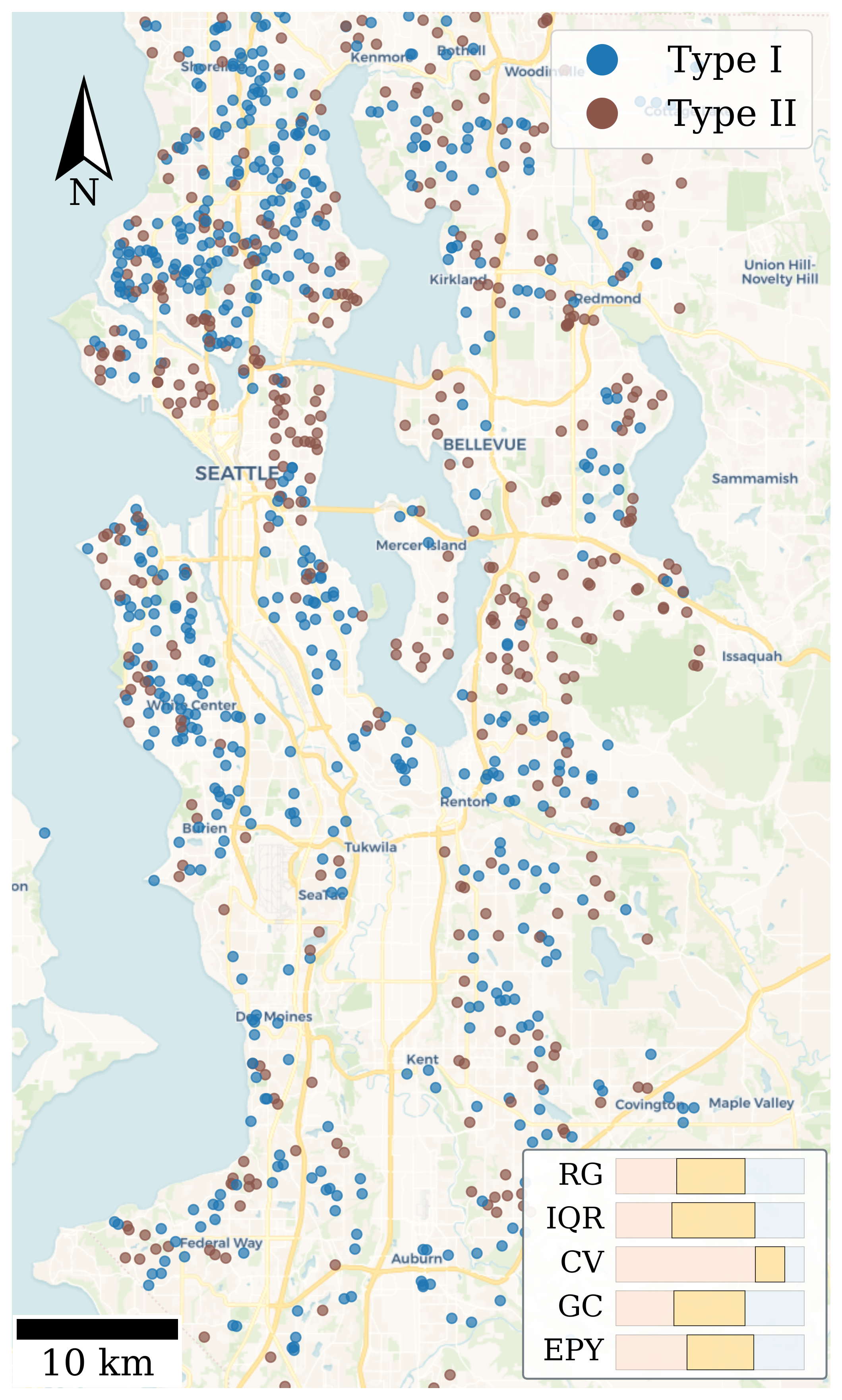}
	  \scriptsize Type I\\
	\includegraphics[width=\textwidth]{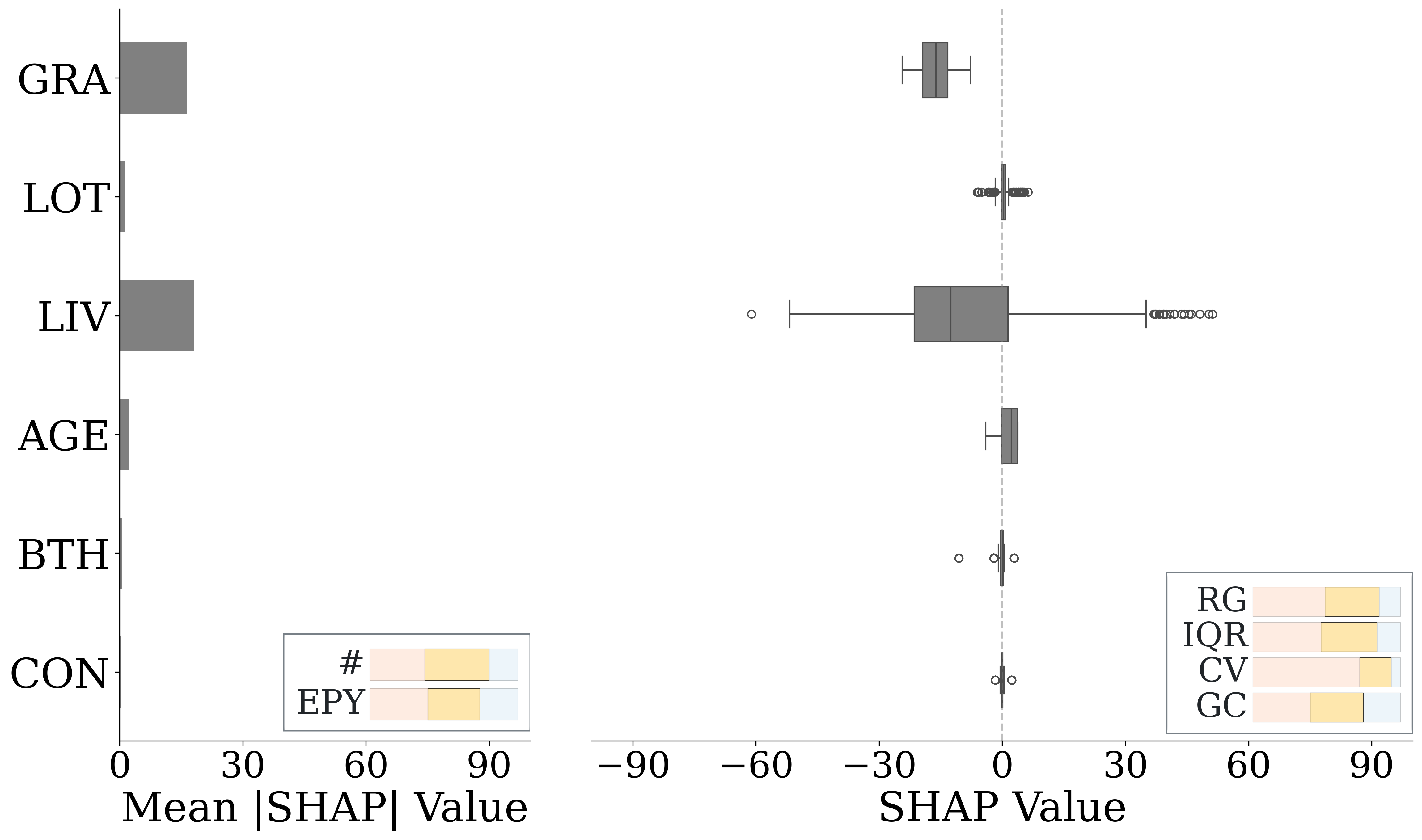}\\
	Type II\\
	\includegraphics[width=\textwidth]{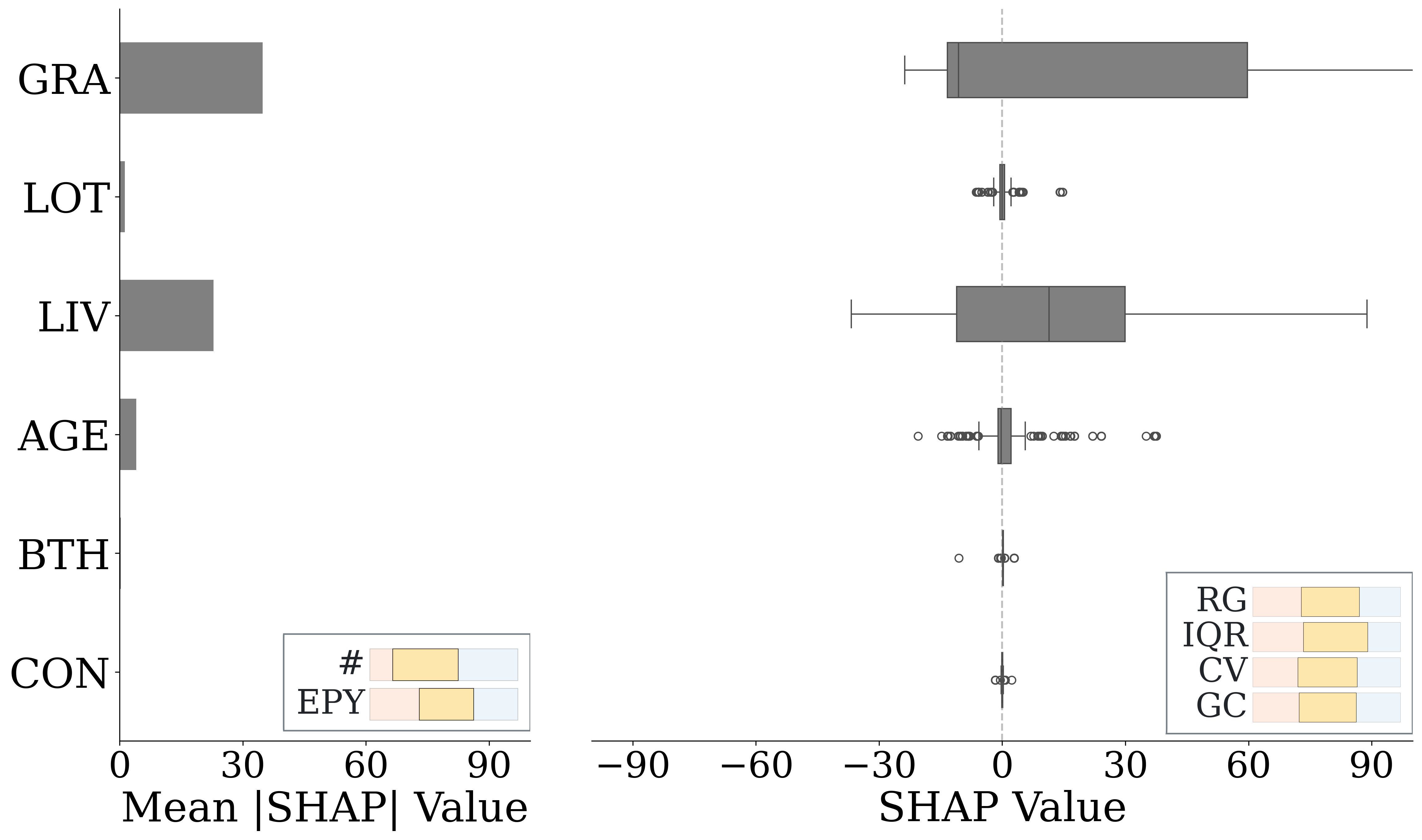}\\
	Type III (absent)\\
	\includegraphics[width=\textwidth]{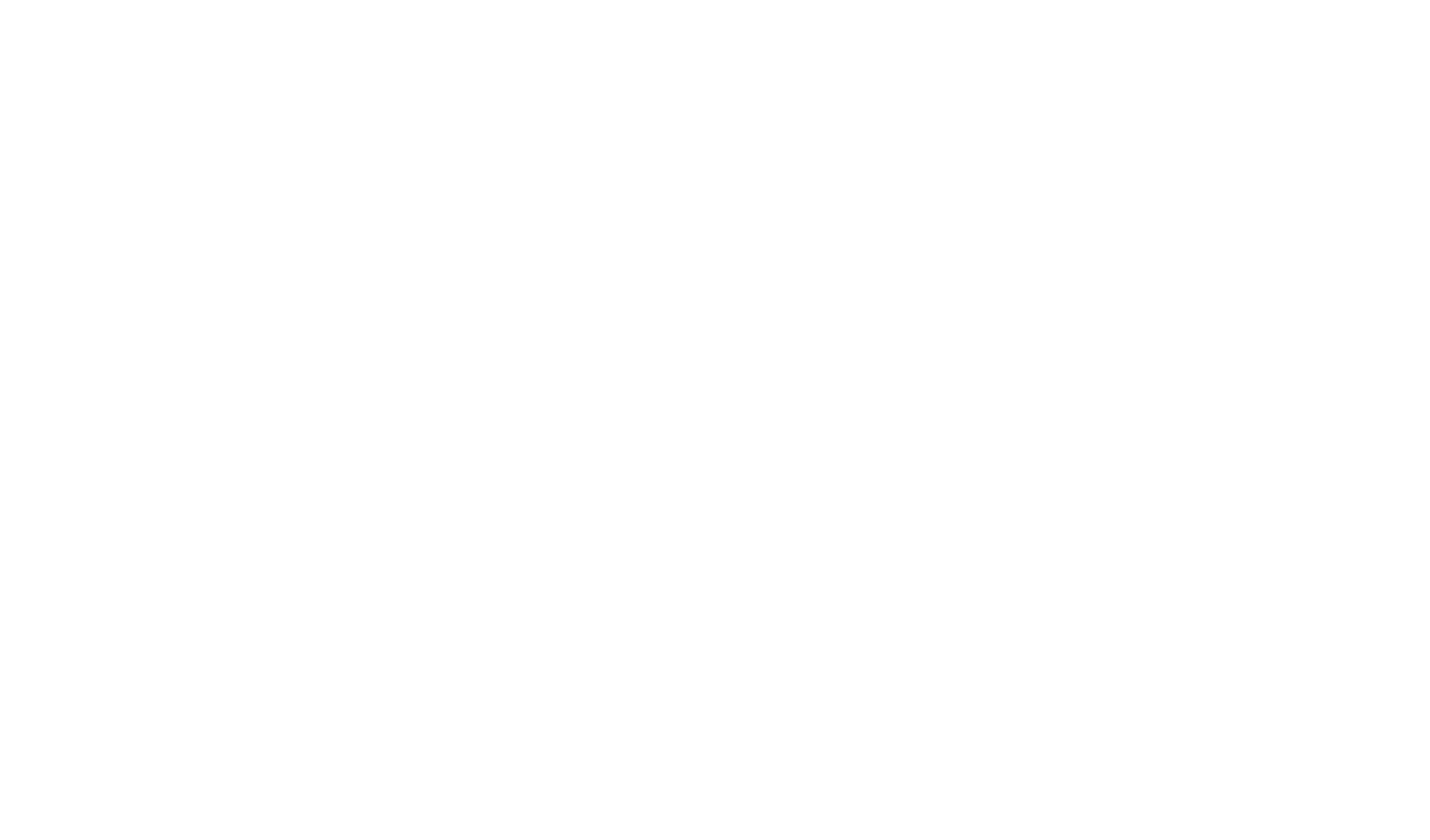}
	 \subcaption{GT\\(msl=6, md=8, $\ell_{2}(feature)$)}
	  \end{subfigure}
	\begin{subfigure}{0.32\textwidth}
	\centering
	\includegraphics[width=\textwidth]{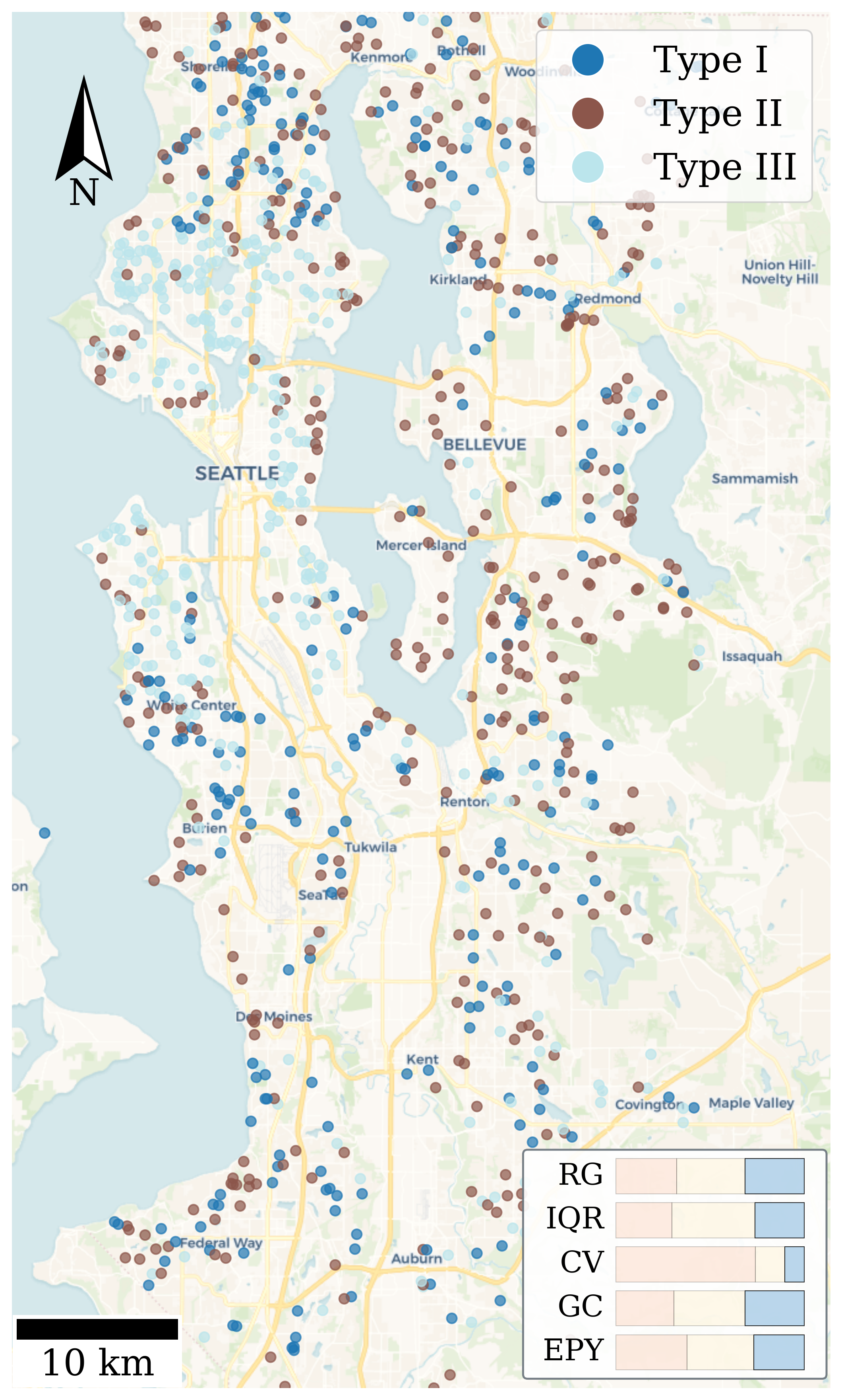}
	  \scriptsize Type I\\
	\includegraphics[width=\textwidth]{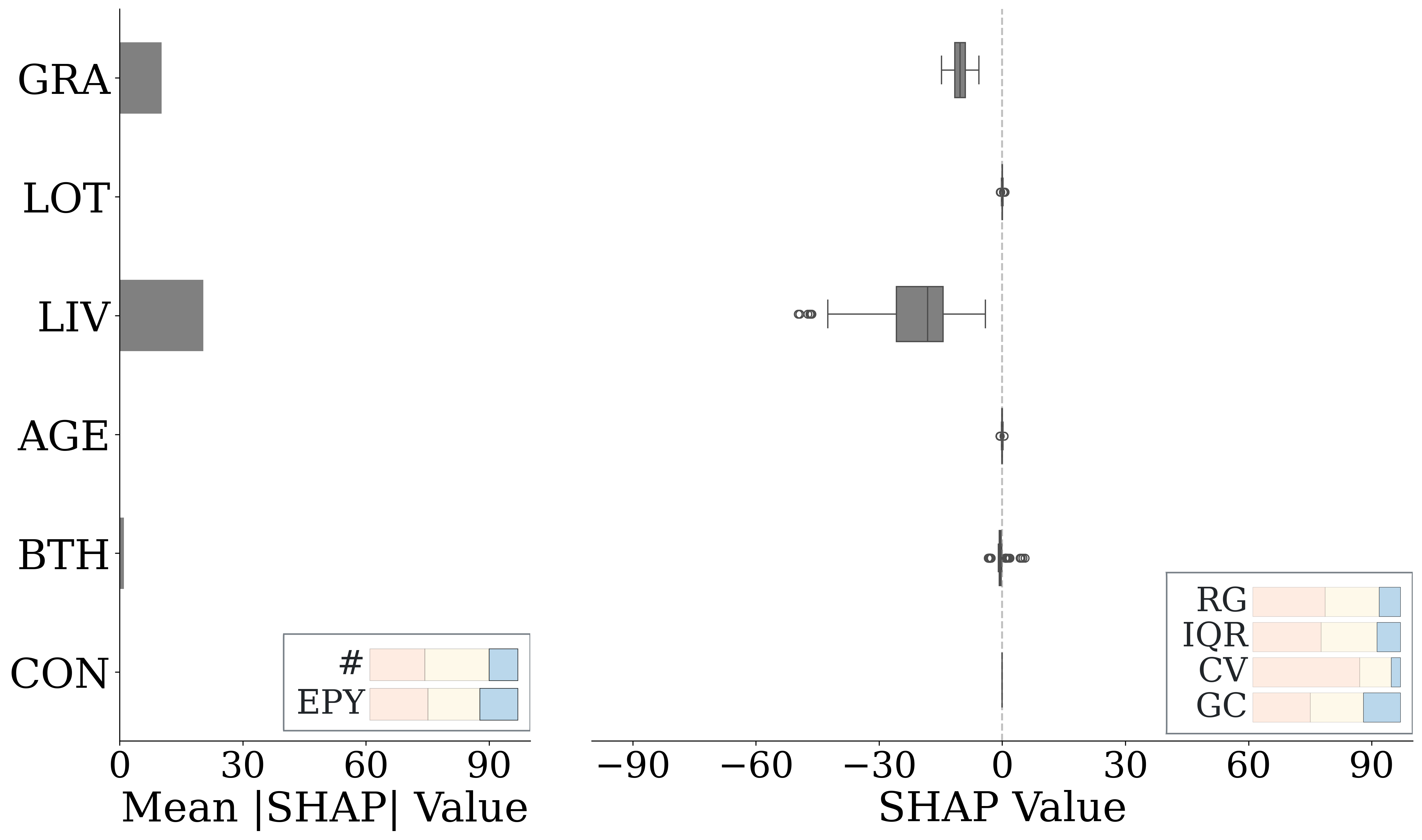}\\
	Type II\\
	\includegraphics[width=\textwidth]{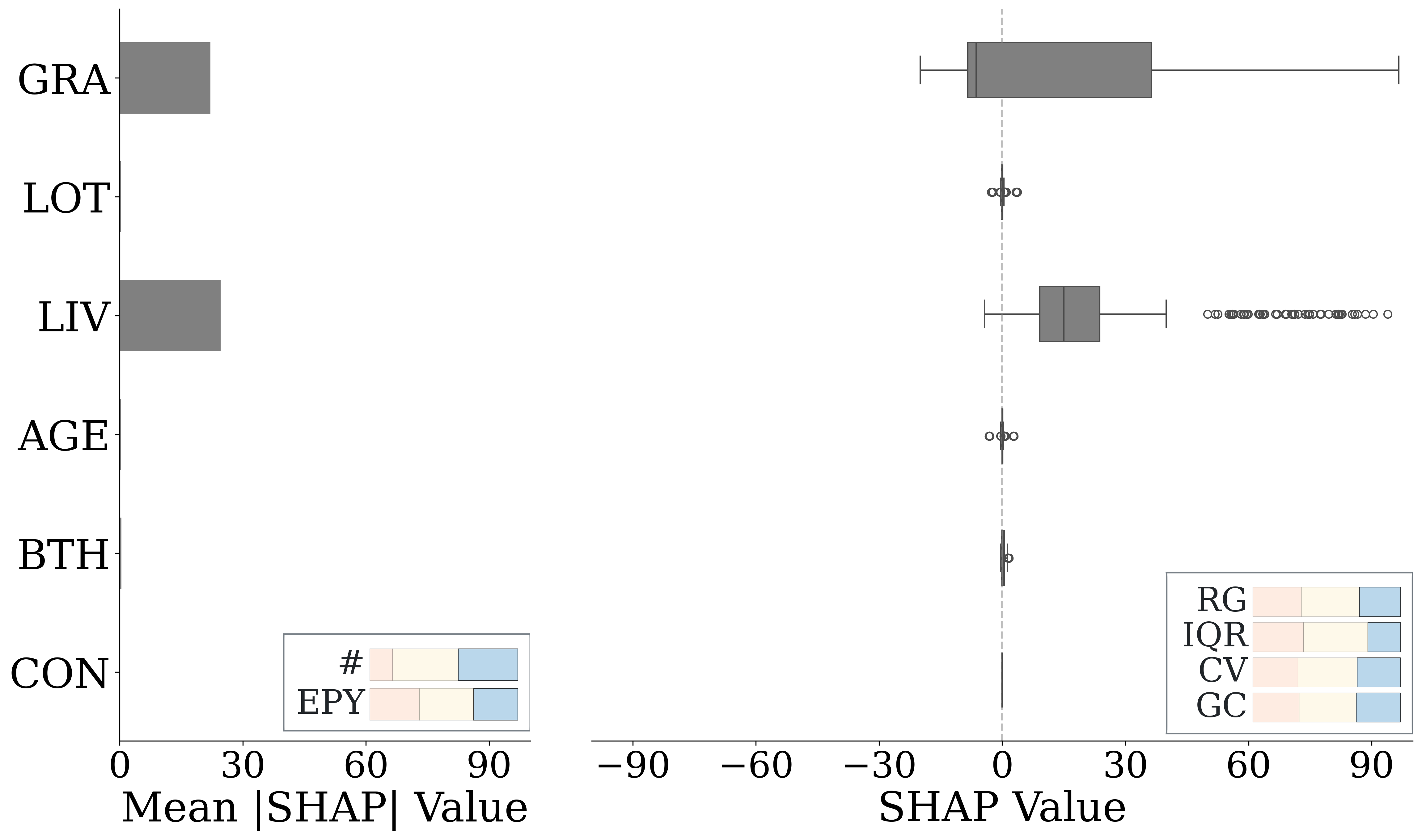}\\
	Type III\\
	\includegraphics[width=\textwidth]{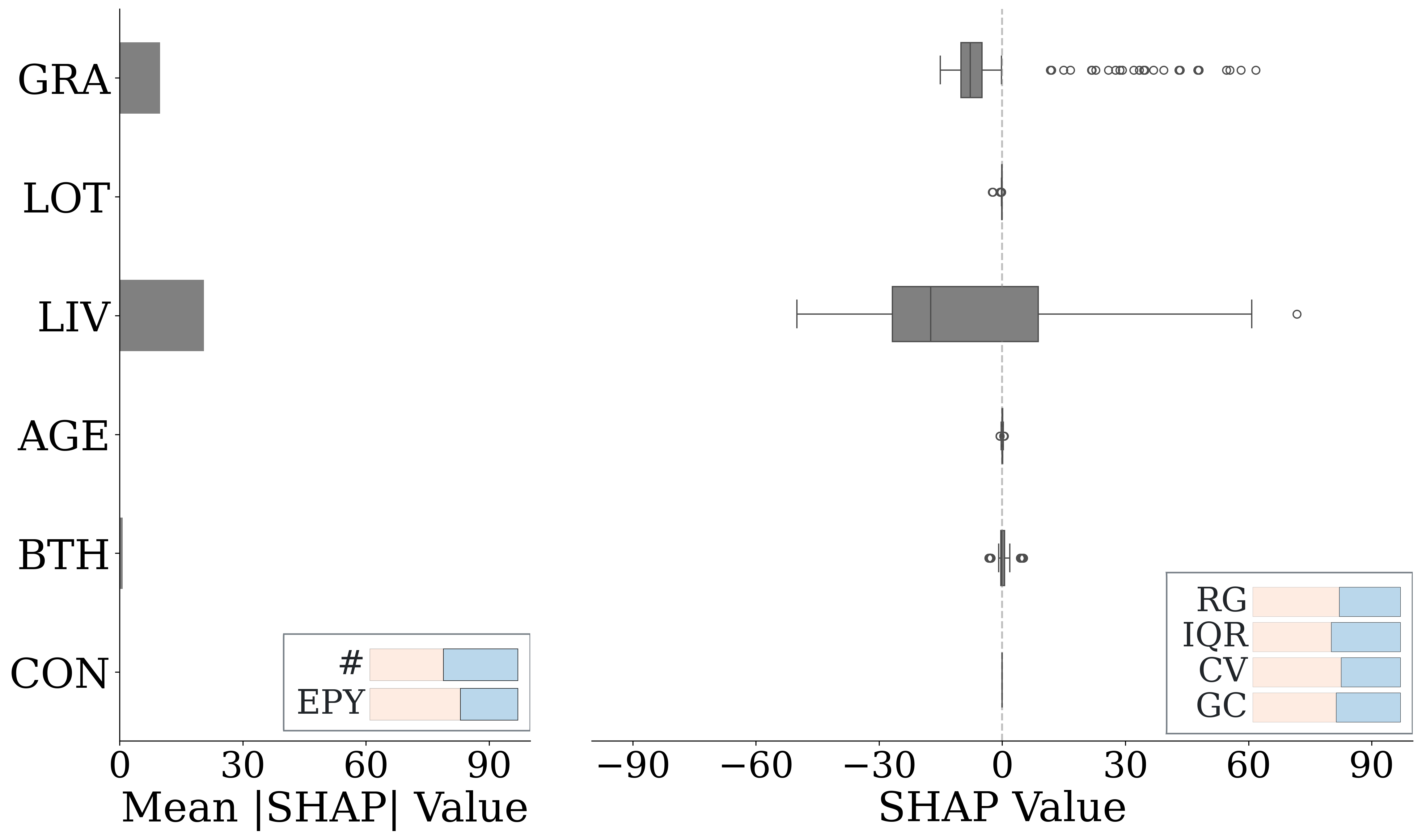}
	 \subcaption{Ours\\(msl=6, md=8, $\ell_{2}(feature)$)}
	 \end{subfigure}
	\caption{The derived spatial communities of houses with similar feature values and attributions in Seattle. The SHAP scores for each feature in each community are compared across the three models based on the predefined dispersion indicators.}
	\label{fig8:seattle-feature}
\end{figure}

\begin{figure}[h!]
\captionsetup[subfigure]{justification=centering}
	\centering
	\begin{subfigure}{0.32\textwidth}
	\centering
	\includegraphics[width=\textwidth]{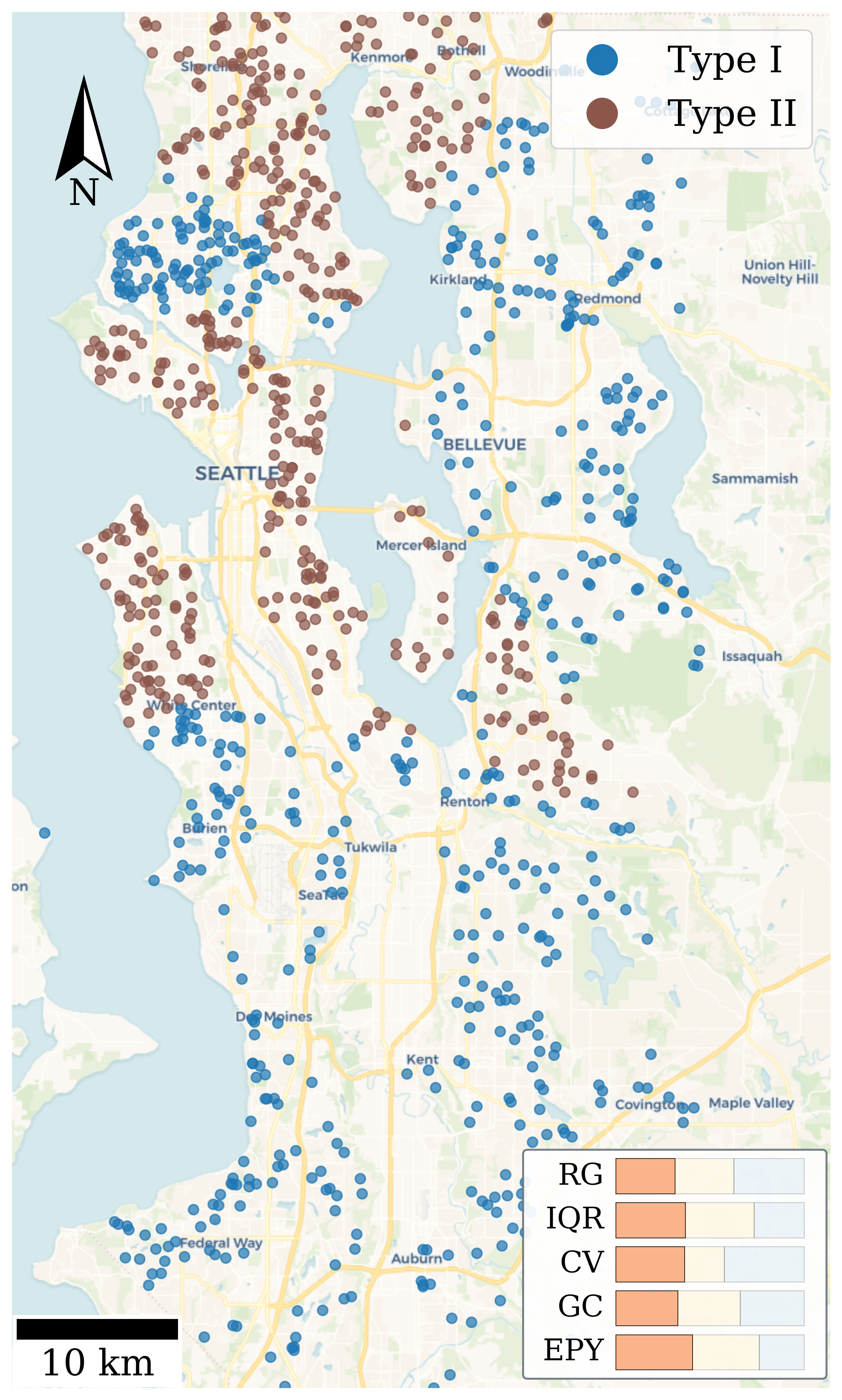}
	  \scriptsize Type I\\
	\includegraphics[width=\textwidth]{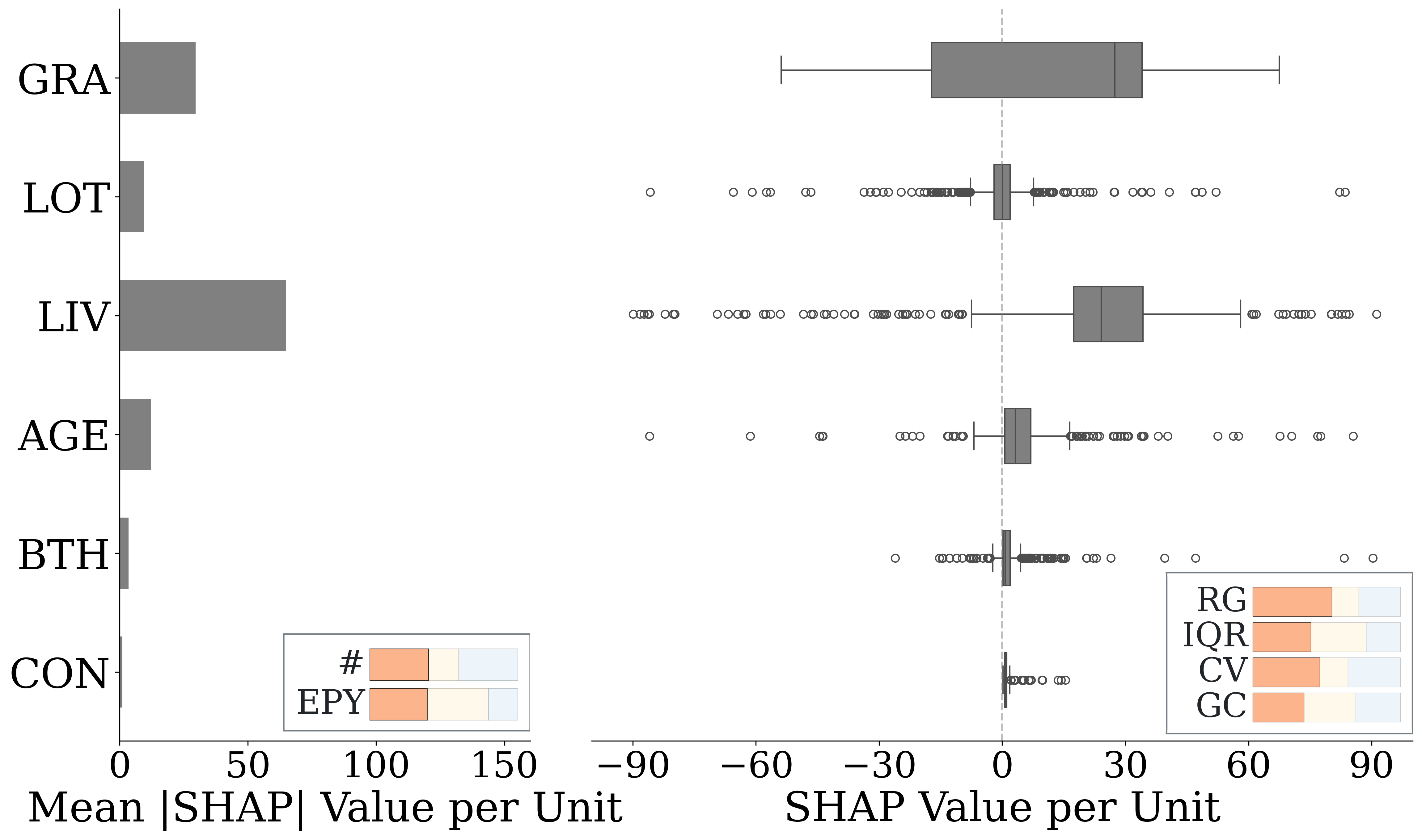}\\
	Type II\\
	\includegraphics[width=\textwidth]{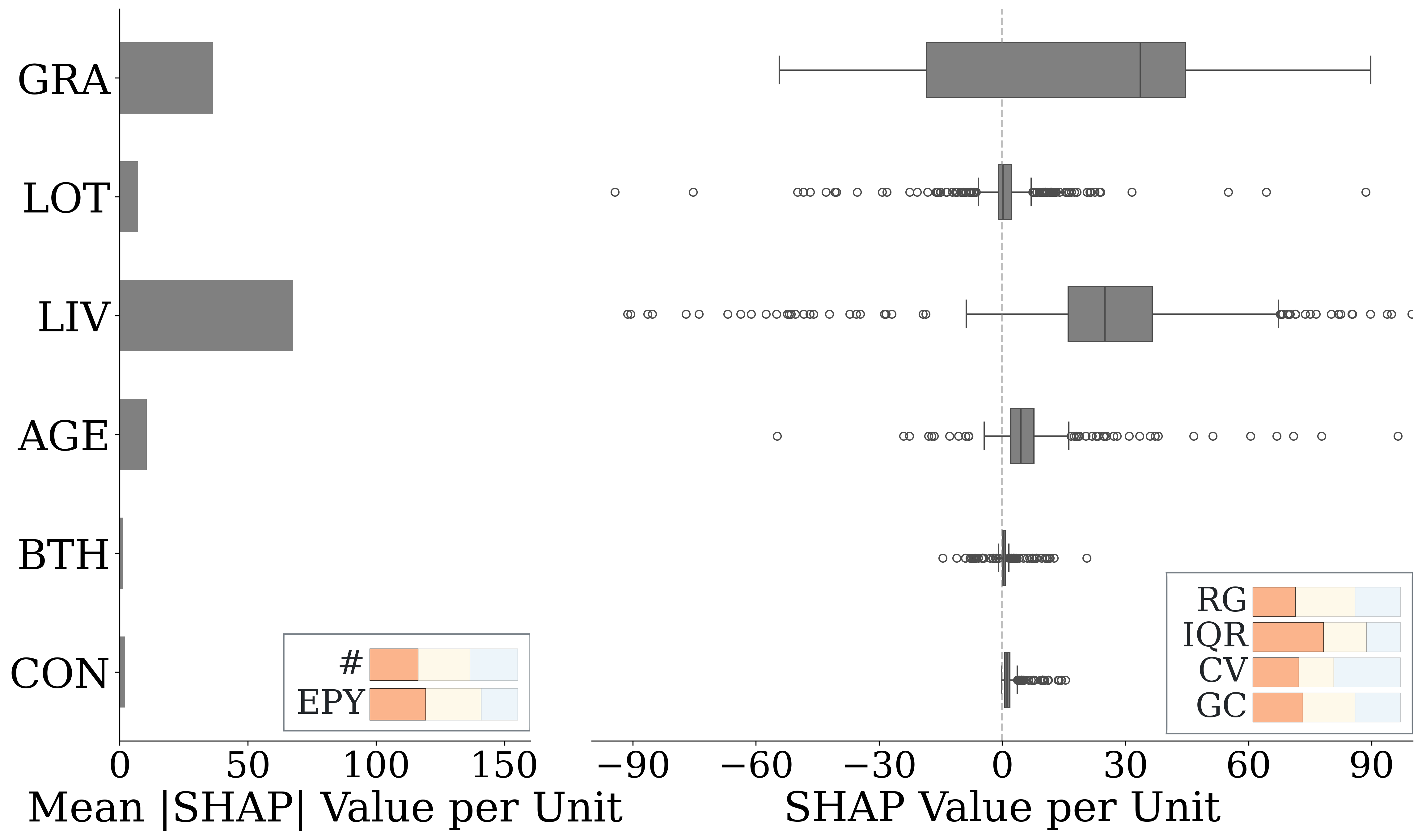}\\
	Type III (absent)\\
	\includegraphics[width=\textwidth]{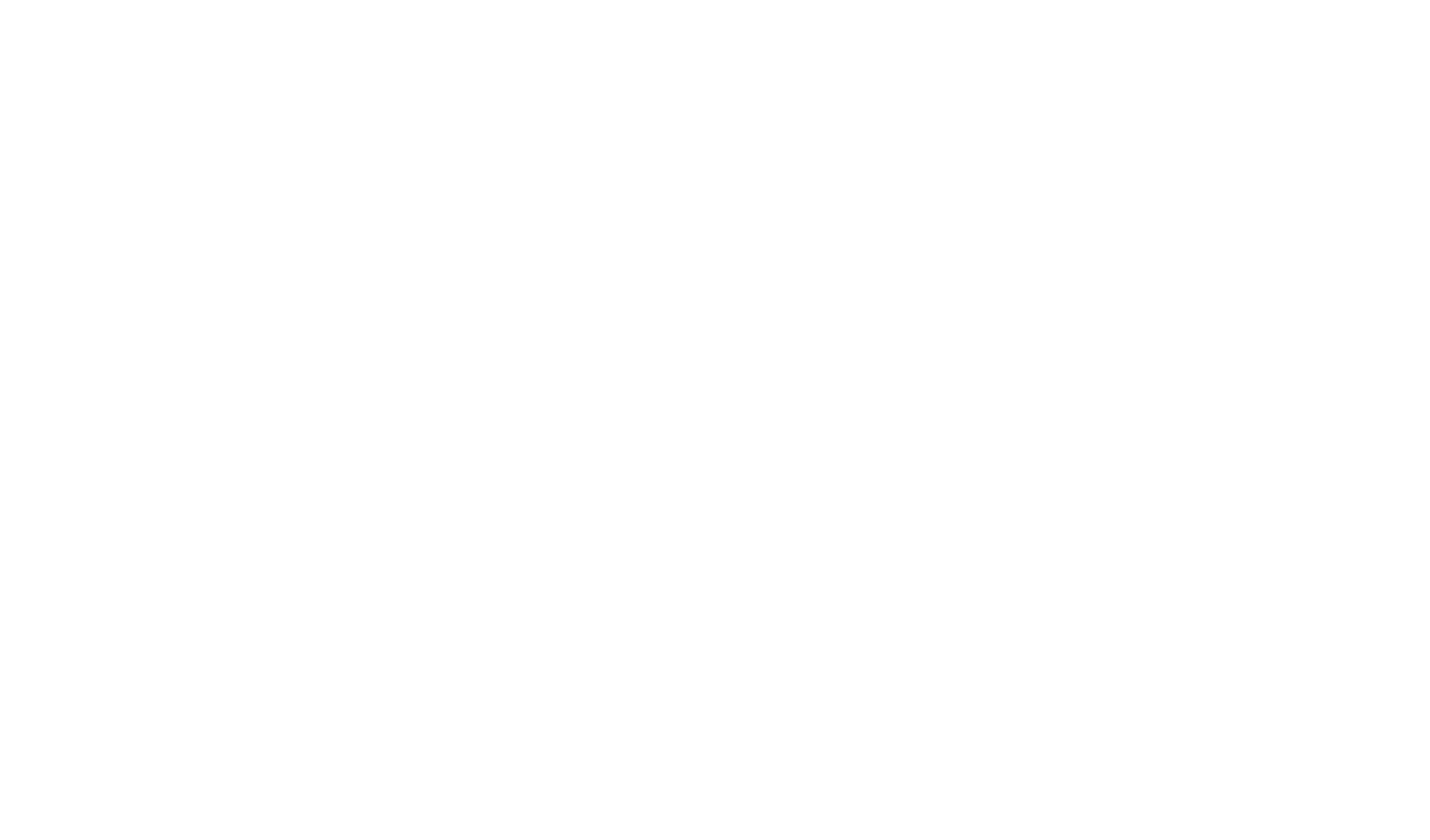}
	 \subcaption{DT\\(msl=6, md=8, $\ell_{2}(gwr)$)}
	 \end{subfigure}
	 \begin{subfigure}{0.32\textwidth}
	\centering
	\includegraphics[width=\textwidth]{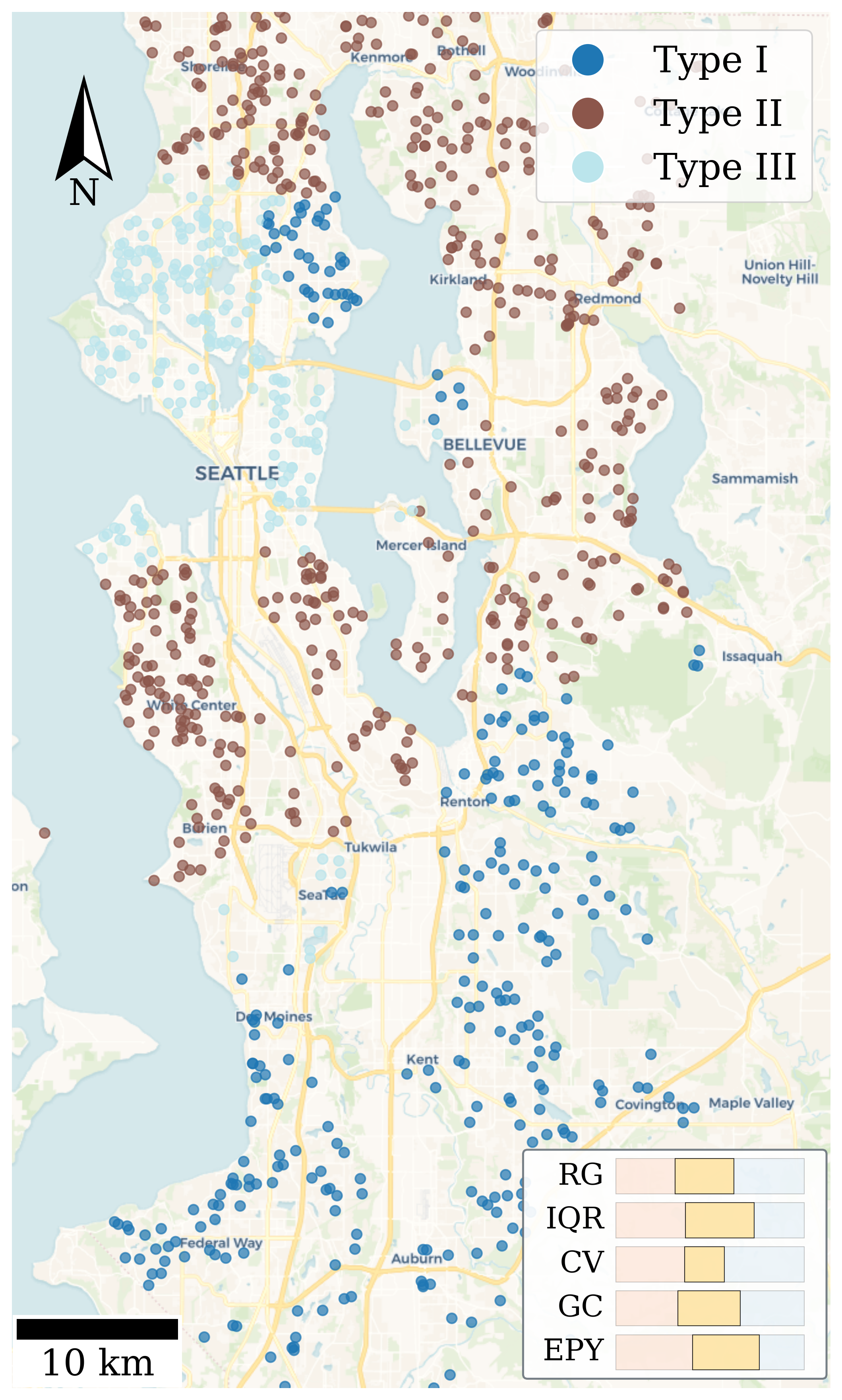}
	  \scriptsize Type I\\
	\includegraphics[width=\textwidth]{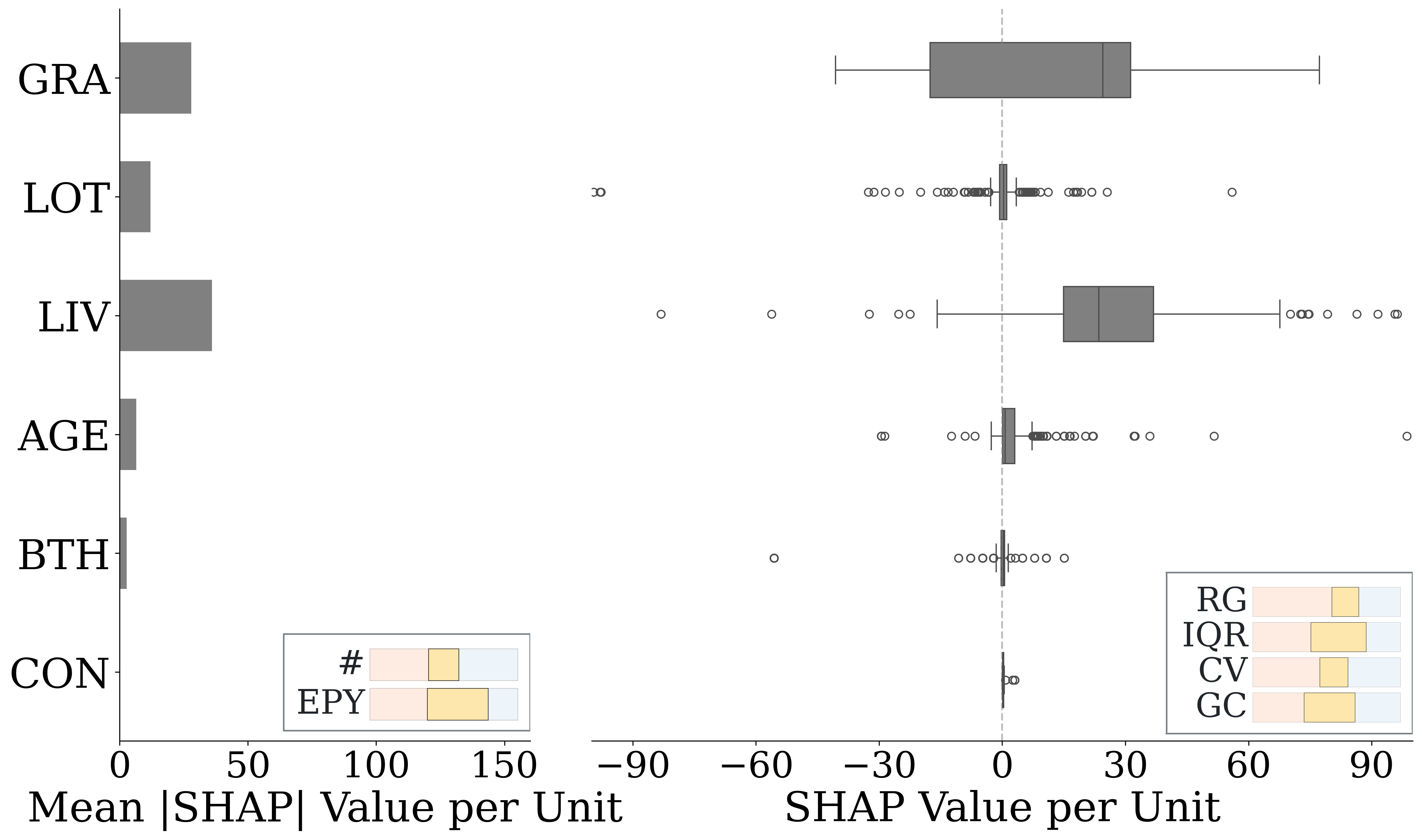}\\
	Type II\\
	\includegraphics[width=\textwidth]{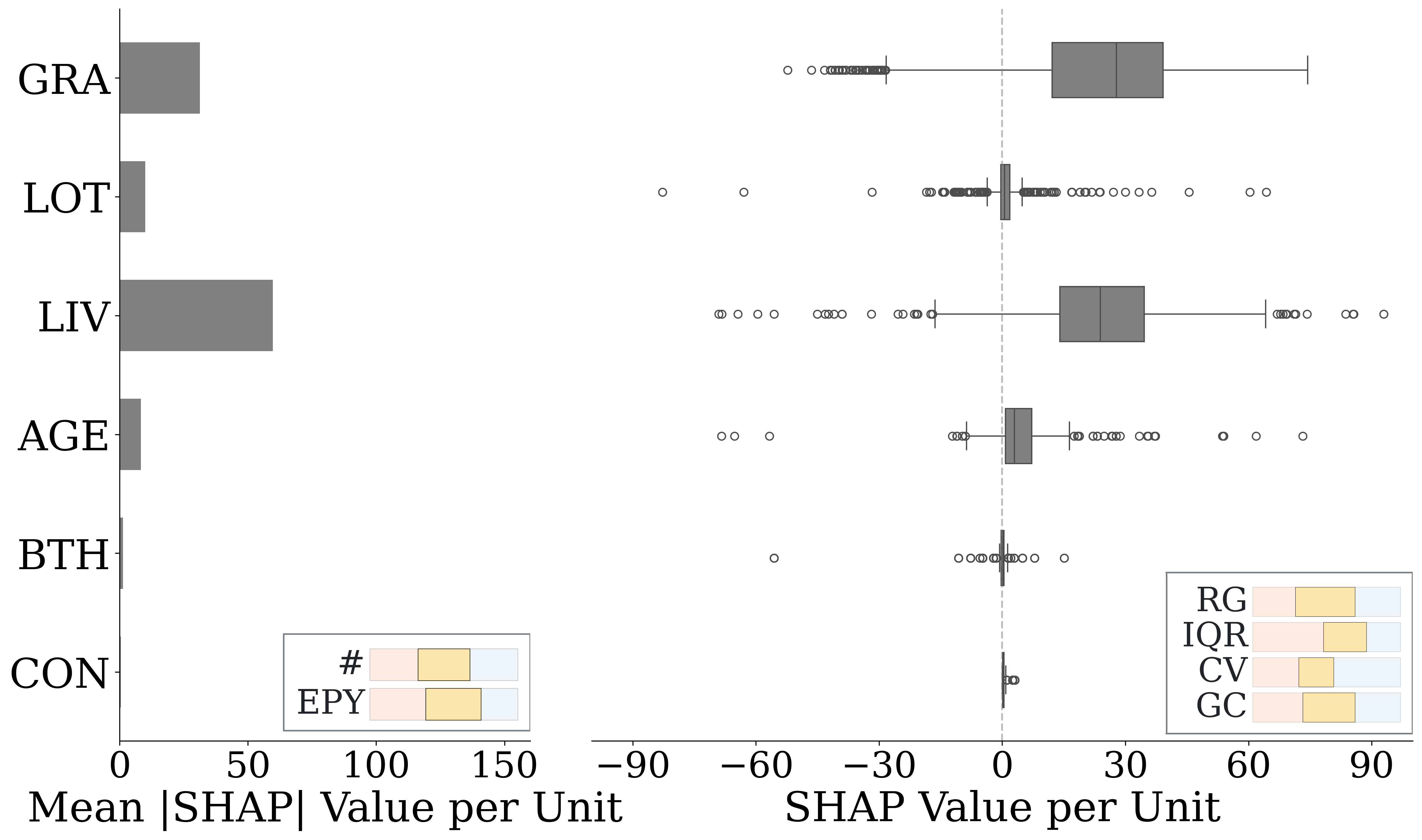}\\
	Type III\\
	\includegraphics[width=\textwidth]{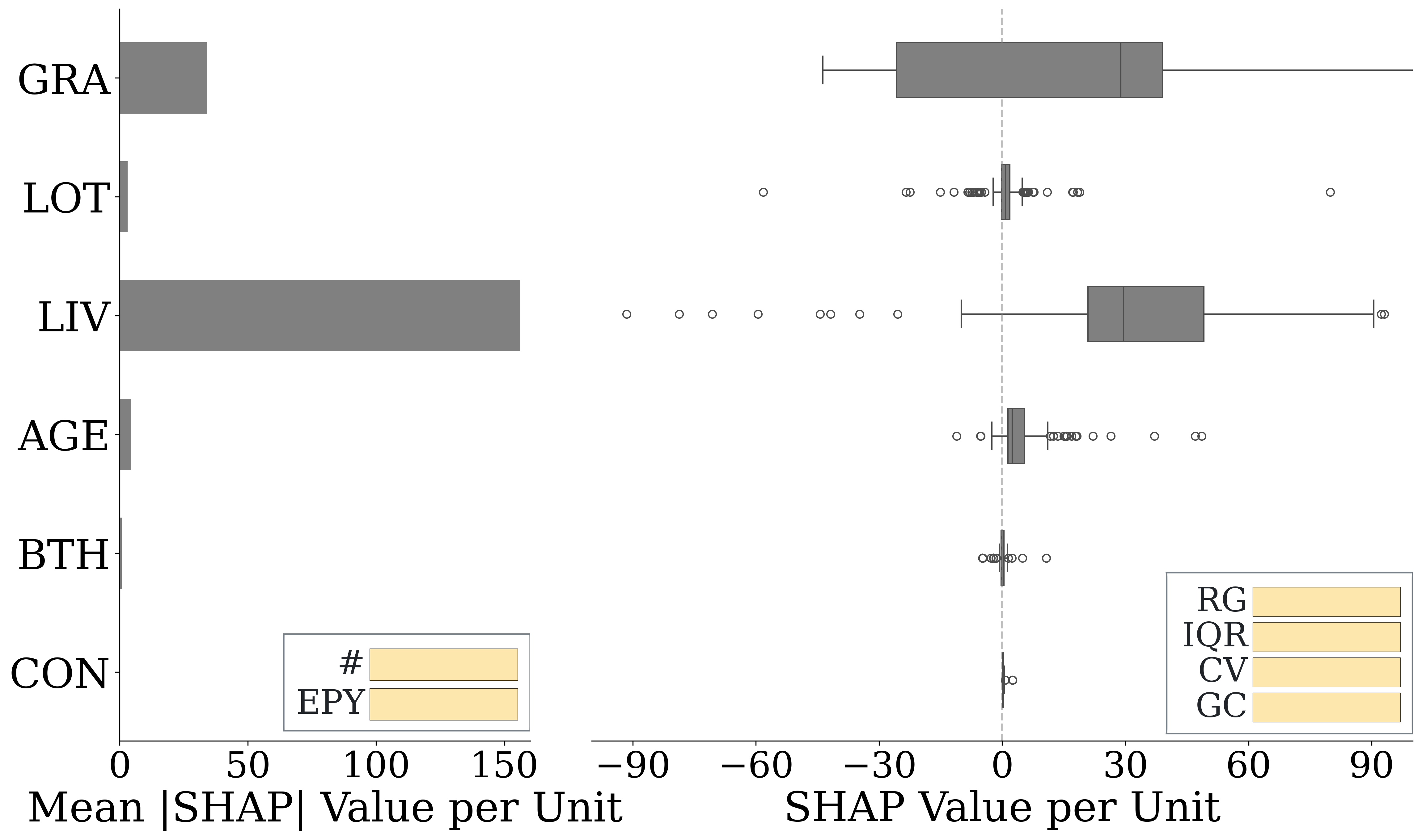}
	 \subcaption{GT\\(msl=6, md=8, $\ell_{2}(gwr)$)}
	  \end{subfigure}
	\begin{subfigure}{0.32\textwidth}
	\centering
	\includegraphics[width=\textwidth]{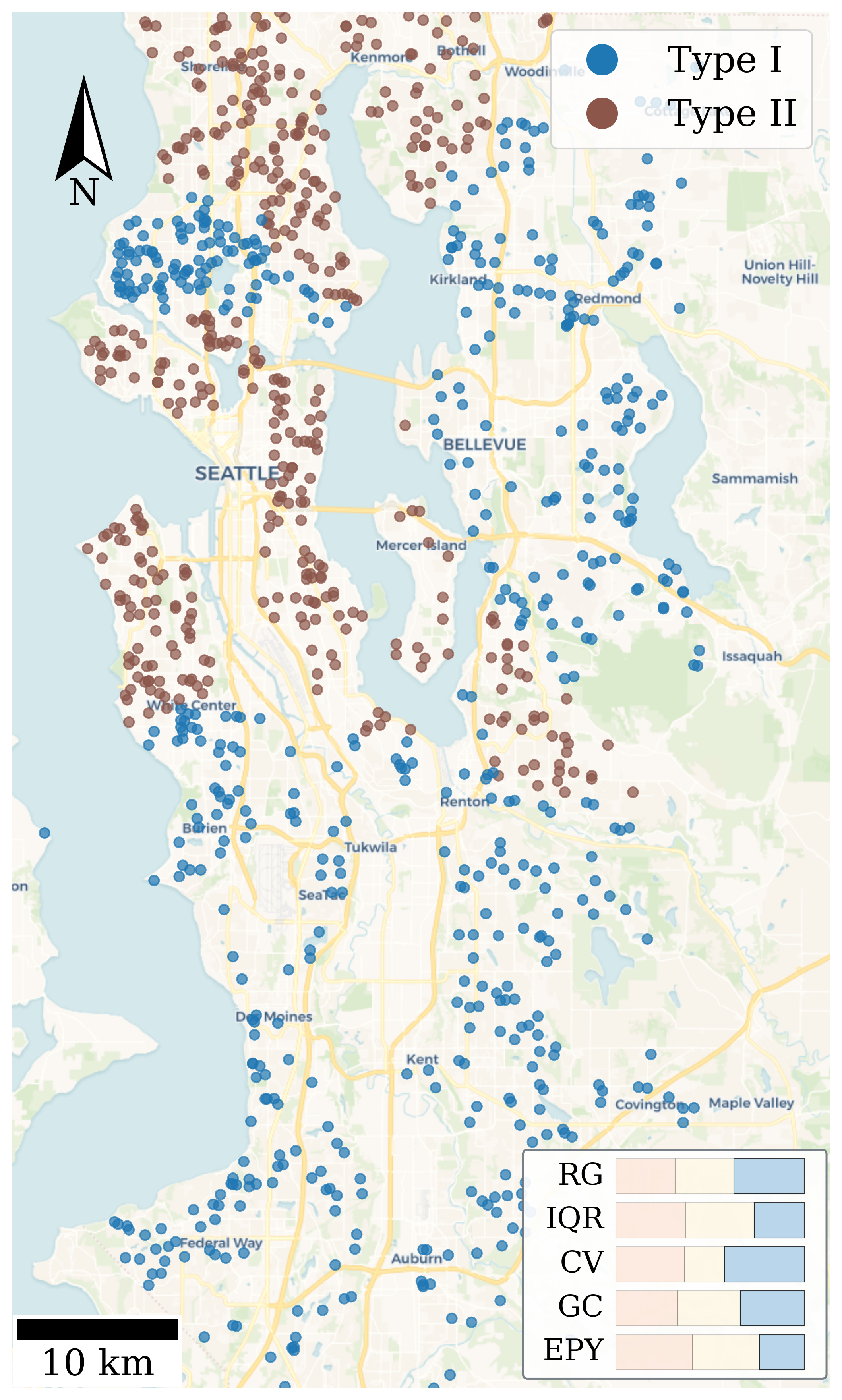}
	  \scriptsize Type I\\
	\includegraphics[width=\textwidth]{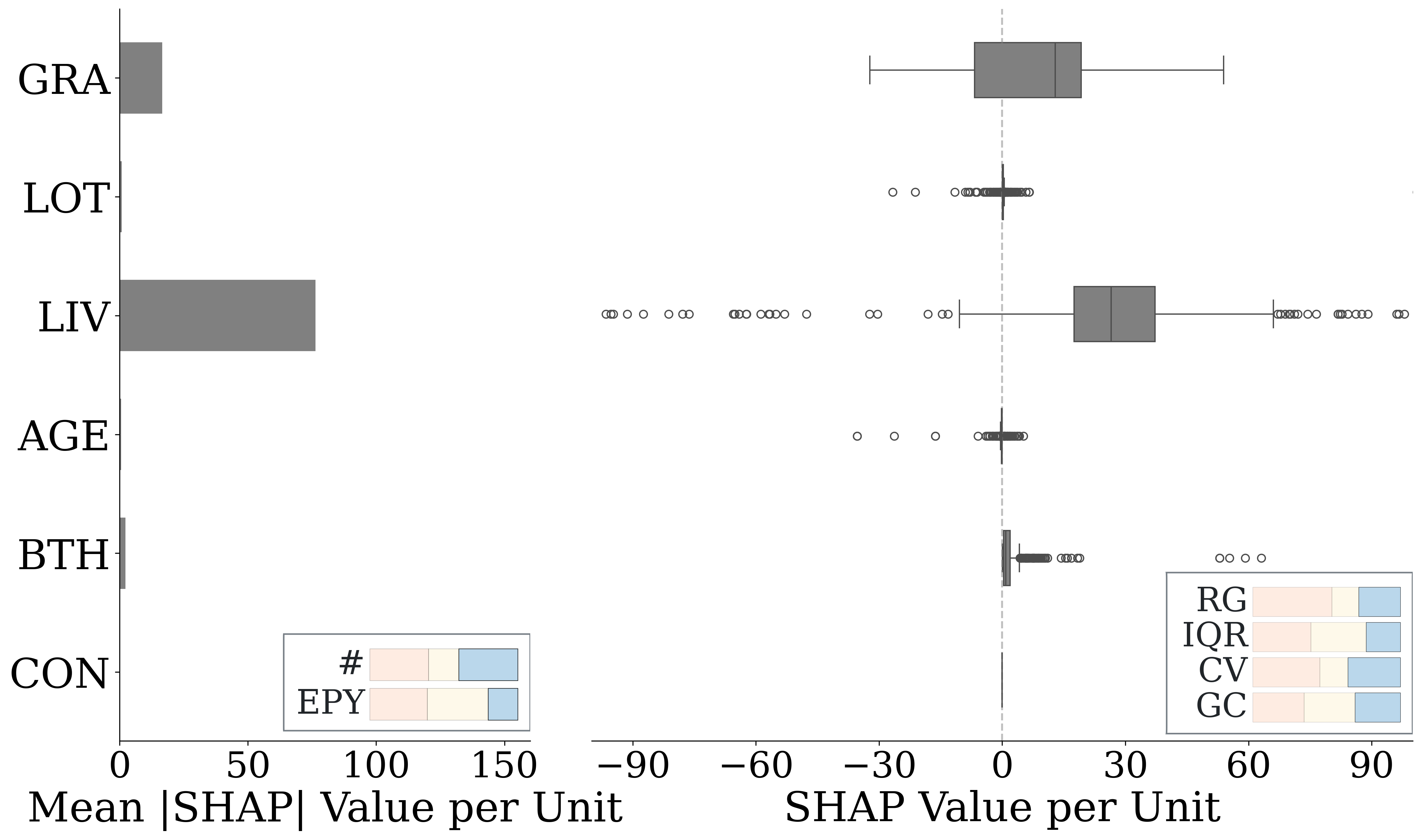}\\
	Type II\\
	\includegraphics[width=\textwidth]{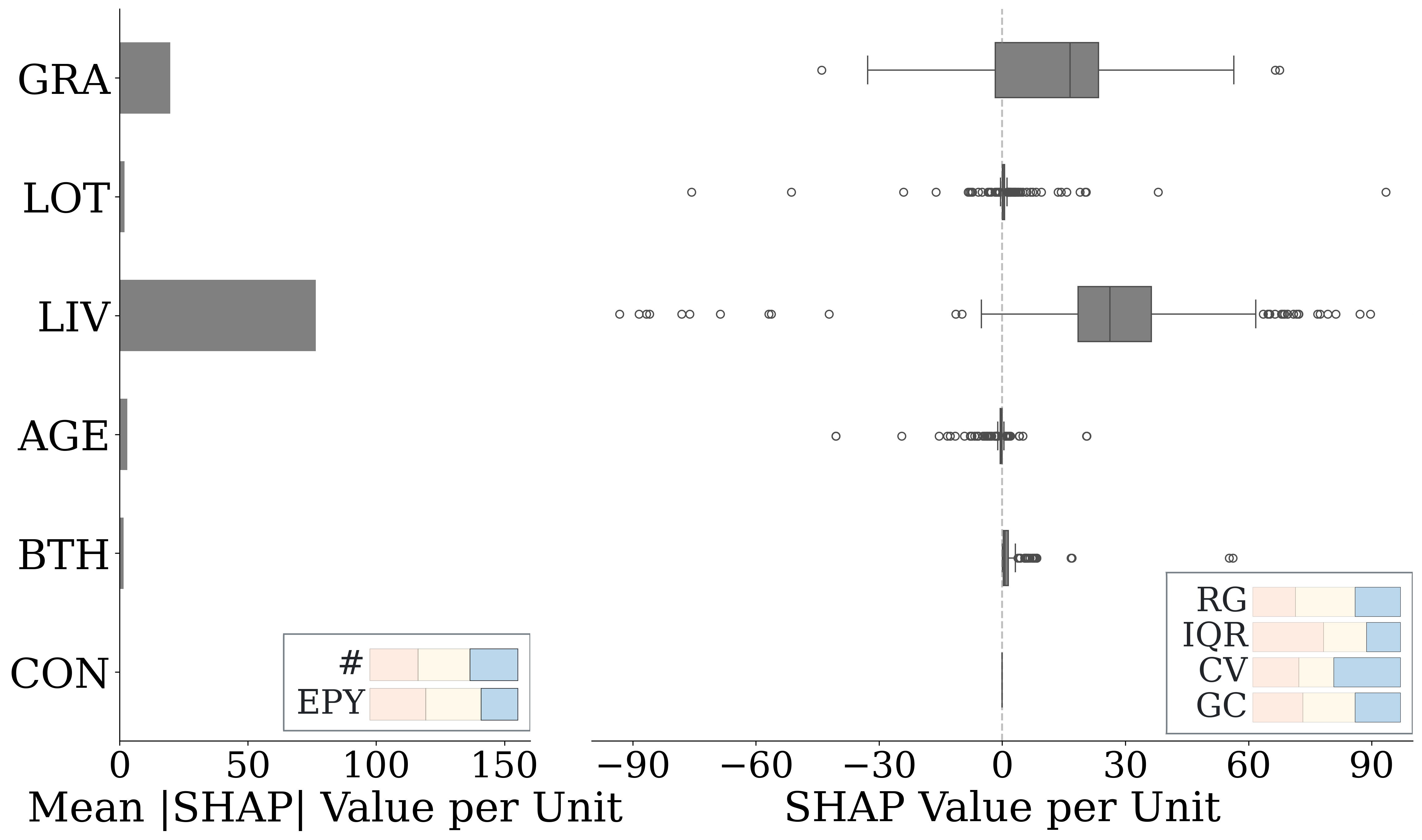}\\
	Type III (absent)\\
	\includegraphics[width=\textwidth]{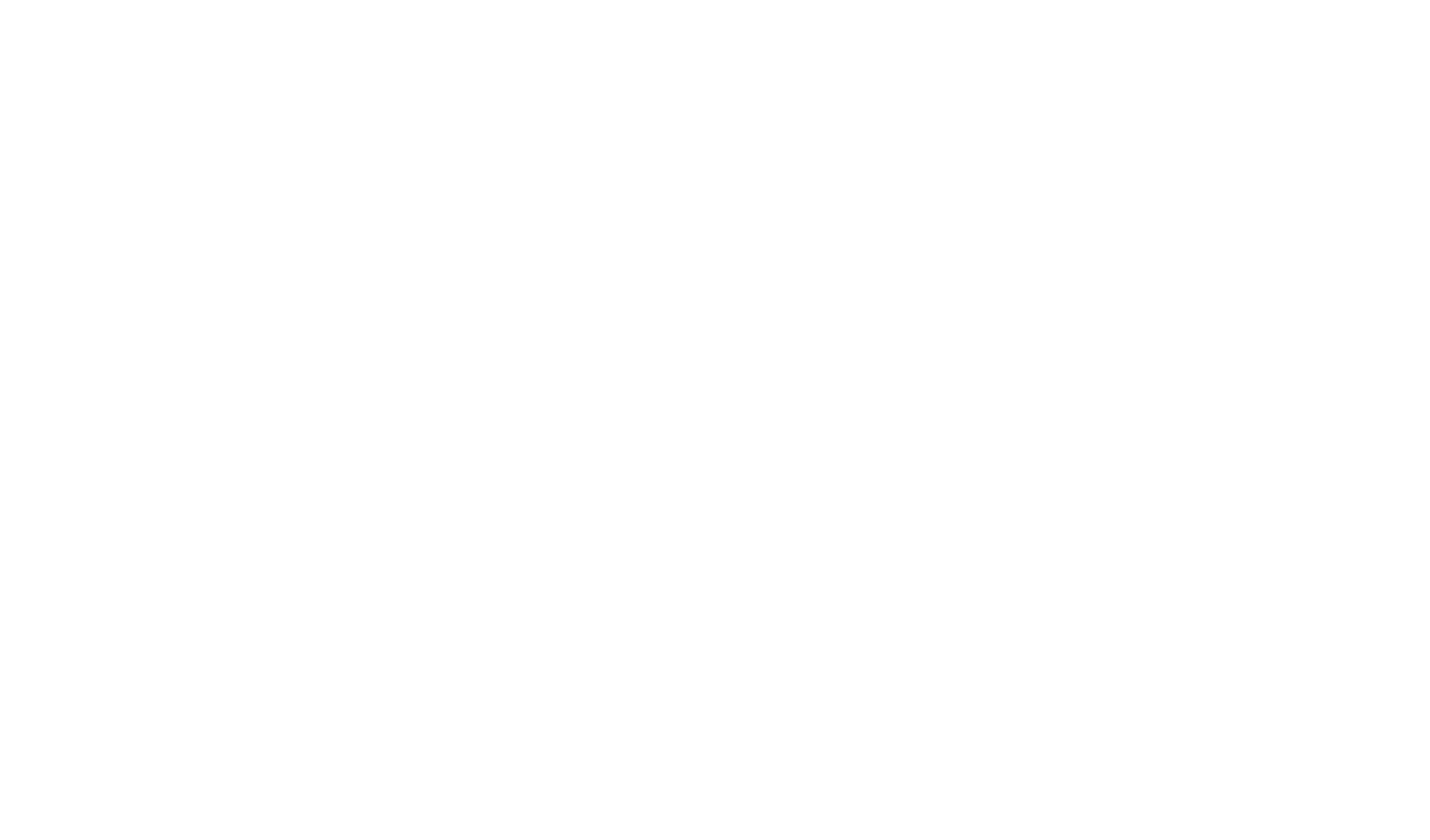}
	 \subcaption{Ours\\(msl=6, md=8, $\ell_{2}(gwr)$)}
	 \end{subfigure}
	\caption{The derived spatial communities of houses with similar feature coefficients and attributions in Seattle. The SHAP scores for each feature in each community are compared across the three models based on the predefined dispersion indicators.}
	\label{fig9:seattle-gwr}
\end{figure}

Instead, the models that account for spatial similarity between samples based on GWR coefficients of their input features (i.e., $\ell_{2}(gwr)$) divide the case studied counties into more spatially cohesive regions (see Figure \ref{fig7:fujian-gwr}). Theoretically, neighboring counties have a higher chance to pose similar feature coefficients in the GWR analysis. Therefore, the derived spatially-cohesive communities align better with the Tobler's first law of geography. Moreover, since the GWR coefficient quantifies the increase of prediction response to a unit increase of feature values, the derived feature attributions are indicators of the importance of each unit of feature values. That is, the derived feature attributions cancel out the difference in magnitude between samples. Similar to the results derived from the models based on feature values, our proposed model assigns high attributions to a smaller set of features (e.g., PRI, IND and IMP) than either the DT model or the GT model. More importantly, the feature attributions derived from our model are more distinctively different between the ``Type I'' counties and the ``Type II'' counties, which clearly reveals their differences in terms of the industrial sector (IND) without the interventions from POP and CLD. From a global perspective, the separation between the North-Western part and the South-Eastern part in Fujian is in a good agreement with the dominant economic development pattern in the last decades \citep{Chen2023}.  

Promisingly, the above findings and conclusions also hold for the house price prediction task in Seattle. As shown in Figure \ref{fig8:seattle-feature}, our proposed model can assign more robust explanations to samples with similar feature values than the DT model and the GT model, even though these samples are spatially separated from each other. Again, within each spatial community, the majority of the dispersion indicators of SHAP values are lower for our proposed model. Whereas, DT and GT divide the counties into groups with more heterogenous feature attributions. When replacing the similarity metric with the GWR coefficients (i.e., $\ell_{2}(gwr)$), our proposed model yields spatial partitions that are highly similar to the results from the DT model as shown in Figure \ref{fig9:seattle-gwr}. Despite that samples are assigned into similar groups based on our model and the DT model, we found that our model attributes features of interest more meaningfully and consistently than the DT model does. This fact provides a strong evidence for the superior of our proposed model towards robust model interpretability.

In summary, by comparing the feature attribution of samples within in the same spatial communities based on above visual analytics, we empirically confirmed that our proposed model yields more robust model explanations than the DT model and the GT model for the two case studies. The resulting spatial patterns are also more meaningful and informative. 

\subsubsection{Ablation Study}

Following the machine learning protocol, we further evaluate the variations of model performance by removing either the Moran's I module or the modularity maximization module from our proposed SX-GeoTree model. Results of the comparative analysis are listed in Table \ref{tab7:fujian-ablation} for the Fujian-county dataset and in Table \ref{tab8:seattle-ablation} for the Seattle-house dataset, respectively. 

\begin{table}[h!]
   \centering
      \caption{Ablation study on the Fujian-county dataset.}
   \begin{tabular}{@{} lllllll@{}} 
      \toprule
      & \multicolumn{2}{c}{\textbf{SX-GeoTree}} & \multicolumn{2}{c}{\textbf{w/o Moran's I}} & \multicolumn{2}{c}{\textbf{w/o Modularity}} \\
      \cmidrule(r){2-3} \cmidrule(r){4-5} \cmidrule(r){6-7}
      Metrics ({\scriptsize{similarity}})    & \scriptsize{$\ell_{2}(feature)$} & \scriptsize{$\ell_{2}(gwr)$} & \scriptsize{$\ell_{2}(feature)$} & \scriptsize{$\ell_{2}(gwr)$} & \scriptsize{$\ell_{2}(feature)$} & \scriptsize{$\ell_{2}(gwr)$}  \\
      \cmidrule(r){2-7}       
       Parameters ({\scriptsize{optimal}}) & \scriptsize{msl=5,md=5} & \scriptsize{msl=5,md=5} & \scriptsize{msl=5,md=5} & \scriptsize{msl=5,md=5} & \scriptsize{msl=5,md=5} & \scriptsize{msl=5,md=5} \\
      \midrule
       \textbf{R}$^{2}$ ({\scriptsize{test}}) 	& \textbf{0.77}	& \textbf{0.77}	& 0.67	& 0.76	& 0.74	& \textbf{0.77}\\
      \textbf{Moran's I} ({\scriptsize{residuals}}) &\textbf{0.0234}	&\textbf{0.0277} 	& 0.0838	& -0.1112	& 0.0376	& 0.0496\\
      \textbf{Modularity} ({\scriptsize{$\ell_{2}(feature)$}})	&\textbf{0.1917}	& 	& 0.1658	&	& 0.1305	& \\
     \textbf{Modularity} ({\scriptsize{$\ell_{2}(gwr)$}})		&	& \textbf{0.1679}	& 	& 0.1353	&	& 0.0654\\ 
      \bottomrule
   \end{tabular}
   \label{tab7:fujian-ablation}
\end{table}

\begin{table}[h!]
   \centering
      \caption{Ablation study on the Seattle-house dataset.}
   \begin{tabular}{@{} lllllll@{}} 
      \toprule
      & \multicolumn{2}{c}{\textbf{SX-GeoTree}} & \multicolumn{2}{c}{\textbf{w/o Moran's I}} & \multicolumn{2}{c}{\textbf{w/o Modularity}} \\
      \cmidrule(r){2-3} \cmidrule(r){4-5} \cmidrule(r){6-7}
      Metrics ({\scriptsize{similarity}})    & \scriptsize{$\ell_{2}(feature)$} & \scriptsize{$\ell_{2}(gwr)$} & \scriptsize{$\ell_{2}(feature)$} & \scriptsize{$\ell_{2}(gwr)$} & \scriptsize{$\ell_{2}(feature)$} & \scriptsize{$\ell_{2}(gwr)$}  \\
      \cmidrule(r){2-7}       
       Parameters ({\scriptsize{optimal}}) & \scriptsize{msl=6,md=8} & \scriptsize{msl=6,md=8} & \scriptsize{msl=6,md=8} & \scriptsize{msl=6,md=8} & \scriptsize{msl=6,md=8} & \scriptsize{msl=6,md=8} \\
      \midrule
       \textbf{R}$^{2}$ ({\scriptsize{test}}) 	& \textbf{0.75}	& \textbf{0.77}	& 0.74	& \textbf{0.77}	& \textbf{0.75}	& \textbf{0.77}\\
      \textbf{Moran's I} ({\scriptsize{residuals}}) & 0.0591	&\textbf{0.0544} 	& 0.1171	& 0.0806	& \textbf{0.0521}	& 0.0572\\
      \textbf{Modularity} ({\scriptsize{$\ell_{2}(feature)$}})	&\textbf{0.1051}	& 	& 0.0559	&	& 0.0522	& \\
     \textbf{Modularity} ({\scriptsize{$\ell_{2}(gwr)$}})		&	& \textbf{0.0765}	& 	& 0.0563	&	& 0.0525\\ 
      \bottomrule
   \end{tabular}
   \label{tab8:seattle-ablation}
\end{table}

In general, our proposed model with both the Moran's I module and the modularity maximization module achieves the best performances in terms of accuracy, spatial evenness of residuals, and consensus of explanations between similar samples. Whereas, the removal of the pre-designed modules from the proposed model undermines the model's performance in the corresponding aspect. Besides, it seems that the Moran's I module and the modularity maximization module mutually cooperate with each other in that the removal of one module also reduces the performance of the other module. The above ablation study quantitatively confirms the effectiveness of the two modules towards robust model prediction and explanation.

\section{Discussion}

Recently, machine learning methods reshape the geography community markedly, and provide advanced, effective, and efficient analytic tools for tackling classic and newly-emerging spatial problems. Mutually, a lot of machine learning methods have been improved by incorporating spatial thinking and knowledges. Nevertheless, the knowledge and know-how gap between the two domains still remains, which urges researchers to deeply understanding machine learning and geography fundamentals in research and practices.      

Our studies have provided a proof-of-concept that classic machine learning methods can be naturally extended from the aspatial space to the spatial space, which is particularly useful for applying machine learning techniques under the geospatial context. Taking the measurement of inter-sample similarity (which is fundamental for almost all machine learning algorithms) as an example, we demonstrated that the concept of ``spatial similarity'' can measure inter-sample similarities in a more informative manner. By incorporating the concept of spatial similarity into the widely-used decision tree regressor, we enhanced the model's ability to achieve smoother predictions and more robust explanations. This ability has enabled us to gain more useful insights in the two exemplar spatial prediction tasks.

Likewise, we also noticed that directly incorporating spatial objectives, such as spatial autocorrelation of residuals and explanations, into the main learning task can undermine the predictive ability of the original model. This limitation suggests that geographers should pay more attention in the architecture of machine learning models, in order to balance the benefit from adding spatial objectives and the compromise of the model's predictive power. Indeed, our proposed model has suffered such dilemmas, which thus direct us to test the more advanced bilevel optimization approach in our future works. As such, we argue that ``spatial'' should be ``specially'' treated in machining learning models.      

\section{Conclusions}

This study introduced SX-GeoTree, a self-explaining geospatial regression tree that explicitly incorporates spatial similarity of feature attributions into the tree induction process. By augmenting recursive binary splitting with (i) impurity reduction (MSE), (ii) spatial residual control (global Moran's I), and (iii) an explanation robustness module based on the modularity of a consensus similarity network (combining GWR coefficient space and SHAP attribution space), the model operationalizes the dual objectives of predictive accuracy and locally stable explanations. 

Methodologically, the work demonstrates that spatial similarity can be defined beyond geometric proximity - via shared localized stimulus-response relationships captured through GWR coefficients - and that aligning this structure with attribution space yields more trustworthy geospatial model interpretations. Conceptually, we reframed local Lipschitz continuity of explanations as a network community preservation problem, enabling scalable enforcement of attribution robustness without per-sample neighborhood optimization.

Practically, SX-GeoTree provides a transparent alternative to black-box spatial learners, producing interpretable splits and spatially consistent explanation fields that may better support policy communication, auditing, and spatial decision support. Empirical evaluation on two contrasting regression tasks (county-level GDP in Fujian and point-wise housing prices in Seattle) showed that SX-GeoTree substantially improves spatial evenness of residuals (e.g., Moran's I reduction of roughly a half over baselines) and the structural coherence of feature attributions (e.g., modularity gains of roughly 2 times over baselines), while maintaining competitive accuracy relative to standard decision trees. Where rich spatial partitioning (oblique / Gaussian splits) dominated (such as the Fujian case study), some accuracy trade-off emerged, highlighting the inherent tension in multi-objective learning under fixed weighted regularization. Ablation analyses confirmed that the Moran's I and modularity components are complementary: removing either degrades both spatial residual structure and explanation stability.

Our limitations include: (1) a proof-of-concept hard tree (no soft / differentiable gating); (2) reliance on fixed (unlearned) regularization weights; (3) use of a single attribution paradigm (TreeSHAP); and (4) no uncertainty quantification for explanations. Future work will explore soft decision tree or neural oblique split parameterizations, bilevel / Stackelberg optimization for dynamic objective balancing, integration of uncertainty- or fairness-aware attribution constraints, and accelerated modularity optimization for large datasets.

Overall, the proposed SX-GeoTree advances geographically informed interpretable machine learning by unifying spatial dependency control and explanation robustness within the tree growth process. The proposed framework offers a transferable template for embedding domain-specific structural priors (here, spatial similarity) directly into interpretable model architectures, supporting more stable and trustworthy geospatial analytics.

\section*{Acknowledgments}

The authors would thank Dr. Xiaoyue Xing from The University of Hong Kong for their constructive suggestions, the National Key Research and Development Program of China for their financial support, and the Academy of Digital China for providing the Fujian-county dataset. During the preparation of this manuscript, the Generative AI tools (e.g. GPT 4.1) provided by the Github Copilot have been used for language improvement .

\section*{Funding}

This work was supported by [National Key Research and Development Program of China] under Grant [2023YFB3906800].

\section*{Disclosure statement}

The authors report there are no competing interests to declare.

\bibliographystyle{chicago}
\bibliography{references}

\newpage
\appendix
\section{Supplementary Materials}

\subsection{Multicollinearity analysis of the Fujian-county data}

\begin{table}[h!]
   \centering
   \caption{Variance Inflation Factors (VIFs) of the selected variables in the Fujian-county dataset}
   \begin{tabular}{@{} lllllllll @{}} 
      \toprule
      \textbf{Feature}   	& X 	& Y 	& PRI 	& TER	& IND	& POP	& CLD	& IMP\\
      \midrule
      \textbf{VIF}      & 1.82	& 2.84	& 2.16	& 2.37	& 2.86	& 2.20	& 2.55	& 3.97 \\
      \bottomrule
   \end{tabular}
   \label{tabA1:fujian-vif}
\end{table}

\subsection{Multicollinearity analysis of the Seattle-house data}

\begin{table}[h!]
   \centering
   \caption{Variance Inflation Factors (VIFs) of the selected variables in the Seattle-house dataset}
   \begin{tabular}{@{} lllllllll@{}} 
      \toprule
      \textbf{Feature}   	& X 	& Y 	& BTH 	& LIV	& LOT	& GRA	& CON	& AGE\\
      \midrule
      \textbf{VIF}      & 1.35	& 1.10	& 2.70	& 3.51	& 1.36	& 2.59	& 1.16	& 1.96 \\
      \bottomrule
   \end{tabular}
   \label{tabA2:seattle-vif}
\end{table}

\newpage
\subsection{Robustness of model explanations in the Fujian-county data}

\begin{table}[h!]
   \centering
   \caption{Dispersion metrics (Fujian-county): intra-community SHAP attribution variability for the two detected communities. Lower values indicate more stable (robust) attributions.}
   \begin{tabular}{@{} l *{6}{r} @{}} 
      \toprule
      & \multicolumn{3}{c}{$\ell_{2}(\text{feature})$} & \multicolumn{3}{c}{$\ell_{2}(\text{gwr})$} \\
      \cmidrule(r){2-4} \cmidrule(r){5-7}
      & \textbf{DT} & \textbf{GT} & \textbf{Ours} & \textbf{DT} & \textbf{GT} & \textbf{Ours} \\
      \midrule
   Range (average)   & 217.14 & 321.22 & 156.30 & 4057.81 & 2058.44 & 1051.62 \\
   Range (Type I)    & 145.93 & 195.79 & 46.70  & 5612.47 & 2194.06 & 542.41  \\
   Range (Type II)   & 288.35 & 446.65 & 265.90 & 2503.16 & 1922.81 & 1560.83 \\
      \midrule
   IQR (average)     & 77.26  & 117.32 & 36.84  & 129.69  & 158.61  & 97.56   \\
   IQR (Type I)      & 37.67  & 51.67  & 8.41   & 134.08  & 106.79  & 65.57   \\
   IQR (Type II)     & 116.86 & 182.98 & 65.28  & 125.30  & 210.43  & 129.54  \\
      \midrule
   CV (average)      & 81.18  & 4.74   & 1.24   & 12.42   & 3.46    & 2.03    \\
   CV (Type I)       & 2.26   & 2.68   & 1.28   & 6.79    & 5.64    & 1.18    \\
   CV (Type II)      & 160.11 & 6.80   & 1.21   & 18.05   & 1.28    & 2.87    \\
      \midrule
   Entropy (average) & 1.65   & 1.58   & 1.57   & 1.87    & 1.57    & 1.60    \\
   Entropy (Type I)  & 1.44   & 1.56   & 1.61   & 1.72    & 1.58    & 1.53    \\
   Entropy (Type II) & 1.85   & 1.59   & 1.52   & 2.03    & 1.55    & 1.67    \\
      \midrule
   Gini (average)    & 0.34   & 0.43   & 0.21   & 0.54    & 0.32    & 0.35    \\
   Gini (Type I)     & 0.29   & 0.29   & 0.14   & 0.53    & 0.30    & 0.26    \\
   Gini (Type II)    & 0.39   & 0.56   & 0.28   & 0.56    & 0.33    & 0.43    \\
      \bottomrule
   \end{tabular}
   \label{tabA3:fujian-dispersion}
\end{table}

\newpage
\subsection{Robustness of model explanations in the Seattle-house data}

\begin{table}[h!]
   \centering
   \caption{Dispersion metrics (Seattle-house): intra-community SHAP attribution variability. Lower values indicate more stable (robust) attributions. A dash (--) indicates the community is absent for that model.}
   \begin{tabular}{@{} l *{6}{r} @{}} 
      \toprule
      & \multicolumn{3}{c}{$\ell_{2}(\text{feature})$} & \multicolumn{3}{c}{$\ell_{2}(\text{gwr})$} \\
      \cmidrule(r){2-4} \cmidrule(r){5-7}
      & \textbf{DT} & \textbf{GT} & \textbf{Ours} & \textbf{DT} & \textbf{GT} & \textbf{Ours} \\
      \midrule
      Range (average)     & 47.99  & 48.14  & 31.45  & 1425.09 & 1139.51 & 1028.29 \\
      Range (Type I)      & 37.25  & 27.77  & 10.95  & 1820.92 & 614.95  & 960.39 \\
      Range (Type II)     & 57.09  & 68.50  & 48.30  & 1029.25 & 1444.62 & 1096.18 \\
      Range (Type III)    & 49.62  & --     & 35.08  & --      & 1358.93 & -- \\
      \midrule
      IQR (average)       & 10.16  & 12.80  & 6.48   & 14.59   & 12.98   & 7.73 \\
      IQR (Type I)        & 7.05   & 5.80   & 2.43   & 13.43   & 12.75   & 7.93 \\
      IQR (Type II)       & 15.56  & 19.80  & 10.05  & 15.74   & 9.53    & 7.53 \\
      IQR (Type III)      & 7.87   & --     & 6.96   & --      & 16.68   & -- \\
      \midrule
      CV (average)        & 12.74  & 7.13   & 4.37   & 10.67   & 6.31    & 11.33 \\
      CV (Type I)         & 24.08  & 7.10   & 2.08   & 12.44   & 5.23    & 9.76 \\
      CV (Type II)        & 5.45   & 7.15   & 5.22   & 8.89    & 6.71    & 12.91 \\
      CV (Type III)       & 8.68   & --     & 5.82   & --      & 6.97    & -- \\
      \midrule
      Entropy (average)   & 1.61   & 1.51   & 1.14   & 1.76    & 1.53    & 1.03 \\
      Entropy (Type I)    & 1.77   & 1.59   & 1.16   & 1.81    & 1.92    & 0.94 \\
      Entropy (Type II)   & 1.30   & 1.43   & 1.16   & 1.71    & 1.70    & 1.12 \\
      Entropy (Type III)  & 1.75   & --     & 1.11   & --      & 0.98    & -- \\
      \midrule
      Gini (average)      & 0.61   & 0.65   & 0.48   & 0.69    & 0.69    & 0.62 \\
      Gini (Type I)       & 0.63   & 0.58   & 0.41   & 0.70    & 0.69    & 0.61 \\
      Gini (Type II)      & 0.58   & 0.71   & 0.55   & 0.69    & 0.72    & 0.63 \\
      Gini (Type III)     & 0.62   & --     & 0.48   & --      & 0.66    & -- \\
      \bottomrule
   \end{tabular}
   \label{tabA4:seattle-dispersion}
\end{table}

\end{document}